\documentclass[conference,compsoc]{IEEEtran}
\IEEEoverridecommandlockouts
\usepackage{cite}
\usepackage{amsmath,amssymb,amsfonts}
\usepackage{algorithmic}
\usepackage{graphicx}
\usepackage{textcomp}
\usepackage{tikz}
\usepackage{xcolor}
\usepackage{hyperref}
\usepackage{caption}
\usepackage{subcaption}
\usepackage{mathrsfs}
\usepackage{multirow}
\usepackage{placeins}

\usepackage[linesnumbered,ruled,vlined]{algorithm2e}

\def\BibTeX{{\rm B\kern-.05em{\sc i\kern-.025em b}\kern-.08em
    T\kern-.1667em\lower.7ex\hbox{E}\kern-.125emX}}
\begin{document}

\newcommand*\circled[1]{\tikz[baseline=(char.base)]{
            \node[shape=circle,fill,inner sep=2pt] (char) {\textcolor{white}{#1}};}}


\title{\textit{Get Rid Of Your Trail}: Remotely Erasing Backdoors in Federated Learning}

\author{
\IEEEauthorblockN{Manaar Alam}
\IEEEauthorblockA{
\textit{New York University Abu Dhabi}\\
Abu Dhabi, United Arab Emirates \\
alam.manaar@nyu.edu}
\and
\IEEEauthorblockN{Hithem Lamri}
\IEEEauthorblockA{
\textit{\'Ecole nationale sup\'erieure d'informatique}\\
Algiers, Algeria \\
kh\_lamri@esi.dz}
\and 
\IEEEauthorblockN{Michail Maniatakos}
\IEEEauthorblockA{
\textit{New York University Abu Dhabi}\\
Abu Dhabi, United Arab Emirates \\
michail.maniatakos@nyu.edu}
}

\maketitle

\begin{abstract}
Federated Learning (FL) enables collaborative deep learning training across multiple participants without exposing sensitive personal data. However, the distributed nature of FL and the unvetted participants' data makes it vulnerable to backdoor attacks. In these attacks, adversaries inject malicious functionality into the centralized model during training, leading to intentional misclassifications for specific adversary-chosen inputs. While previous research has demonstrated successful injections of persistent backdoors in FL, the persistence also poses a challenge, as their existence in the centralized model can prompt the central aggregation server to take preventive measures to penalize the adversaries. Therefore, this paper proposes a methodology \emph{that enables adversaries to effectively remove backdoors from the centralized model} upon achieving their objectives or upon suspicion of possible detection. The proposed approach extends the concept of machine unlearning and presents strategies to preserve the performance of the centralized model and simultaneously prevent over-unlearning of information unrelated to backdoor patterns, making the adversaries stealthy while removing backdoors.
To the best of our knowledge, this is the first work that explores machine unlearning in FL to remove backdoors to the benefit of adversaries. 
Exhaustive evaluation considering image classification scenarios demonstrates the efficacy of the proposed method in efficient backdoor removal from the centralized model, injected by state-of-the-art attacks across multiple configurations.
\end{abstract}


\section{Introduction}
Federated Learning (FL)~\cite{DBLP:conf/aistats/McMahanMRHA17,DBLP:journals/corr/KonecnyMYRSB16,DBLP:journals/corr/KonecnyMRR16} is a widespread decentralized deep learning framework that facilitates massively distributed collaborative training involving thousands or even millions of participants~\cite{DBLP:conf/mlsys/BonawitzEGHIIKK19,DBLP:journals/corr/abs-1811-03604}. The primary objective of FL is to collaboratively develop a high-accuracy shared model by aggregating locally fine-tuned models on each participant's dataset. FL offers better generalization than conventional machine learning techniques as the local models are trained on more diverse and potentially larger decentralized data sources, resulting in enhanced model performance~\cite{DBLP:conf/aistats/McMahanMRHA17}. Furthermore, the inherent operations of FL also provide a promising framework in which participants are not required to disclose sensitive or personal information to the central model aggregation server or other participants in order to train the shared model, which is convenient for settings where data privacy is of utmost importance. Recently, FL has gained a huge surge in popularity across several real-world applications, from smartphones~\cite{DBLP:conf/www/YangWXCBLL21}, finance~\cite{DBLP:series/lncs/LongT0Z20}, advertising~\cite{DBLP:journals/corr/abs-2209-15635}, and healthcare~\cite{DBLP:journals/tist/AntunesCKYE22}, to autonomous vehicles~\cite{DBLP:journals/tits/LiTZL022}, where data sensitivity poses a challenge to conventional data-sharing approaches.

In most FL scenarios, it is typically assumed that the central aggregation server does not have the capability to validate the training data of participants or to control their individual training processes directly. Consequently, despite the numerous advantages that FL offers, it has been demonstrated to be vulnerable to a specific type of security threat known as \textit{backdoor attacks}~\cite{DBLP:conf/aistats/BagdasaryanVHES20,DBLP:conf/nips/WangSRVASLP20,DBLP:conf/icml/ZhangPSYMMR022,DBLP:conf/iclr/XieHCL20,DBLP:journals/corr/abs-1911-07963,DBLP:journals/corr/abs-2011-07429,DBLP:journals/corr/abs-2208-06176,DBLP:journals/corr/abs-2210-09305,DBLP:journals/corr/abs-2205-13523}. During such an attack, an adversary can manipulate the local models of a select group of participants by poisoning their local datasets with so-called \textit{trigger patterns}\footnote{A trigger pattern is a specific input feature or set of features that, when present, causes a model with a backdoor to produce incorrect adversary-chosen predictions.}, subsequently embedding effective backdoor functionality into the shared global model. Consequently, the compromised global model produces incorrect adversary-chosen predictions for specific inputs containing the trigger patterns while preserving the model's performance on legitimate tasks.

The growing incidence of backdoor attacks in the context of FL has prompted considerable efforts to devise effective defense strategies. A fundamental approach in countering these attacks involves preventing backdoor insertions through anomaly detection methods~\cite{DBLP:conf/nips/BlanchardMGS17,DBLP:conf/acsac/ShenTS16,DBLP:conf/uss/NguyenRCYMFMMMZ22,DBLP:conf/ndss/RiegerNMS22,DBLP:journals/corr/abs-2202-11196,DBLP:journals/corr/abs-2301-09508,DBLP:journals/corr/abs-2210-12873}, which help identify malicious participants by measuring the similarities between participants' local models and the global model prior to aggregation. An alternative strategy involves deliberately incorporating selective noise into participants' local model~\cite{DBLP:conf/icassp/MiaoYHLH22,DBLP:journals/corr/abs-1911-07963}, diminishing the impact of backdoor insertions. Additionally, robust byzantine-tolerant aggregation schemes have also been proposed as another line of defense~\cite{DBLP:conf/icml/YinCRB18,DBLP:conf/aaai/OzdayiKG21,DBLP:conf/aistats/PandaMBCM22,DBLP:journals/corr/abs-2102-02402}. Nevertheless, these approaches frequently result in reduced global model accuracy or have been shown to be vulnerable to more sophisticated attacks~\cite{DBLP:conf/icml/ZhangPSYMMR022,DBLP:journals/corr/abs-2301-08170}. Another line of defense strategy has recently been developed that leverages participant feedback to detect backdoor attacks~\cite{DBLP:conf/icdcs/AndreinaMMK21,DBLP:conf/ih/ZhaoWLLM21,DBLP:journals/corr/abs-2210-07714,DBLP:journals/corr/abs-2011-01767}, analyzing the global model with benign data procured from trustworthy participants.

Recent advancements in backdoor attacks in FL concentrate on the persistence of backdoors in the global model for extended periods~\cite{DBLP:conf/aistats/BagdasaryanVHES20,DBLP:conf/icml/ZhangPSYMMR022,DBLP:journals/corr/abs-2205-13523}. This is due to the fact that backdoors do not naturally persist through multiple rounds when adversaries cease to inject them~\cite{DBLP:conf/icml/ZhangPSYMMR022}. However, the persistence of backdoors poses challenges for adversaries, as their detection in the global model can prompt the central aggregation server to implement preventive actions. Such actions may include analyzing participants based on historical contributions and penalizing those who consistently contribute malicious updates~\cite{DBLP:journals/corr/abs-2011-10464}. 

After the attackers have achieved their desired objectives or have reasons to believe they can be detected, they may want to remove the backdoors and erase their footprint. While software backdoors can be easily removed with a software update, backdoor removal in FL cannot be as simple due to the decentralized nature of the learning process. \textit{Therefore, in this paper, we propose a methodology that enables adversaries to effectively remove backdoors from the global model.} 
It should be emphasized that we focus on backdoor removal and not backdoor injection. For the latter, we employ state-of-the-art backdoor insertion techniques;  developing a new backdoor insertion method is beyond the scope of this study.

Our proposed methodology extends the technique of \textit{machine unlearning}~\cite{DBLP:conf/sp/CaoY15,DBLP:conf/sp/BourtouleCCJTZL21,DBLP:conf/nips/SekhariAKS21}, an emerging approach allowing the selective elimination of specific data points' influence on a trained machine learning model without necessitating complete retraining from scratch. Machine unlearning is driven by data privacy regulations, such as the General Data Protection Regulation~(GDPR) and California Consumer Privacy Act~(CCPA), which grant users the right to revoke their data. The primary objective of machine unlearning is to facilitate the efficient removal of data instances that should no longer be used by the model, particularly in cases where retraining the entire model would be resource-intensive or time-consuming. Previous research has explored machine unlearning within the context of FL to ensure data privacy, maintain regulatory compliance, and uphold users' right to be forgotten~\cite{DBLP:conf/infocom/LiuXYWL22,DBLP:conf/iwqos/LiuMYWL21,DBLP:journals/network/WuGWHZD22,DBLP:journals/corr/abs-2207-05521}. \emph{However, leveraging machine unlearning to benefit an adversary in an FL framework remains unexplored.} While machine unlearning has been employed as a defense against backdoor attacks in conventional machine learning contexts~\cite{DBLP:conf/infocom/LiuFCLMWM22}, its  application to the training dynamics of FL is yet to be investigated. Our study highlights that directly applying such methods within an FL framework leads to poor solutions for the removal of backdoors, thereby necessitating appropriate investigation and adaptation for optimal performance.

The implementation of machine unlearning as a method to remove backdoors poses multiple challenges that must be addressed to ensure its effectiveness. One such challenge is the occurrence of \textit{catastrophic forgetting}~\cite{DBLP:journals/network/WuGWHZD22,DBLP:journals/corr/abs-2207-05521,DBLP:journals/corr/abs-2003-10933}, a phenomenon where features not intended for unlearning are unintentionally removed during the unlearning process, negatively impacting the overall performance of the model. Additionally, the unlearning procedure can inadvertently generate models that deviate significantly from the global model, increasing the probability of the central aggregation server detecting malicious activity. It is also crucial to recognize that the parameters of the global model possess varying degrees of significance in determining its outcome. Uniformly modifying all parameters during unlearning can produce suboptimal results as it disregards the differential significance of individual parameters within the FL framework.
To address these concerns, we employ two strategies: (1) \textit{memory preservation}, preserving the global model's memory on the legitimate task, and (2) \textit{dynamic penalization}, deterring the over-unlearning of information unrelated to trigger patterns, thereby promoting a more balanced and efficient unlearning process.

In our study, we assess the effectiveness of the proposed backdoor removal methodology by utilizing two well-established image classification datasets, CIFAR-10~\cite{krizhevsky2009learning} and CIFAR-100~\cite{krizhevsky2009learning}, in conjunction with two standard neural network classifiers, VGG-11~\cite{DBLP:journals/corr/SimonyanZ14a} and ResNet18~\cite{DBLP:conf/cvpr/HeZRS16}. Our evaluation incorporates three commonly used natural \textit{semantic backdoor patterns}\footnote{We primarily focus on semantic triggers, which constitute natural images that do not require pixel modification, making them more threatening than another type of trigger pattern known as \textit{pixel-pattern triggers}~\cite{DBLP:conf/aistats/BagdasaryanVHES20,DBLP:journals/corr/abs-2205-13523}. Pixel-pattern triggers necessitate an attacker to modify the pixels of digital images, particularly at test time, so that the model can misclassify the altered image. Despite our primary focus on semantic triggers, the proposed methodology can be equally applied to pixel-pattern triggers.}~\cite{DBLP:conf/aistats/BagdasaryanVHES20,DBLP:conf/icml/ZhangPSYMMR022,DBLP:journals/corr/abs-1911-07963} and two state-of-the-art backdoor insertion strategies - \textit{Constrain-and-Scale}~\cite{DBLP:conf/aistats/BagdasaryanVHES20} and \textit{Neurotoxin}~\cite{DBLP:conf/icml/ZhangPSYMMR022} - across various FL configurations. The results of our comprehensive analysis demonstrate that, regardless of any FL configuration, the proposed methodology consistently and successfully removes backdoors from the global model without alerting the central aggregation server.

\subsection*{Contribution}
The summary of the key contributions of this paper is described as follows:
\begin{itemize}
    \item We present a methodology that \textit{enables adversaries within an FL framework to effectively remove backdoors from the global model}, thereby erasing their footprint and evading detection by the central aggregation server.
    \item We utilize the principles of machine unlearning to design the backdoor removal mechanism, which, to the best of our knowledge, has not been explored previously in the context of benefiting adversaries in an FL framework.
    \item We introduce two strategies – \textit{memory preservation} and \textit{dynamic penalization} – that effectively address the challenges of catastrophic forgetting inherent in machine unlearning and maintain stealthiness throughout the unlearning process.
\end{itemize}

The remainder of this paper is organized as follows: Section~\ref{sec:prlim} outlines the preliminary concepts used throughout the paper. Section~\ref{sec:threat_model} illustrates the problem landscape and the threat model utilized in this paper. Section~\ref{sec:method} provides an in-depth study of the proposed backdoor removal methodology. Section~\ref{sec:setup} outlines the experimental setup used in this paper, while Section~\ref{sec:results} comprehensively demonstrates the experimental results. Section~\ref{sec:related_work} reviews relevant literature. Finally, Section~\ref{sec:conclusion} concludes the paper with a brief discussion on future research direction.

\section{Preliminaries}\label{sec:prlim}
In this section, we provide a brief overview of the concepts used in this paper. We first discuss the basic working principles of FL. Next, we discuss the critical challenges posed by backdoor attacks on FL. Finally, we discuss the notion of machine unlearning, an essential technique that facilitates the removal of specific data from a trained model while maintaining its overall performance.

\subsection{Federated Learning}
FL is a privacy-centric, decentralized framework designed for training collaborative machine learning models. FL framework remains in compliance with the GDPR and CCPA by eliminating the need for central data storage - as is customary in traditional machine learning approaches. This framework ensures data privacy by exchanging only model updates, such as gradients or weights, instead of raw data. Consequently, the FL framework provides a secure and privacy-preserving solution for training machine learning models on distributed datasets, making it especially suitable for applications that handle sensitive or confidential data. Let us consider an FL framework with $n$ participants $\{p_1, \dots, p_n\}$ and a central aggregation server $\mathbf{CS}$ that collaboratively train a global model $\mathcal{G}$. In each training round $r$, the FL framework comprises of the following three steps:
\begin{itemize}
    \item \textit{Participant Selection:} $\mathbf{CS}$ selects a random subset of $m$ participants $\mathcal{S}_m$ that satisfy predefined conditions to participate in the training and broadcasts the current global model $\mathcal{G}^{(r)}$ to all selected participants, where $\mathcal{G}^{(r)}$ is the global model at the $r$-th training~round.
    \item \textit{Local Model Updates:} Starting from $\mathcal{G}^{(r)}$, each selected participant $p_i \in \mathcal{S}_m$ trains respective local models $\mathcal{L}^{(r+1)}_{p_i}$ using its private data $\mathcal{D}_{p_i}$ by running standard optimization algorithms, and transmits the difference $\mathcal{L}^{(r+1)}_{p_i} - \mathcal{G}^{(r)}$ back to $\mathbf{CS}$, where $\mathcal{L}^{(r+1)}_{p_i}$ is the updated local model of participant $p_i$ after it is selected at the $r$-th training round.
    \item \textit{Global Aggregation:} $\mathbf{CS}$ aggregates all the received updates into the global model $\mathcal{G}^{(r+1)}$.
\end{itemize}
In this work, without loss of generality, we assume that $\mathbf{CS}$ employs \textit{Federated Averaging} (FedAvg)~\cite{DBLP:conf/aistats/McMahanMRHA17} aggregation method as it is commonly applied in FL~\cite{DBLP:conf/ccs/BonawitzIKMMPRS17,DBLP:conf/nips/SmithCST17} and related works on backdoor attacks in FL~\cite{DBLP:conf/aistats/BagdasaryanVHES20,DBLP:conf/nips/WangSRVASLP20,DBLP:conf/icml/ZhangPSYMMR022}. Under FedAvg, the global model is updated by performing a weighted average on the received updates as:
    \begin{equation}
        \mathcal{G}^{(r+1)} = \mathcal{G}^{(r)} + \sum_{p_i \in \mathcal{S}_m}\frac{n_{p_i}}{n_{\mathcal{S}_m}}(\mathcal{L}^{(r+1)}_{p_i} - \mathcal{G}^{(r)})
    \end{equation}
\noindent where $n_{p_i} = |\mathcal{D}_{p_i}|$ and $n_{\mathcal{S}_m} = \sum_{p_i \in \mathcal{S}_m} n_{p_i}$ is the total number of training data used at the selected round. Learning does not stop even after the global model $\mathcal{G}$ converges. Participants, throughout their deployment, continuously update $\mathcal{G}$. A malicious participant thus always has an opportunity to be selected and influence $\mathcal{G}$.

\subsection{Backdoor Attacks in FL}
FL frameworks are becoming increasingly vulnerable to a major security threat known as backdoor attacks~\cite{DBLP:conf/aistats/BagdasaryanVHES20,DBLP:conf/nips/WangSRVASLP20,DBLP:conf/icml/ZhangPSYMMR022,DBLP:conf/iclr/XieHCL20,DBLP:journals/corr/abs-1911-07963,DBLP:journals/corr/abs-2011-07429,DBLP:journals/corr/abs-2208-06176,DBLP:journals/corr/abs-2210-09305,DBLP:journals/corr/abs-2205-13523}, In these attacks, an adversary, denoted as $\mathcal{A}$, strategically manipulates the local models of a subset of compromised participants to create poisoned local models. As these poisoned models are aggregated into the global model $\mathcal{G}$, the latter's properties are gradually altered due to the accumulation of poisoned updates. The primary objective of $\mathcal{A}$ is to develop a poisoned global model $\hat{\mathcal{G}}$ that functions normally for the majority of inputs but produces incorrect targeted predictions for specific adversary-chosen inputs known as the \textit{trigger set}. Backdoor attacks in FL are typically carried out using a combination of \textit{model-replacement} and \textit{data-poisoning} strategies~\cite{DBLP:conf/aistats/BagdasaryanVHES20,DBLP:conf/nips/WangSRVASLP20,DBLP:conf/icml/ZhangPSYMMR022}. Model replacement involves $\mathcal{A}$ manipulating the training process or modifying the trained models of compromised participants, while data poisoning comprises $\mathcal{A}$ manipulating the training dataset of these participants. In order to inject backdoors into the global model, $\mathcal{A}$ first introduces manipulated data containing the trigger set into the training datasets of compromised participants, which can be accomplished by flipping data labels or embedding triggers into data samples, such as adding a unique pixel pattern to images~\cite{DBLP:conf/aistats/BagdasaryanVHES20,DBLP:journals/corr/abs-2205-13523}, in conjunction with label flipping~\cite{DBLP:conf/aistats/BagdasaryanVHES20,DBLP:conf/nips/WangSRVASLP20,DBLP:conf/icml/ZhangPSYMMR022}. In the subsequent step, $\mathcal{A}$ adjusts the training process execution by modifying the parameters of the resulting local models trained with poisoned data and scaling the model updates to optimize the attack's impact in the global model while evading detection by the anomaly detector of the central aggregation server~\cite{DBLP:conf/nips/BlanchardMGS17,DBLP:conf/acsac/ShenTS16,DBLP:conf/uss/NguyenRCYMFMMMZ22,DBLP:conf/ndss/RiegerNMS22,DBLP:journals/corr/abs-2202-11196,DBLP:journals/corr/abs-2301-09508}.

\subsection{Machine Unlearning}
Machine unlearning is an emerging area of research that focuses on the efficient removal or "unlearning" of specific data points from a trained model~\cite{DBLP:conf/sp/CaoY15,DBLP:conf/sp/BourtouleCCJTZL21,DBLP:conf/nips/SekhariAKS21}. Machine unlearning has gained significant importance due to growing concerns surrounding data privacy, regulatory compliance, and the need to rectify mislabeled or outdated data, as well as accommodate user requests for data removal. Numerous studies have explored various approaches to machine unlearning~\cite{DBLP:conf/cvpr/GolatkarAS20,DBLP:conf/eccv/GolatkarAS20,DBLP:conf/aaai/GravesNG21,DBLP:conf/icml/GuoGHM20}, with approximate unlearning being a notable approach that aims to eliminate the impact of selected data points on the model without requiring a complete retraining process~\cite{DBLP:conf/nips/SekhariAKS21,DBLP:conf/alt/Neel0S21,DBLP:journals/corr/abs-2209-02299}. Approximate unlearning is achieved by employing techniques such as influence functions to iteratively estimate the effect of the data points targeted for removal on the model parameters and subsequently adjusting the model parameters based on the estimated influence of the data point~\cite{DBLP:conf/sp/BourtouleCCJTZL21,DBLP:conf/aaai/MarchantRA22,DBLP:conf/ndss/WarneckePWR23}. The approximate unlearning process is not only computationally efficient but also offers a balance between data deletion and model accuracy. Owing to the benefits of approximate unlearning, our proposed backdoor removal method is designed following the principles of approximate unlearning.

\section{Problem Landscape and Threat Model}\label{sec:threat_model}
In this section, we first present an overview of the problem landscape within a conventional FL framework. Subsequently, we describe the threat model considered in the context of this paper.

\subsection{Problem Landscape}
\begin{figure}[!t]
    \centering
    \includegraphics[width=0.79\linewidth]{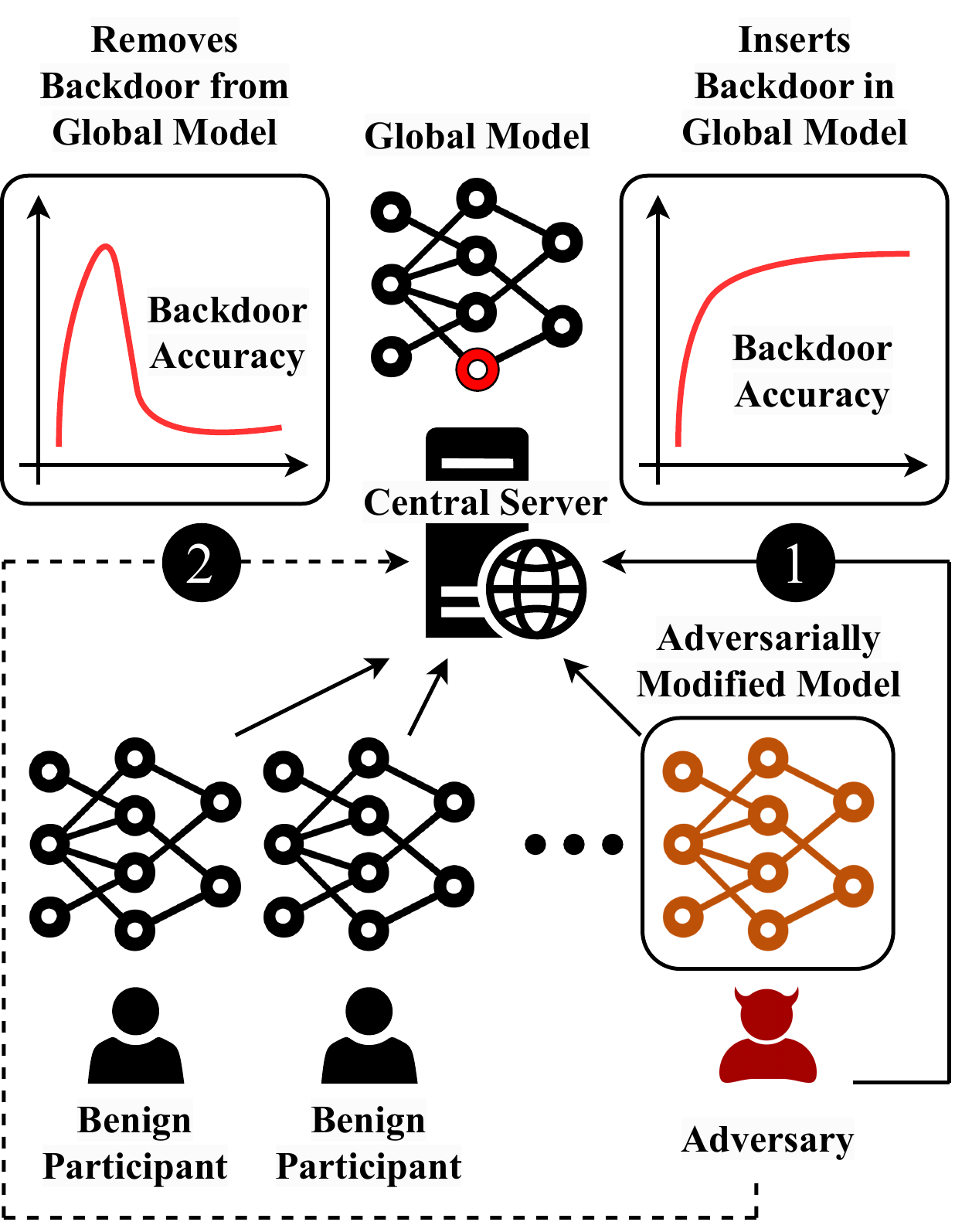}
    \caption{\textbf{Problem Landscape:} A Federated Learning framework featuring a single compromised participant under the control of an adversary. The adversary is capable of \protect\circled{1} embedding backdoors into the global model, achieving high \textit{backdoor accuracy} (ref. Section~\ref{sec:eval_metric}), and \protect\circled{2} removing backdoors from the global model using our proposed methodology, resulting in low \textit{backdoor accuracy}.}
    \label{fig:threat_model}
\end{figure}

We consider an FL framework in which an adversary $\mathcal{A}$ controls a single compromised participant $p_c$. Although our primary focus is on one compromised participant, recent research demonstrates that an increased number of compromised participants can further amplify the effectiveness of the attack~\cite{DBLP:conf/aistats/BagdasaryanVHES20,DBLP:conf/raid/FungYB20,DBLP:journals/corr/abs-1911-07963,DBLP:conf/nips/WangSRVASLP20,DBLP:conf/iclr/XieHCL20}. $\mathcal{A}$ covertly embeds triggers or patterns into the global model $\mathcal{G}$ by adversarially modifying the compromised local model $\mathcal{L}_{p_c}$, leading the modified global model to generate a \textit{desired incorrect output} upon encountering these triggers during inference, known as \textit{targeted backdoor attacks}. The modified global model functions accurately for the majority of inputs while displaying malicious behavior only for specified triggers, thereby maintaining its stealthiness. Additionally, $\mathcal{A}$ can conceal traces of malicious modification of the global model by selectively removing backdoors from it through modification of $\mathcal{L}_{p_c}$. A brief overview of the problem landscape is demonstrated~in~Figure~\ref{fig:threat_model}.

\subsection{Threat Model}
\subsubsection{Capability of the Adversary}
Following existing research on backdoor attacks in FL~\cite{DBLP:conf/aistats/BagdasaryanVHES20,DBLP:conf/nips/WangSRVASLP20,DBLP:conf/icml/ZhangPSYMMR022,DBLP:conf/iclr/XieHCL20}, we consider an adversary $\mathcal{A}$ with complete control over a malicious participant.
The adversary $\mathcal{A}$ possesses: \textbf{(1)} \textit{control over the compromised participant} - it has authority over the local training data $\mathcal{D}_{p_c}$, it can dictate the local training procedures as well as manipulate hyperparameters such as epochs and learning rate, it can modify the weights of the resulting local model $\mathcal{L}^{(r+1)}_{p_c}$ before submitting it for aggregation, and it has the flexibility to adaptively change local training from round to round, and \textbf{(2)} \textit{white-box access to the received global model} $\mathcal{G}$, allowing for direct parameter inspection.
However, $\mathcal{A}$ neither has control over other benign participants nor has access to their data or local updates. Furthermore, $\mathcal{A}$ has no control over the aggregation algorithm that combines participants' updates into the global model and can only interact with the central server $\mathbf{CS}$ through the compromised client. We assume that $\mathcal{A}$ generates the local model by correctly applying the FL-prescribed training algorithm to its local data and participates in local model training if and only if it is selected by the server.

\subsubsection{Goal of the Adversary}
The objectives of the adversary $\mathcal{A}$ are two-fold: \textbf{(1)} \textit{Impact:} $\mathcal{A}$ first crafts a poisoning dataset $\mathcal{D}$, which is a combination of the local training data of the compromised participant $\mathcal{D}_{p_c}$ and a strategically selected trigger dataset $\mathcal{D}_{t}$. The trigger dataset consists of specific attacker-chosen inputs and desired incorrect label $t$. By utilizing this new dataset $\mathcal{D}$, the adversary trains a malicious local model $\hat{\mathcal{L}}^{(r+1)}_{p_c}$, initialized with the received global model $\mathcal{G}^{(r)}$. The goal of the adversary is to manipulate the global model $\mathcal{G}^{(r)}$ into a modified version $\hat{\mathcal{G}}^{(r+1)}$, such that $\hat{\mathcal{G}}^{(r+1)}$ yields the desired incorrect predictions for any input $x$ belonging to $\mathcal{D}_{t}$. $\mathcal{A}$ also aims to construct $\hat{\mathcal{G}}^{(r+1)}$ such that it closely mimics the behavior of $\mathcal{G}^{(r)}$ on all other inputs not in $\mathcal{D}_{t}$, i.e.,
\begin{equation}
\hat{\mathcal{G}}^{(r+1)}(x) = \begin{cases}
t \neq \mathcal{G}^{(r)}(x) & \forall x \in \mathcal{D}_t\\
\mathcal{G}^{(r)}(x) & \forall x \notin \mathcal{D}_t
\end{cases}
\end{equation}
\textbf{(2)}~\textit{Stealthiness:} $\mathcal{A}$ aims to construct $\hat{\mathcal{L}}^{(r+1)}_{p_c}$ as indistinguishable as possible from the local models of other benign participants to make it difficult for $\mathbf{CS}$ to identify any malicious activity. $\mathcal{A}$ aims to keep high similarity of $\hat{\mathcal{L}}^{(r+1)}_{p_c}$ with other benign models that can be estimated by comparing $\hat{\mathcal{L}}^{(r+1)}_{p_c}$ to $\mathcal{G}^{(r)}$ or to a local model trained on benign data $\mathcal{D}_{p_c}$ as $\mathcal{A}$ does not have access to the local updates of other benign participants.

\subsubsection{Poisoning Strategies}\label{sec:poisoning}
Following existing research on backdoor attacks in FL~\cite{DBLP:conf/aistats/BagdasaryanVHES20,DBLP:conf/nips/WangSRVASLP20,DBLP:journals/corr/abs-1911-07963,DBLP:journals/corr/abs-2205-13523}, we consider that the adversary $\mathcal{A}$ carries out \textit{model replacement attack}
in the global model by transmitting back $\frac{n_{\mathcal{S}_m}}{n_{p_c}}(\hat{\mathcal{L}}^{(r+1)}_{p_c} - \mathcal{G}^{(r)}) + \mathcal{G}^{(r)}$ instead of $\hat{\mathcal{L}}^{(r+1)}_{p_c} - \mathcal{G}^{(r)}$ to the central server $\mathbf{CS}$ whenever selected. Also, following recent research~\cite{DBLP:conf/aistats/BagdasaryanVHES20,DBLP:conf/nips/WangSRVASLP20,DBLP:conf/icml/ZhangPSYMMR022,DBLP:journals/corr/abs-1911-07963,DBLP:journals/corr/abs-2205-13523}, we consider three distinct poisoning strategies through which an adversary may be chosen for participation. These strategies include: \textbf{(1)} \textit{continuous selection}, wherein the compromised participant is selected at every training round; \textbf{(2)} \textit{fixed-frequency selection}, wherein the compromised participant is involved in every $f$ training rounds\footnote{In the literature~\cite{DBLP:conf/aistats/BagdasaryanVHES20,DBLP:journals/corr/abs-1911-07963}, $f$ is usually chosen to be equal to $\frac{1}{|\mathcal{S}_m|}$.}; and \textbf{(3)} \textit{random selection}, wherein each participant is selected uniformly without imposing any constraints. 

\section{Proposed Methodology}\label{sec:method}
In this paper, our primary contribution is a robust approach that effectively removes backdoors from the global model within an FL framework. While the exploration of backdoor insertion in FL has garnered significant research interest~\cite{DBLP:conf/aistats/BagdasaryanVHES20,DBLP:conf/nips/WangSRVASLP20,DBLP:conf/icml/ZhangPSYMMR022,DBLP:conf/iclr/XieHCL20,DBLP:journals/corr/abs-1911-07963,DBLP:journals/corr/abs-2011-07429,DBLP:journals/corr/abs-2208-06176,DBLP:journals/corr/abs-2210-09305,DBLP:journals/corr/abs-2205-13523}, our focus in this section lies predominantly in discussing a comprehensive backdoor removal strategy applicable to any state-of-the-art backdoor insertion techniques. We first discuss the motivation and challenges associated with the proposed methodology, followed by an in-depth discussion of the approach. 

\subsubsection*{Motivation}
\begin{figure}[!t]
    \centering
    \includegraphics[width=\linewidth]{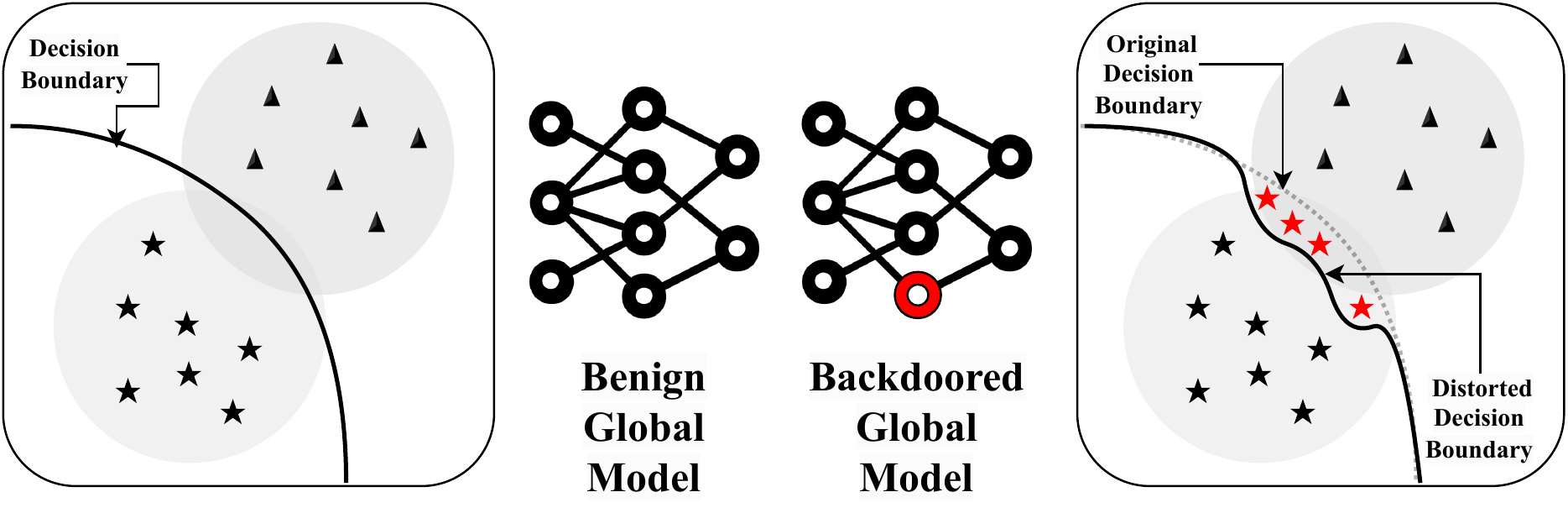}
    \caption{A basic illustration of decision boundary distortion in the global model resulting from a typical backdoor attack, highlighting the impact on classification outcomes.}
    \label{fig:unlearning_motivation}
\end{figure}

In a typical backdoor attack in an FL framework, an adversary poisons the dataset of a compromised participant with trigger images, consequently deceiving the participant into learning the trigger patterns during the local model training phase. Subsequently, the trigger patterns are transferred to the global model as well. As a result, the decision boundaries of the global model experience distortion, diverging from that of an uncorrupted model. Figure~\ref{fig:unlearning_motivation} provides a basic illustration of the distortion of decision boundaries in the global model due to a typical backdoor attack.
Interestingly, this distortion offers a potential avenue for identifying the presence of backdoors within the global model~\cite{DBLP:conf/icdcs/AndreinaMMK21,DBLP:conf/ih/ZhaoWLLM21,DBLP:journals/corr/abs-2210-07714,DBLP:journals/corr/abs-2011-01767}. However, a sophisticated attacker can conceal the traces of malicious modifications in the global model by \textit{unlearning} the trigger patterns, thereby causing the decision boundary to resemble that of~an~uncorrupted~model.

\subsubsection*{Challenges}
Employing machine unlearning as a means to eliminate backdoor patterns introduces several challenges.
\begin{itemize}
    \item[\textbf{C1.}] First, it suffers from the issue of \textit{catastrophic forgetting}\cite{DBLP:journals/network/WuGWHZD22,DBLP:journals/corr/abs-2207-05521,DBLP:journals/corr/abs-2003-10933}, a phenomenon in which features not intended for unlearning may unintentionally be removed, thereby adversely impacting the overall performance of the model. 
    \item[\textbf{C2.}] Second, the unlearning process can lead to the generation of models that substantially deviate from the global model, inadvertently increasing the likelihood of the central server detecting the presence of malicious activity. 
    \item[\textbf{C3.}] Third, the parameters of any model possess varying degrees of importance with respect to the outcome of that model. Modifying all parameters uniformly can yield suboptimal results, as this approach fails to account for the differential significance of individual~parameters. 
\end{itemize}

Appropriately addressing these challenges is essential for developing and implementing effective machine unlearning strategies to remove the impact of backdoor patterns.

\subsection{Backdoor Removal using Machine Unlearning}
The basic principle of backdoor removal using machine unlearning is derived from the following observation about \textit{gradient descent} based neural network learning. Given a neural network model $\mathcal{F}$ whose learning objective is $\mathscr{L}$, its learnable parameters $\theta_i$ at the $i$-th training iteration are updated as follows:
\begin{equation}
    \theta_{i+1} \leftarrow \theta_i - \frac{\partial\mathscr{L}}{\partial\theta_i}
\end{equation}
where $\frac{\partial\mathscr{L}}{\partial\theta_i}$ represents the gradient of the model update. Correspondingly, the reversed learning process, i.e., \textit{gradient ascent}, can be expressed as follows:
\begin{equation}
    \theta_{i} \leftarrow \theta_{i+1} + \frac{\partial\mathscr{L}}{\partial\theta_i}
\end{equation}

Thus the loss of trigger pattern unlearning $\mathscr{L}_{\mathcal{U}}$ can be written as follows.
\begin{equation}\label{eq:basic_unlearning}
    \mathscr{L}_{\mathcal{U}} = - \mathscr{L}_{CE}(\mathcal{F}_{\theta_i} (x_b), y_b)) 
\end{equation}
where $\mathscr{L}_{CE}$ is the \textit{cross-entropy} loss function, $x_b$ is the desired trigger pattern required to be unlearned, and $y_b$ is the target backdoor label. 

In the context of an FL framework for removing the trace of backdoor insertions in the global model, the adversary $\mathcal{A}$ aims to eliminate the backdoor patterns whenever the central server selects a compromised participant $p_c$ during a specific training round $r$. In order to achieve this goal, $\mathcal{A}$ initializes the local model $\mathcal{L}^{(r)}_{p_c}$ with the the received global model $\mathcal{G}^{(r)}$ and utilizes Equation~(\ref{eq:basic_unlearning}) on the local model training to unlearn the backdoor behavior. Essentially, the unlearning loss for the adversary to remove backdoor patterns and produce the modified local model $\hat{\mathcal{L}}^{(r+1)}_{p_c}$ can be described as follows:
\begin{equation}\label{eq:unlearning}
    \mathscr{L}_{\mathcal{U}} = - \mathscr{L}_{CE}(\mathcal{L}^{(r)}_{p_c} (x_m), t)) 
\end{equation}
where $x_m \in \mathcal{D}_t$ (the trigger dataset) and $t$ is the desired incorrect label.

However, the loss depicted in Equation~(\ref{eq:unlearning}) may introduce the notable challenge of \textit{catastrophic forgetting}, as discussed in the challenge \textbf{C1}, which could result in considerable performance degradation for model $\hat{\mathcal{L}}^{(r+1)}_{p_c}$ in comparison to other benign participants. Such an occurrence may raise suspicions at the central server concerning malicious activities conducted by the compromised participant. In order to address the concern, we adopt two strategies: \textbf{(1)} \textit{memory preservation:} utilizing benign dataset of the compromised participant $\mathcal{D}_{p_c}$ for preserving the memory of the global model $\mathcal{G}^{(r)}$ over $\mathcal{D}_{p_c}$, and \textbf{(2)} \textit{dynamic penalization:} implementing a dynamic penalty mechanism designed to penalize over-unlearning of the information not associated with trigger patterns in the trigger dataset $\mathcal{D}_{t}$.

\subsubsection{Memory Preservation}
In order to implement the memory preservation strategy, we propose a modification to the loss function $\mathscr{L}_{\mathcal{U}}$ as mentioned in Equation~(\ref{eq:unlearning}). Specifically, we augment $\mathscr{L}_{\mathcal{U}}$ with both the loss on benign data $\mathcal{D}_{p_c}$ of the compromised participant and the loss on backdoor data $\mathcal{D}_{t}$. The loss on benign data is used to preserve the model's memory of the participant's non-malicious behavior. At the same time, the loss of backdoor data enables the model to unlearn the trigger patterns associated with malicious activity. By jointly optimizing these two types of losses, we can ensure that the model $\hat{\mathcal{L}}^{(r+1)}_{p_c}$ retains its knowledge of benign data, even as it unlearns the malicious behavior, thereby assisting in addressing the challenge of catastrophic forgetting. The loss function incorporating the memory preservation strategy can be written as follows:
\begin{equation}\label{eq:unlearning_mp}
    \mathscr{L}_{\mathcal{U}} = \mathscr{L}_{CE}(\mathcal{L}^{(r)}_{p_c} (x_b), y_b)) - \mathscr{L}_{CE}(\mathcal{L}^{(r)}_{p_c} (x_m), t))
\end{equation}
where $x_b \in \mathcal{D}_{p_c}$ is the benign data and $y_b$ is the corresponding benign label. 

\subsubsection{Dynamic Penalization}\label{sec:dynamic_penalization}
In order to implement the dynamic penalization strategy, we propose a method that imposes a penalty on the updated weights of the local model $\mathcal{L}^{(r)}_{p_c}$ in proportion to the extent of their variation relative to the $L_1$-norm of the weights of the global model $\mathcal{G}^{(r)}$. We augment the penalty with the loss function outlined in Equation~(\ref{eq:unlearning_mp}), with the objective of confining the model parameters of $\mathcal{L}^{(r)}_{p_c}$ to prevent substantial deviation from the model parameters of $\mathcal{G}^{(r)}$. By optimizing the loss function along with this penalty, we ensure that $\mathcal{L}^{(r)}_{p_c}$ remains in close proximity to $\mathcal{G}^{(r)}$ while eliminating any backdoor pattern. The strategy assists in addressing the challenge of catastrophic forgetting and also avoids producing suspicious model updates that could raise alerts at the central server, as discussed in the challenge \textbf{C2}. The modified loss function incorporating the dynamic penalization strategy can be written as follows:
\begin{equation}\label{eq:unlearn_penalty}
    \begin{split}
        \mathscr{L}_{\mathcal{U}} = \mathscr{L}_{CE}(\mathcal{L}^{(r)}_{p_c} (x_b), y_b)) & - \mathscr{L}_{CE}(\mathcal{L}^{(r)}_{p_c} (x_m), t)) \\ & + \gamma \cdot \|\theta_{p_c} - \theta_{\mathcal{G}}\|_{1}
    \end{split}
\end{equation}
where $\gamma$ represents a hyperparameter that modulates the influence of the penalty term on the loss function, $\|\cdot\|_1$ denotes the $L_1$-norm, $\theta_{p_c}$ and $\theta_{\mathcal{G}}$ are the model parameters of $\mathcal{L}^{(r)}_{p_c}$ and $\mathcal{G}^{(r)}$, respectively.

Within a neural network model, it is crucial to recognize that individual parameters possess varying degrees of influence on the classification outcome. Imposing uniform penalties on all parameters may not produce optimal results, as this approach neglects the diverse importance of individual parameters, as discussed in challenge \textbf{C3}. In order to address this challenge, we propose integrating weights into the penalty term, attributing a distinct weight to each parameter according to its importance. The importance is derived from two components: one that emphasizes benign classification (denoted as $\mathcal{I}^{(b)}_{\theta}$), while the other concentrates on backdoor patterns (denoted as $\mathcal{I}^{(m)}_{\theta}$). Consequently, the importance is measured using the ratio $\frac{\mathcal{I}^{(b)}_{\theta}}{\mathcal{I}^{(m)}_{\theta}}$. By adopting this approach, parameters exhibiting higher $\mathcal{I}^{(b)}_{\theta}$ and lower $\mathcal{I}^{(m)}_{\theta}$ values will be subjected to increased penalties, ensuring minimal deviation during training iterations, as these parameters are highly significant for benign classification and less so for backdoor classification. On the other hand, parameters with lower $\mathcal{I}^{(b)}_{\theta}$ and higher $\mathcal{I}^{(m)}_{\theta}$ values will encounter more relaxed constraints, as these parameters are less significant for benign classification and highly significant for backdoor classification. Modulating these parameters accordingly will lead to improved results in terms of maintaining accuracy for benign classification while simultaneously unlearning backdoor patterns. The computation of $\mathcal{I}^{(b)}_{\theta}$ and $\mathcal{I}^{(m)}_{\theta}$ can be written as follows:
\begin{equation}\label{eq:importance}
    \begin{split}
        \mathcal{I}^{(b)}_{\theta} = \frac{1}{|\mathcal{D}_{p_c}|}\left|\frac{\partial\mathscr{L}_{CE}(\mathcal{L}^{(r)}_{p_c} (x_b), y_b))}{\partial\theta_{p_c}}\right|\\
        \mathcal{I}^{(m)}_{\theta} = \frac{1}{|\mathcal{D}_{t}|}\left|\frac{\partial\mathscr{L}_{CE}(\mathcal{L}^{(r)}_{p_c} (x_m), t))}{\partial\theta_{p_c}}\right|
    \end{split}
\end{equation}

The modified loss function outlined in Equation~(\ref{eq:unlearn_penalty}) incorporating weighted importance can be written as follows:

\begin{equation}\label{eq:final_unlearn}
    \begin{split}
        \mathscr{L}_{\mathcal{U}} = \mathscr{L}_{CE}(\mathcal{L}^{(r)}_{p_c} (x_b), y_b)) & - \mathscr{L}_{CE}(\mathcal{L}^{(r)}_{p_c} (x_m), t)) \\ & + \gamma \cdot \|\omega \cdot (\theta_{p_c} - \theta_{\mathcal{G}})\|_{1}
    \end{split}
\end{equation}
where $\omega$ is defined as $\frac{\mathcal{I}^{(b)}_{\theta}}{\mathcal{I}^{(m)}_{\theta}}$, used for the dot product with the parameter deviation before computing the $L_1$-norm for the penalty term. We use the final loss function defined in Equation~(\ref{eq:final_unlearn}) to remove backdoor patterns from the local model and subsequently from the global model. We outline the complete backdoor removal procedure for the compromised participant in Algorithm~\ref{algo:unlearn}.
\begin{algorithm}[!t]
\caption{Backdoor Removal Using Machine Unlearning by the Compromised Participant $p_c$}\label{algo:unlearn}
\KwIn{Global model at the $r$-th Federated Learning round $\mathcal{G}^{(r)}$, benign dataset of the compromised participant $\mathcal{D}_{p_c}$, trigger dataset $\mathcal{D}_{t}$}
\KwOut{Local model $\hat{\mathcal{L}}^{(r+1)}_{p_c}$ having no backdoor patterns}
Initialize local model $\mathcal{L}^{(r)}_{p_c}$ with $\mathcal{G}^{(r)}$ \\
$\theta_{\mathcal{G}} \gets$ extracted parameters of $\mathcal{G}^{(r)}$\\
\For{each local unlearning iteration}{
    Compute $\mathcal{I}^{(b)}_{\theta}$ and $\mathcal{I}^{(m)}_{\theta}$ using Equation~(\ref{eq:importance})\\
    Compute $\omega \gets \frac{\mathcal{I}^{(b)}}{\mathcal{I}^{(m)}_{\theta}}$\\
    $\theta_{p_c} \gets$ extracted parameters of $\mathcal{L}^{(r)}_{p_c}$\\
    Compute weighted penalty term $\|\omega \cdot (\theta_{p_c} - \theta_{\mathcal{G}})\|_{1}$\\
    Compute unlearning loss $\mathscr{L}_{\mathcal{U}}$ using Equation~(\ref{eq:final_unlearn})\\
    Update $\mathcal{L}^{(r)}_{p_c}$ using standard optimization algorithm considering the loss $\mathscr{L}_{\mathcal{U}}$ 
}
$\hat{\mathcal{L}}^{(r+1)}_{p_c} \gets \mathcal{L}^{(r)}_{p_c} - \mathcal{G}^{(r)}$\\
Return $\hat{\mathcal{L}}^{(r+1)}_{p_c}$
\end{algorithm}

\section{Experimental Setup}\label{sec:setup}
In this section, we briefly outline the dataset and training configurations, attack strategies, and evaluation metrics utilized in our study, providing the basis for understanding the experiments and results presented in subsequent sections.

\subsection{Dataset and Training Configurations}
We assess the effectiveness of our proposed method in the context of an image classification task within an FL framework. To this end,  we utilize two widely-adopted datasets, CIFAR-10~\cite{krizhevsky2009learning} and CIFAR-100~\cite{krizhevsky2009learning}, as benchmarks. We employ two standard neural network architectures to serve as classifiers: VGG-11~\cite{DBLP:journals/corr/SimonyanZ14a} for CIFAR-10 and ResNet-18~\cite{DBLP:conf/cvpr/HeZRS16} for CIFAR-100, containing approximately $9.7M$ and $11.1M$ parameters, respectively. We adopt an FL framework wherein participants collaboratively train the global model using training data that is independently and identically distributed among them. We use the deep learning framework \textit{PyTorch}\cite{DBLP:conf/nips/PaszkeGMLBCKLGA19} for all implementations.

Adhering to the FL training configuration presented in~\cite{DBLP:conf/aistats/BagdasaryanVHES20,DBLP:conf/icml/ZhangPSYMMR022}, we train the global model with a total of $n=100$ participants, wherein 10 participants are selected randomly in each FL round, i.e., $|\mathcal{S}_m| = 10$. In each round, the chosen participants train their individual local models for 2 epochs with a learning rate of 0.1. The local training employs \textit{stochastic gradient descent} optimization and \textit{negative log-likelihood loss}. The central server utilizes the FedAvg aggregation method as the baseline implementation.

\subsection{Attack Strategies}
In our evaluation, we consider one out of the total $100$ participants to be malicious. The malicious participant is selected based on the three poisoning mechanisms outlined in Section~\ref{sec:poisoning}. We examine two open-sourced state-of-the-art backdoor insertion attacks in FL - \textit{Constraint and Scale}~\cite{DBLP:conf/aistats/BagdasaryanVHES20}, and \textit{Neurotoxin}~\cite{DBLP:conf/icml/ZhangPSYMMR022} - well-established for their ability to embed stealthy backdoors and maintain persistence across multiple FL training rounds. We consider three distinct categories of commonly used \textit{semantic targeted variants} for backdoor triggers: (1) misclassifying all ``green cars'' as ``birds''~\cite{DBLP:conf/aistats/BagdasaryanVHES20,DBLP:conf/nips/WangSRVASLP20,DBLP:conf/icml/ZhangPSYMMR022}, (2) misclassifying a subset of images from class label ``5'' as class label ``9''~\cite{DBLP:conf/icml/ZhangPSYMMR022}, and (3) edge case samples (e.g., misclassifying images of Southwest airplanes as ``truck'' for a CIFAR-10 classifier)~\cite{DBLP:conf/nips/WangSRVASLP20}. The proposed method remains equally applicable to other trigger patterns and target labels. Table~\ref{table:attack_settings} provides an overview of the prevalent attack settings found in the literature, which we have considered in our evaluation.
\begin{table}[!t]
\caption{Overview of the backdoor insertion settings considered in the evaluation.}
\label{table:attack_settings}
\resizebox{\columnwidth}{!}{
\begin{tabular}{c|c|c|c|c}
\hline
\textbf{ID} & \textbf{Dataset} & \textbf{Model} & \textbf{Trigger Pattern} & \textbf{Attack Method} \\ \hline
1 & \multirow{3}{*}{CIFAR-10} & \multirow{3}{*}{VGG-11} & \begin{tabular}[c]{@{}c@{}}Green Cars\\ to Bird\end{tabular} & \begin{tabular}[c]{@{}c@{}}Constraint\\ and Scale~\cite{DBLP:conf/aistats/BagdasaryanVHES20}\end{tabular} \\ \cline{1-1} \cline{4-5} 
2 &  &  & \begin{tabular}[c]{@{}c@{}}Label 5\\ to Label 9\end{tabular} & Neurotoxin~\cite{DBLP:conf/icml/ZhangPSYMMR022} \\ \cline{1-1} \cline{4-5} 
3 &  &  & \begin{tabular}[c]{@{}c@{}}Edge\\ Case\end{tabular} & Neurotoxin~\cite{DBLP:conf/icml/ZhangPSYMMR022} \\ \hline
4 & CIFAR-100 & ResNet-18 & \begin{tabular}[c]{@{}c@{}}Label 5\\ to Label 9\end{tabular} & Neurotoxin~\cite{DBLP:conf/icml/ZhangPSYMMR022} \\ \hline
\end{tabular}}
\end{table}

\subsection{Evaluation Metrics}\label{sec:eval_metric}
In order to assess the effectiveness of the proposed method, we consider the following two metrics: \textbf{(1)} $Acc_{\mathcal{B}}$ (\textit{Backdoor Accuracy}), which represents the accuracy of a model in the backdoor task, quantifying the proportion of backdoor images for which the model generates target outputs as determined by the adversary, and \textbf{(2)} $Acc_{\mathcal{M}}$ (\textit{Main Task Accuracy}), which represents the accuracy of a model in its primary task, quantifying the proportion of benign images for which the model produces accurate predictions. During backdoor insertion, the adversary aims to maximize $Acc_{\mathcal{B}}$ while minimizing its impact on $Acc_{\mathcal{M}}$. In contrast, when removing backdoors, the aim is to minimize $Acc_{\mathcal{B}}$ while minimizing its impact on $Acc_{\mathcal{M}}$. Additionally, we utilize the $L_2$-norm as a measure of the deviation of model parameters from the global model, serving as an indicator of stealthiness.

\section{Experimental Results}\label{sec:results}
In this section, we present a detailed analysis of the experimental results for the proposed backdoor removal method using machine unlearning. The analysis also includes an in-depth ablation study, which evaluates the impact of various factors on the performance and efficacy of the proposed approach.

\subsection{Effectiveness of Backdoor Unlearning}
\begin{figure*}[!t]
     \centering
     \begin{subfigure}[t]{0.24\linewidth}
         \centering
         \includegraphics[width=\linewidth]{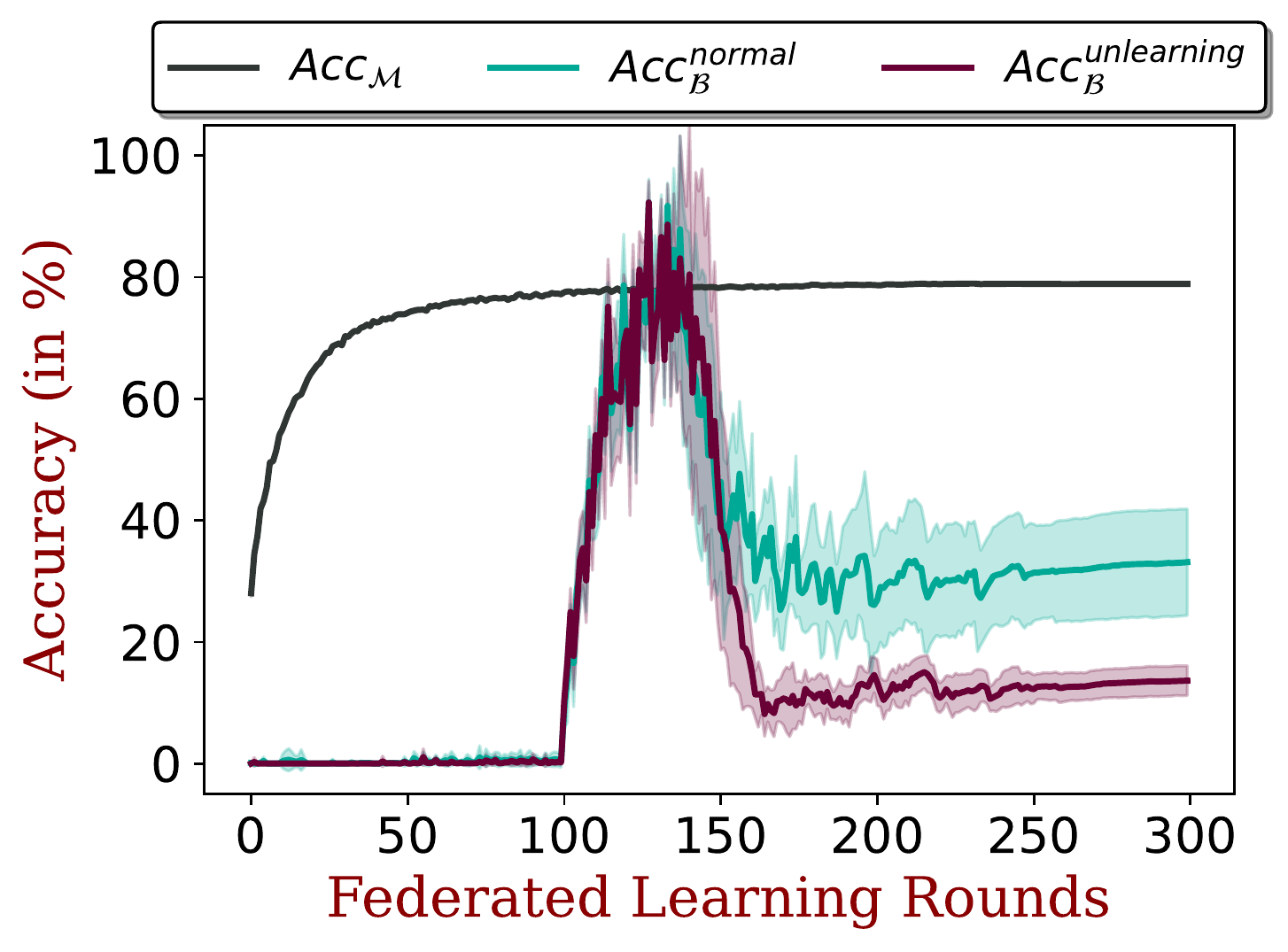}
         \caption{\textbf{ID1:} Continuous Selection}
     \end{subfigure}
     \begin{subfigure}[t]{0.24\linewidth}
         \centering
         \includegraphics[width=\linewidth]{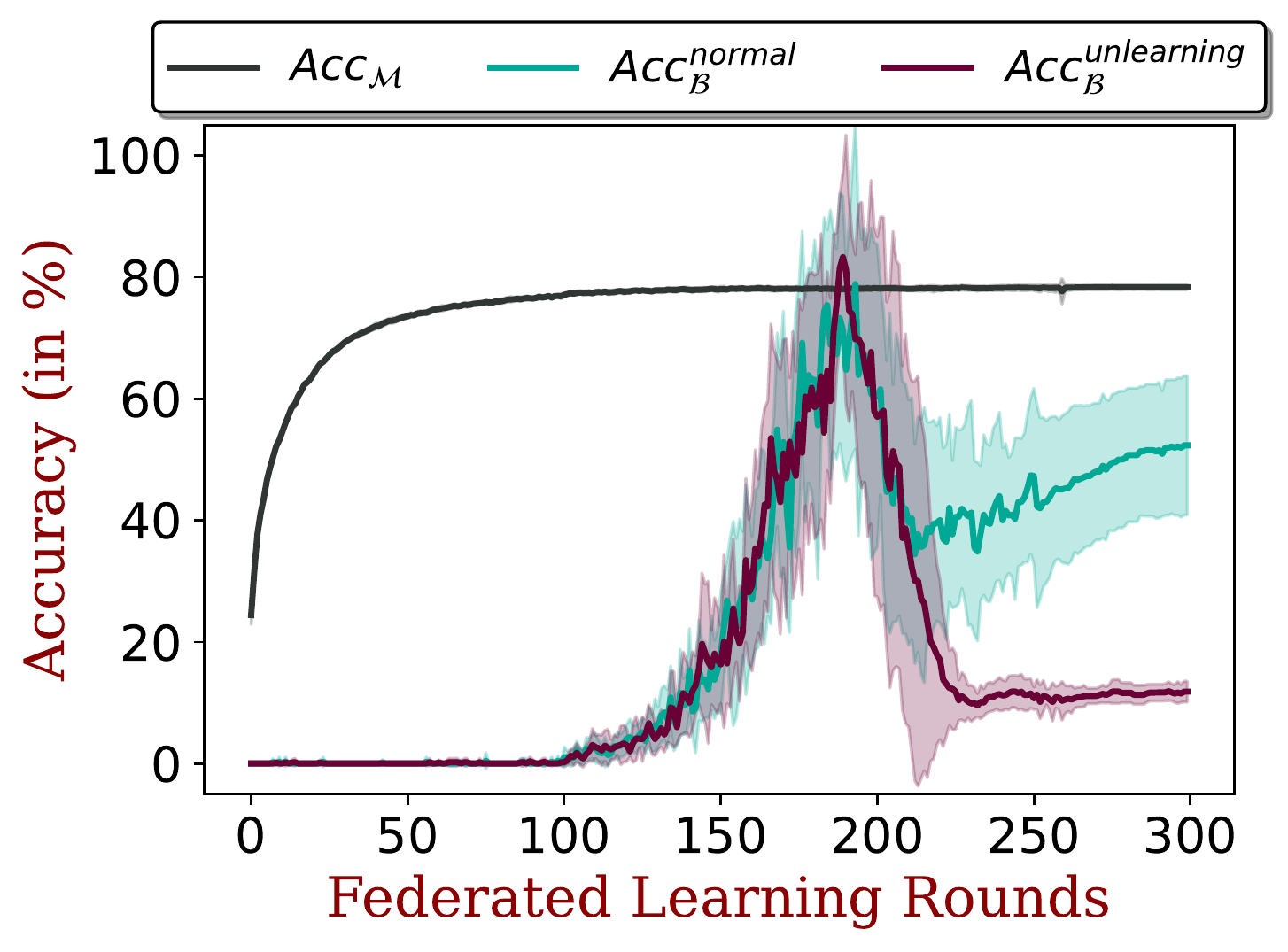}
         \caption{\textbf{ID2:} Continuous Selection}
     \end{subfigure}
     \begin{subfigure}[t]{0.24\linewidth}
         \centering
         \includegraphics[width=\linewidth]{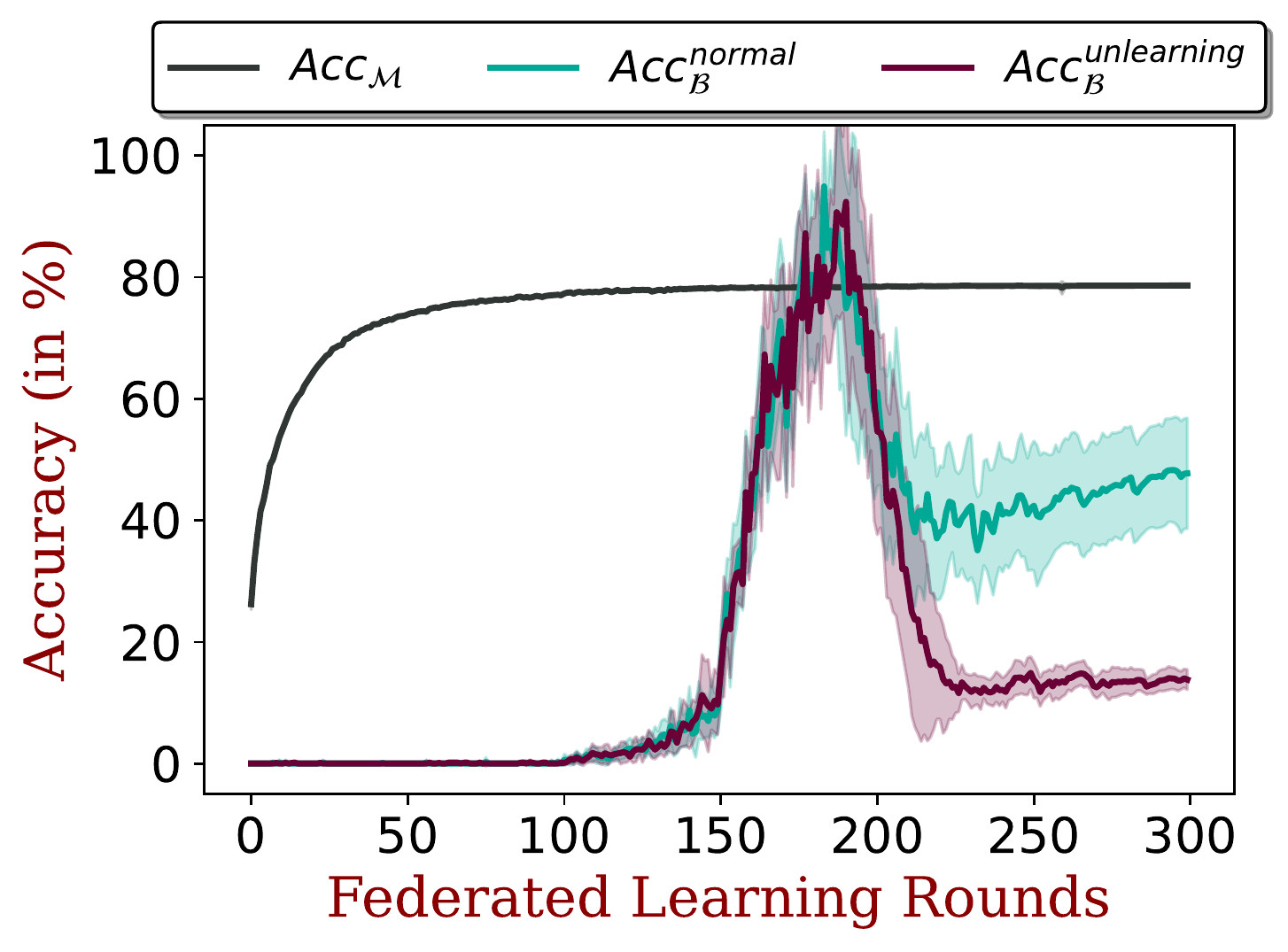}
         \caption{\textbf{ID3:} Continuous Selection}
     \end{subfigure}
     \begin{subfigure}[t]{0.24\linewidth}
         \centering
         \includegraphics[width=\linewidth]{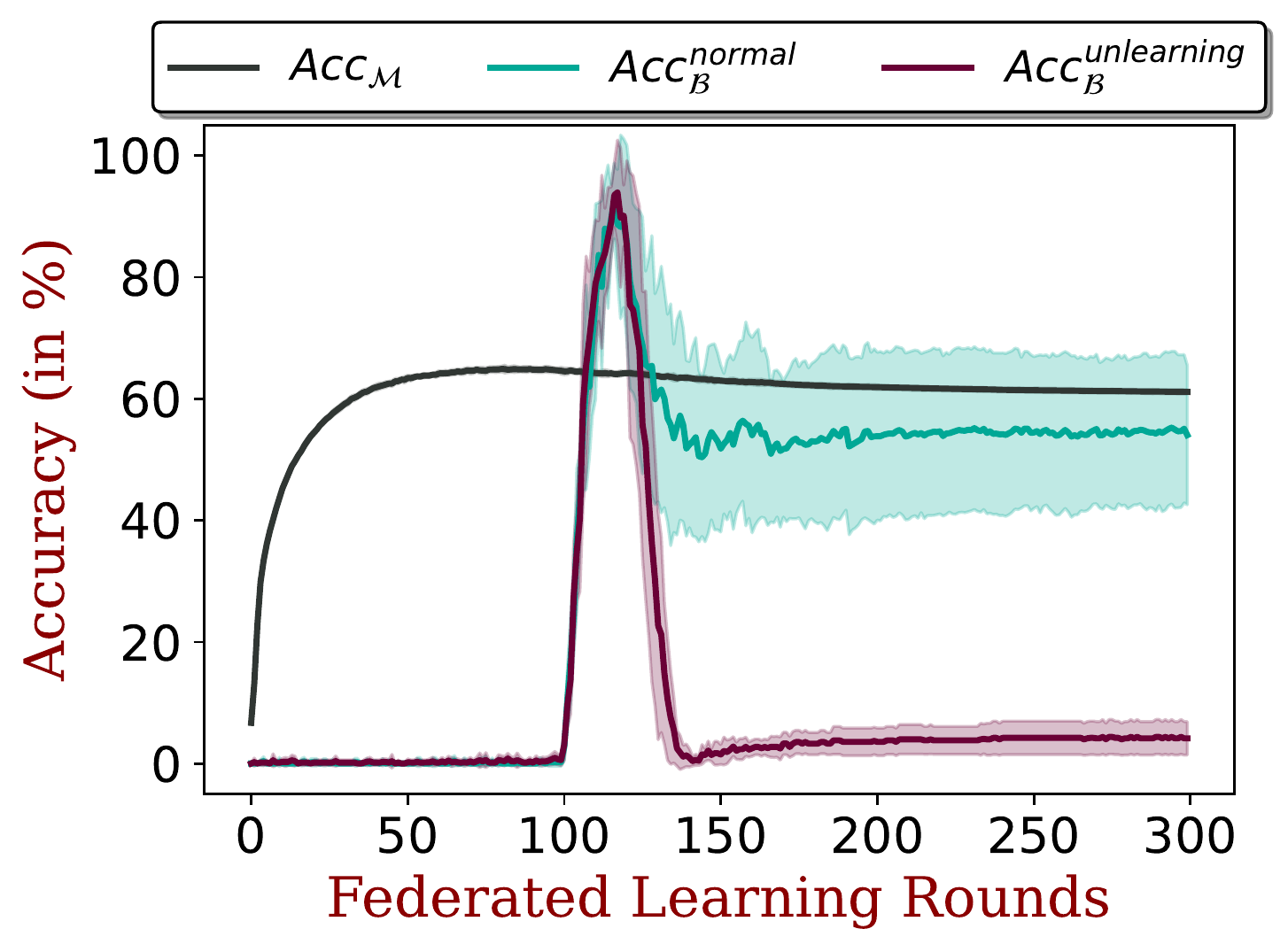}
         \caption{\textbf{ID4:} Continuous Selection}
     \end{subfigure}
     \begin{subfigure}[t]{0.24\linewidth}
         \centering
         \includegraphics[width=\linewidth]{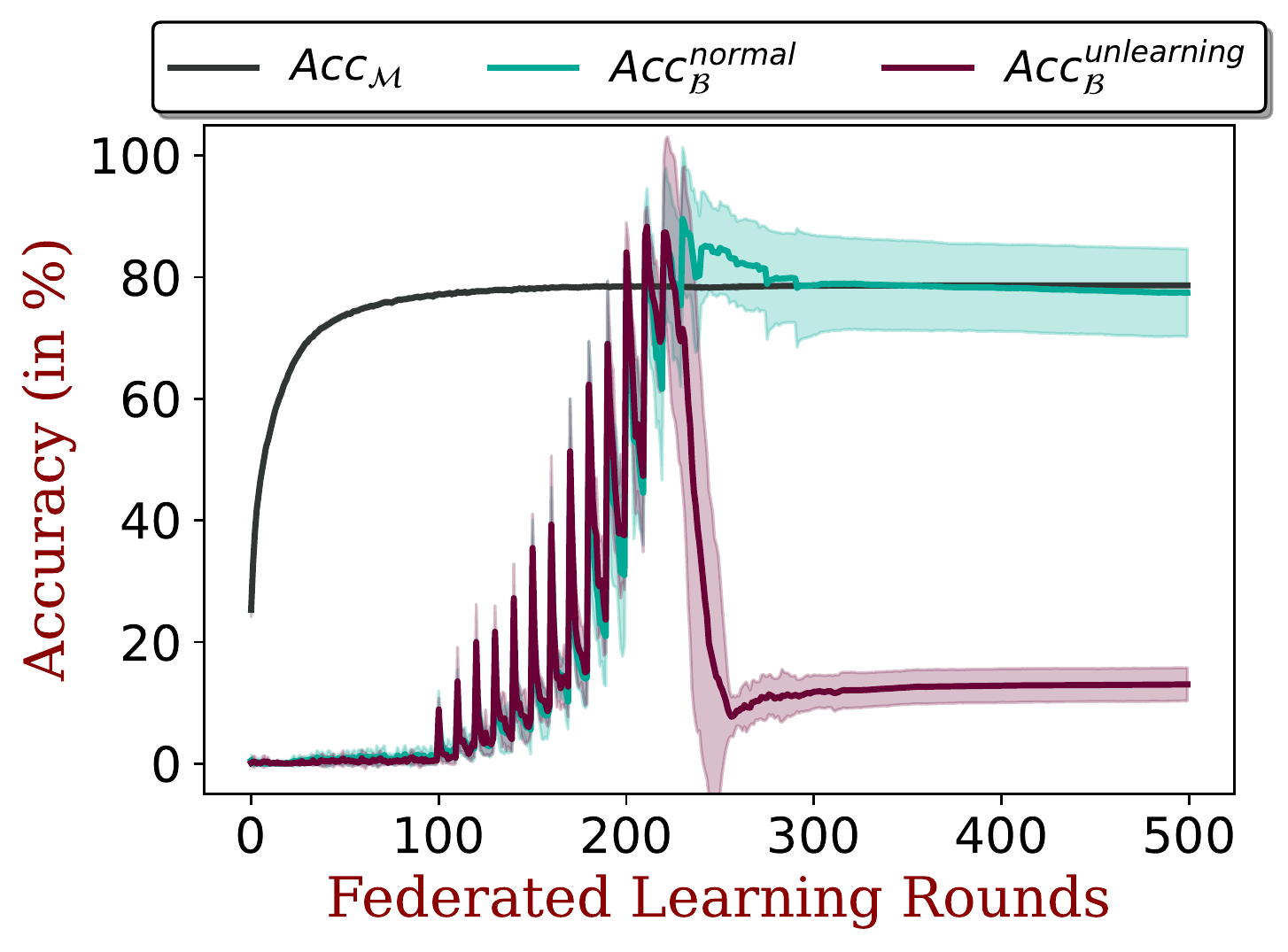}
         \caption{\textbf{ID1:} Fixed-Freq. Selection}
     \end{subfigure}
     \begin{subfigure}[t]{0.24\linewidth}
         \centering
         \includegraphics[width=\linewidth]{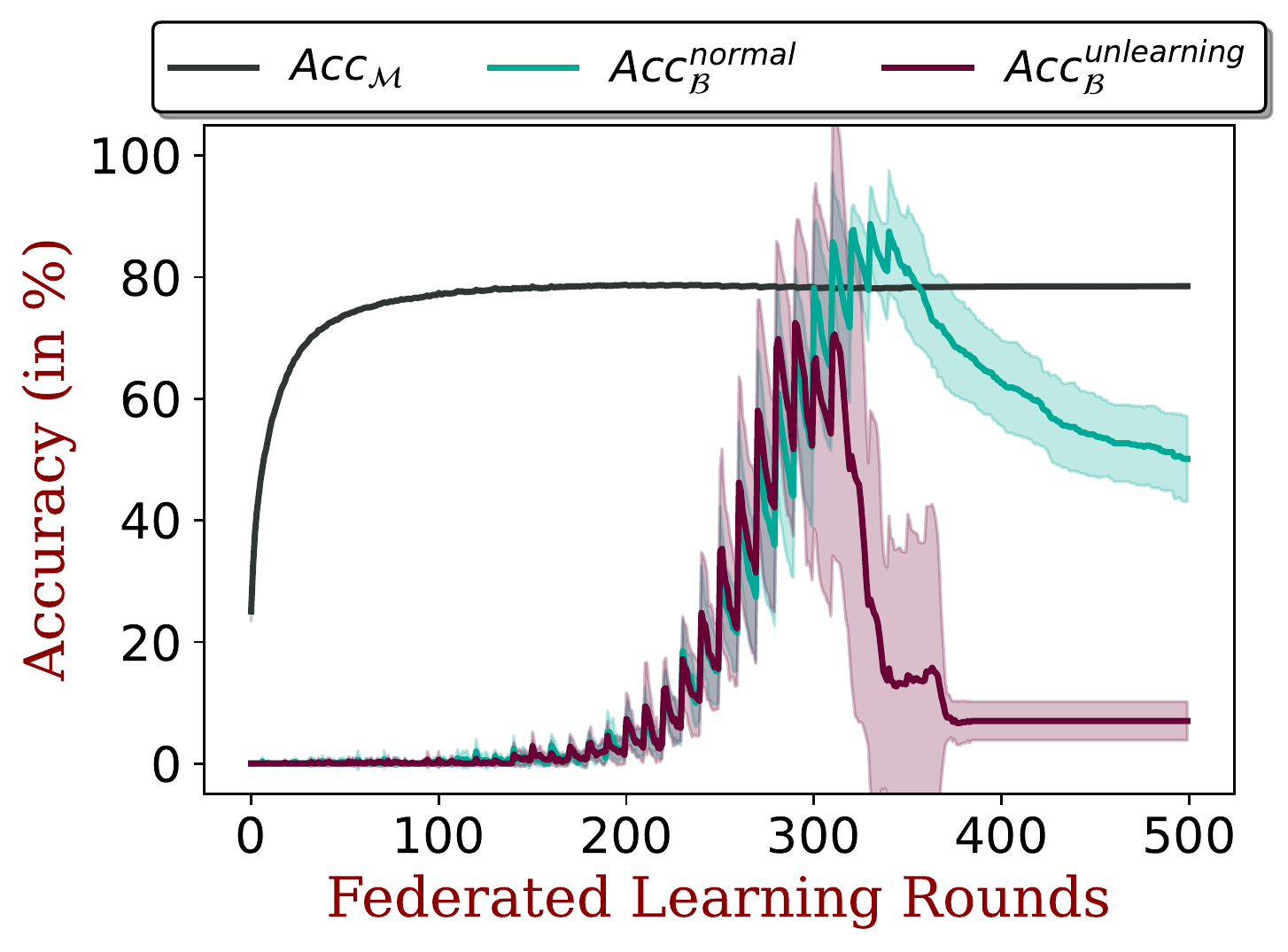}
         \caption{\textbf{ID2:} Fixed-Freq. Selection}
     \end{subfigure}
     \begin{subfigure}[t]{0.24\linewidth}
         \centering
         \includegraphics[width=\linewidth]{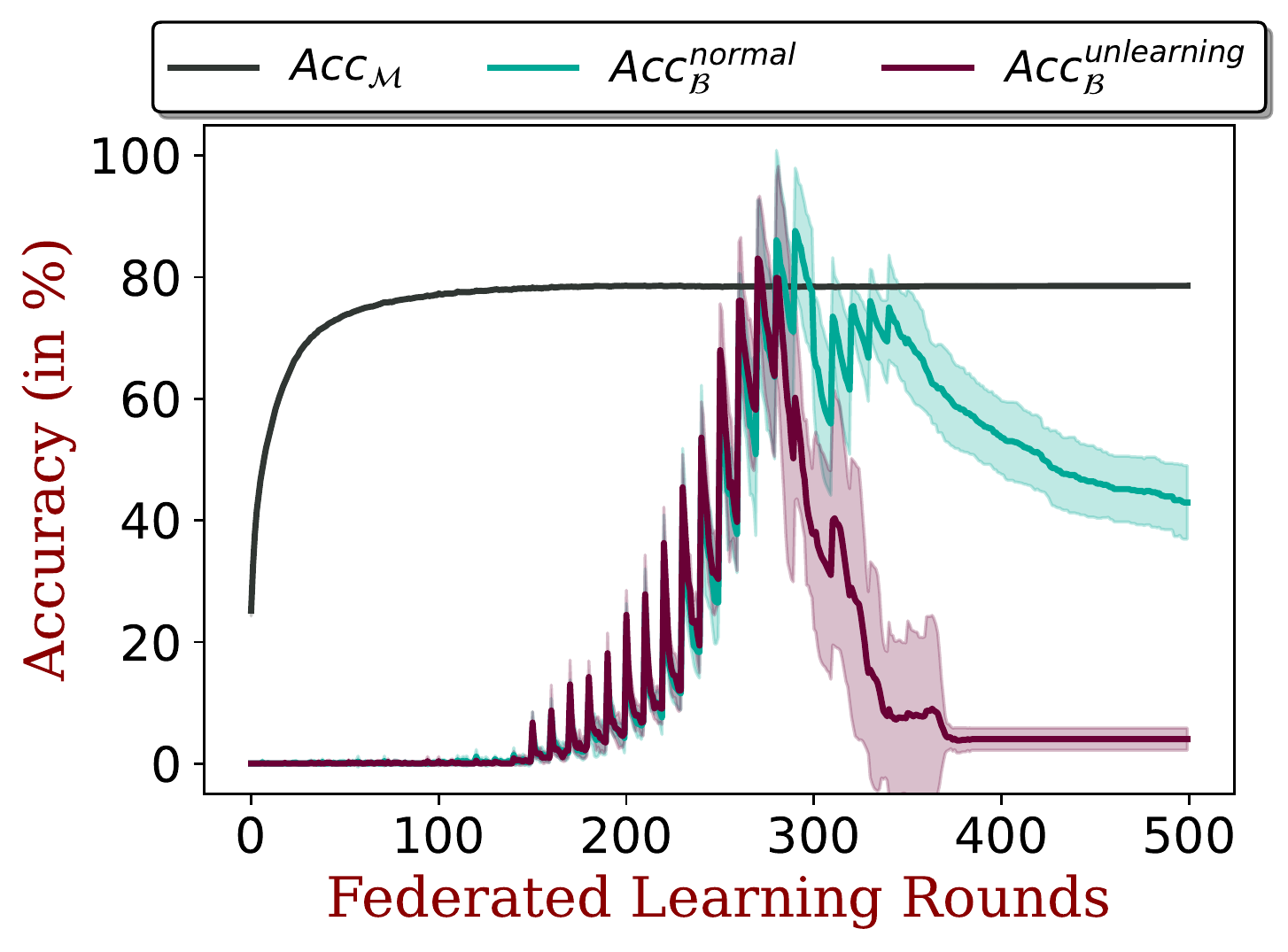}
         \caption{\textbf{ID3:} Fixed-Freq. Selection}
     \end{subfigure}
     \begin{subfigure}[t]{0.24\linewidth}
         \centering
         \includegraphics[width=\linewidth]{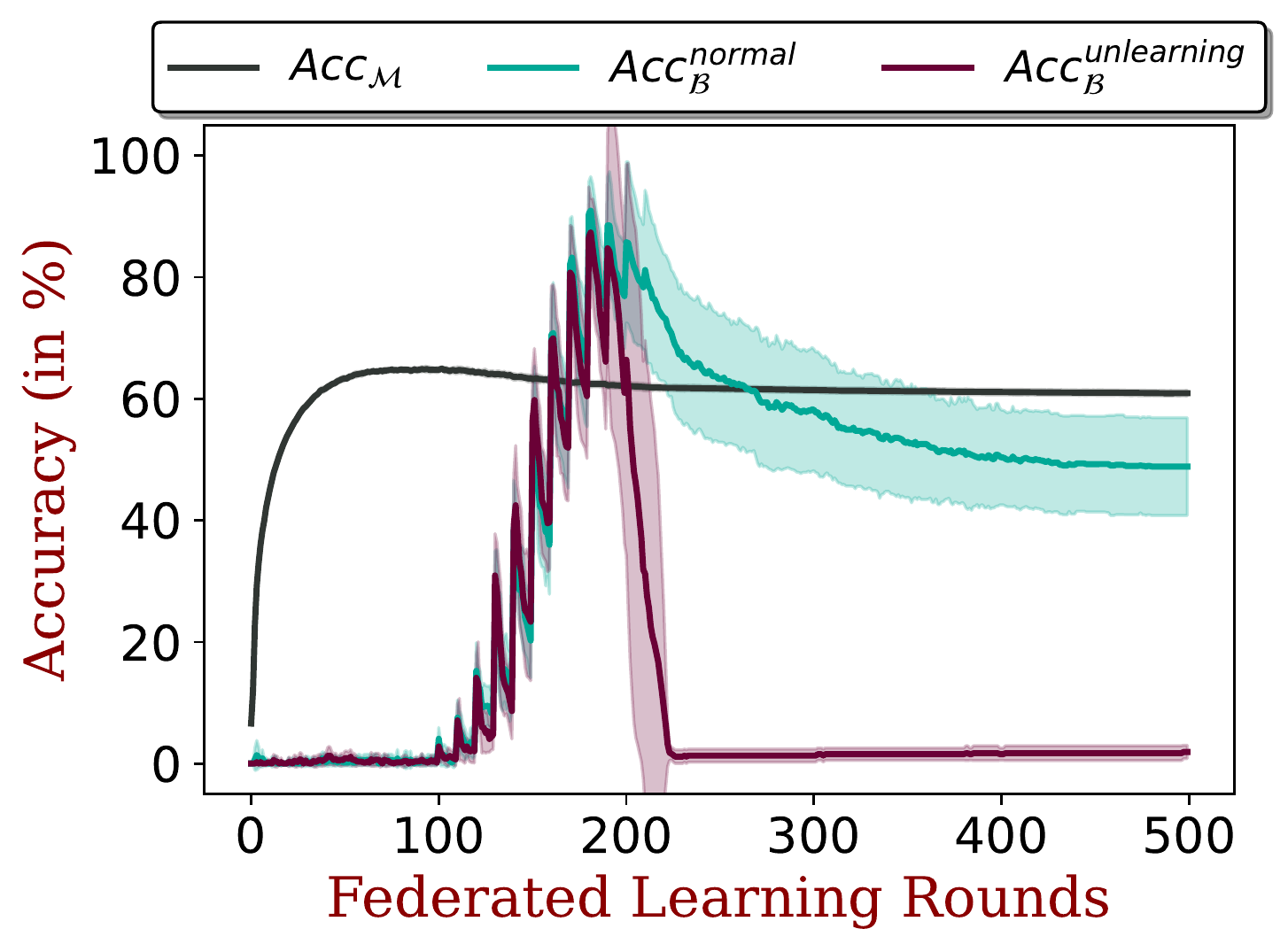}
         \caption{\textbf{ID4:} Fixed-Freq. Selection}
     \end{subfigure}
     \begin{subfigure}[t]{0.24\linewidth}
         \centering
         \includegraphics[width=\linewidth]{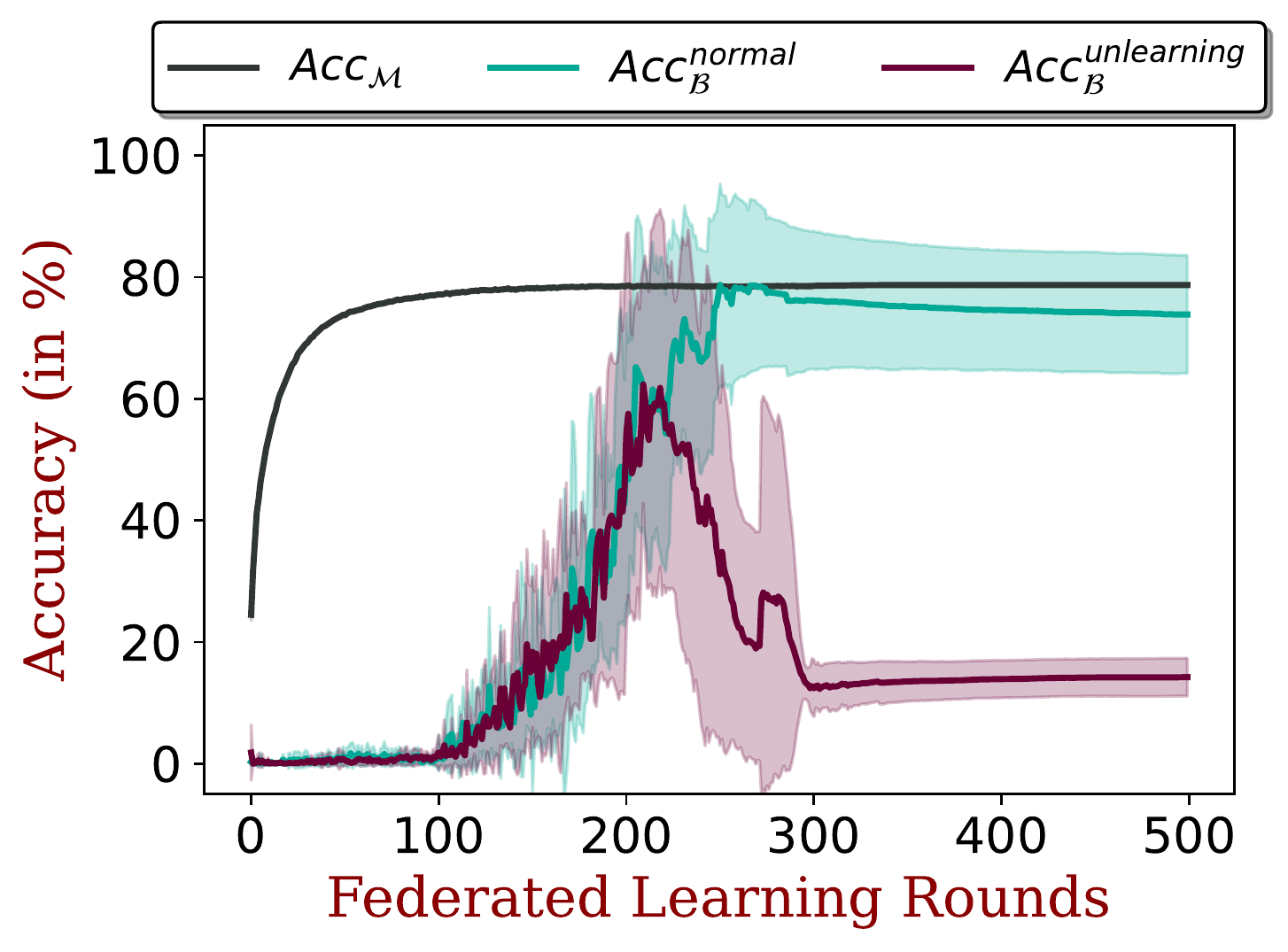}
         \caption{\textbf{ID1:} Random Selection}
     \end{subfigure}
     \begin{subfigure}[t]{0.24\linewidth}
         \centering
         \includegraphics[width=\linewidth]{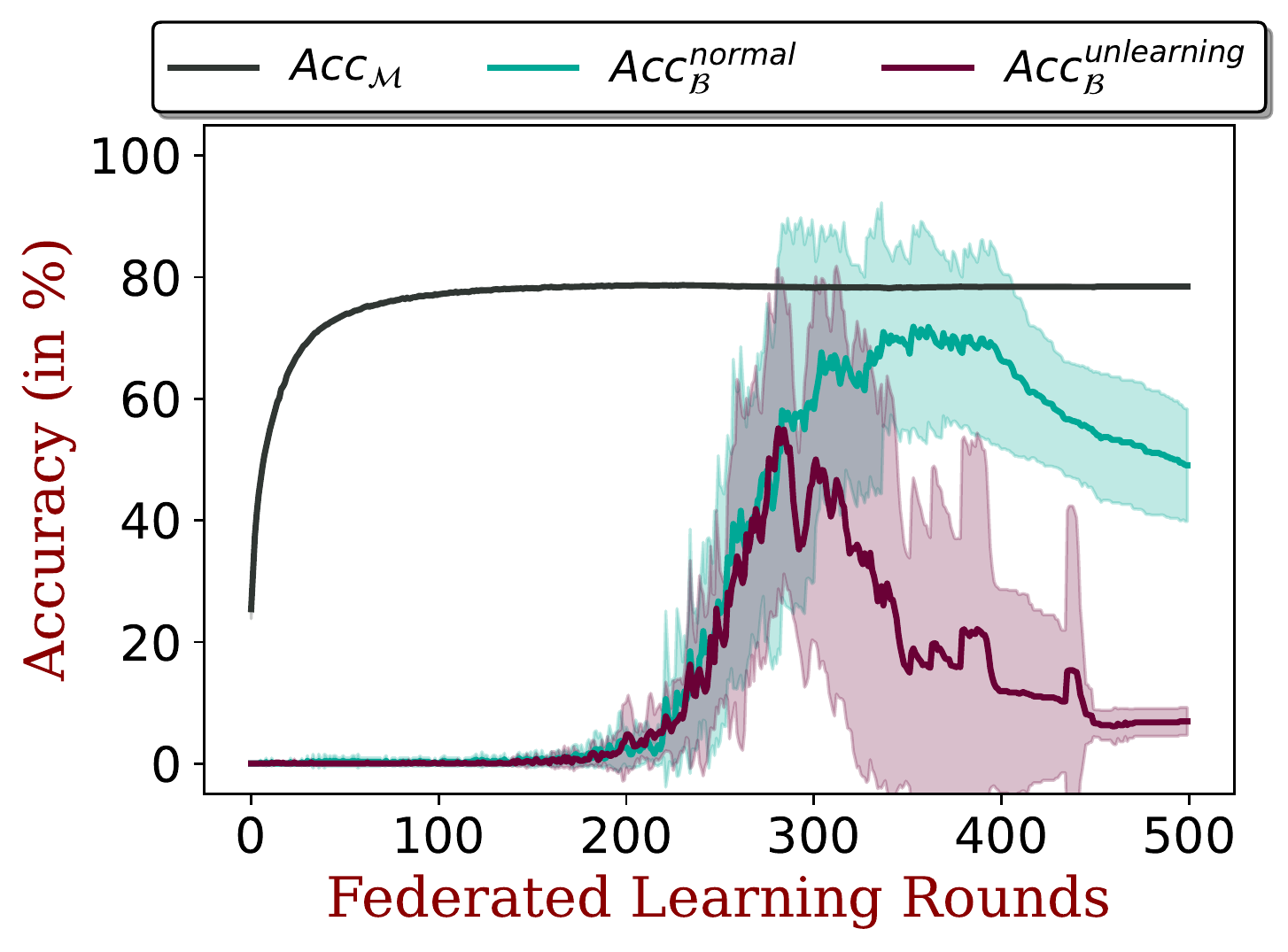}
         \caption{\textbf{ID2:} Random Selection}
     \end{subfigure}
     \begin{subfigure}[t]{0.24\linewidth}
         \centering
         \includegraphics[width=\linewidth]{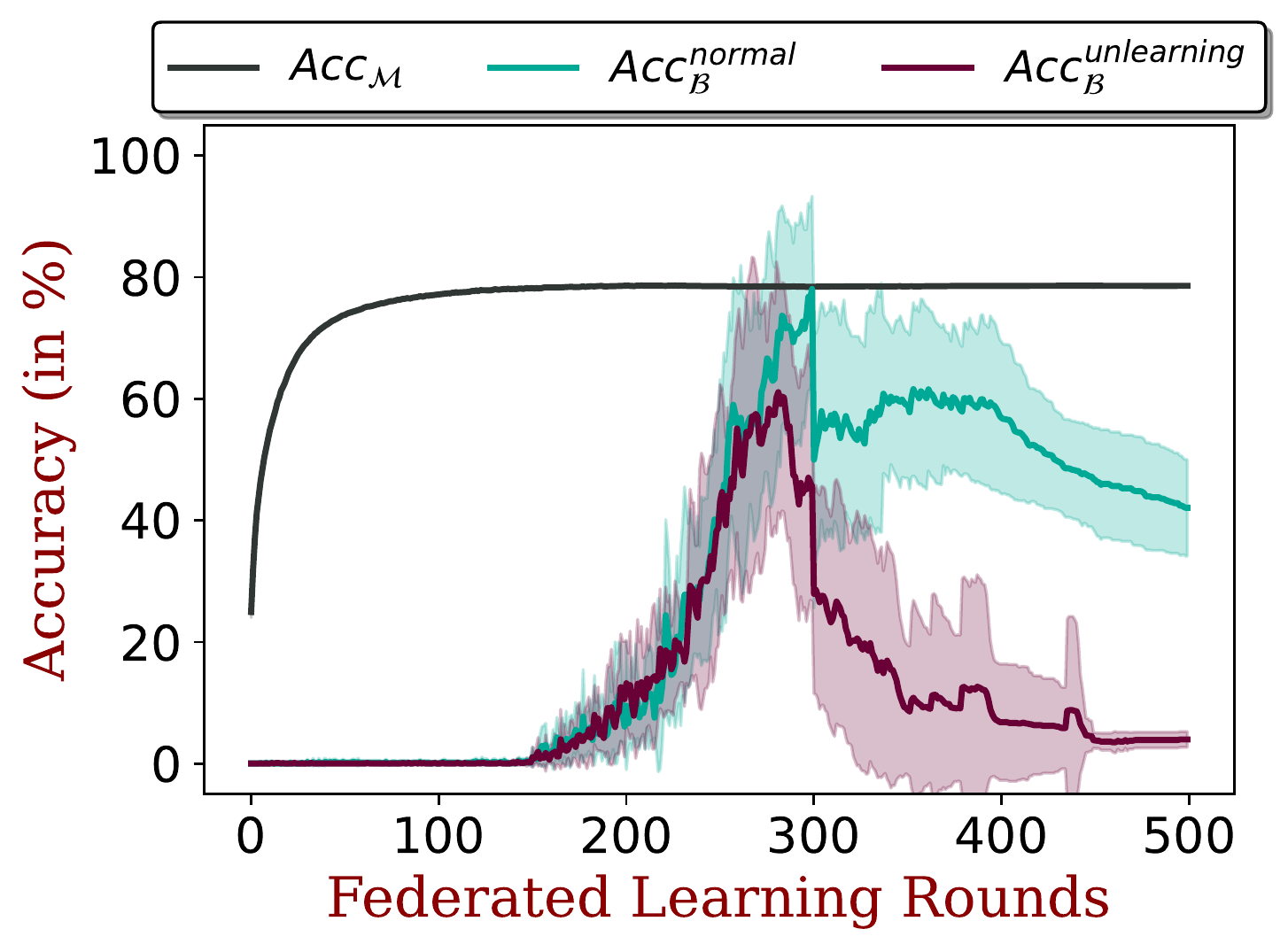}
         \caption{\textbf{ID3:} Random Selection}
     \end{subfigure}
     \begin{subfigure}[t]{0.24\linewidth}
         \centering
         \includegraphics[width=\linewidth]{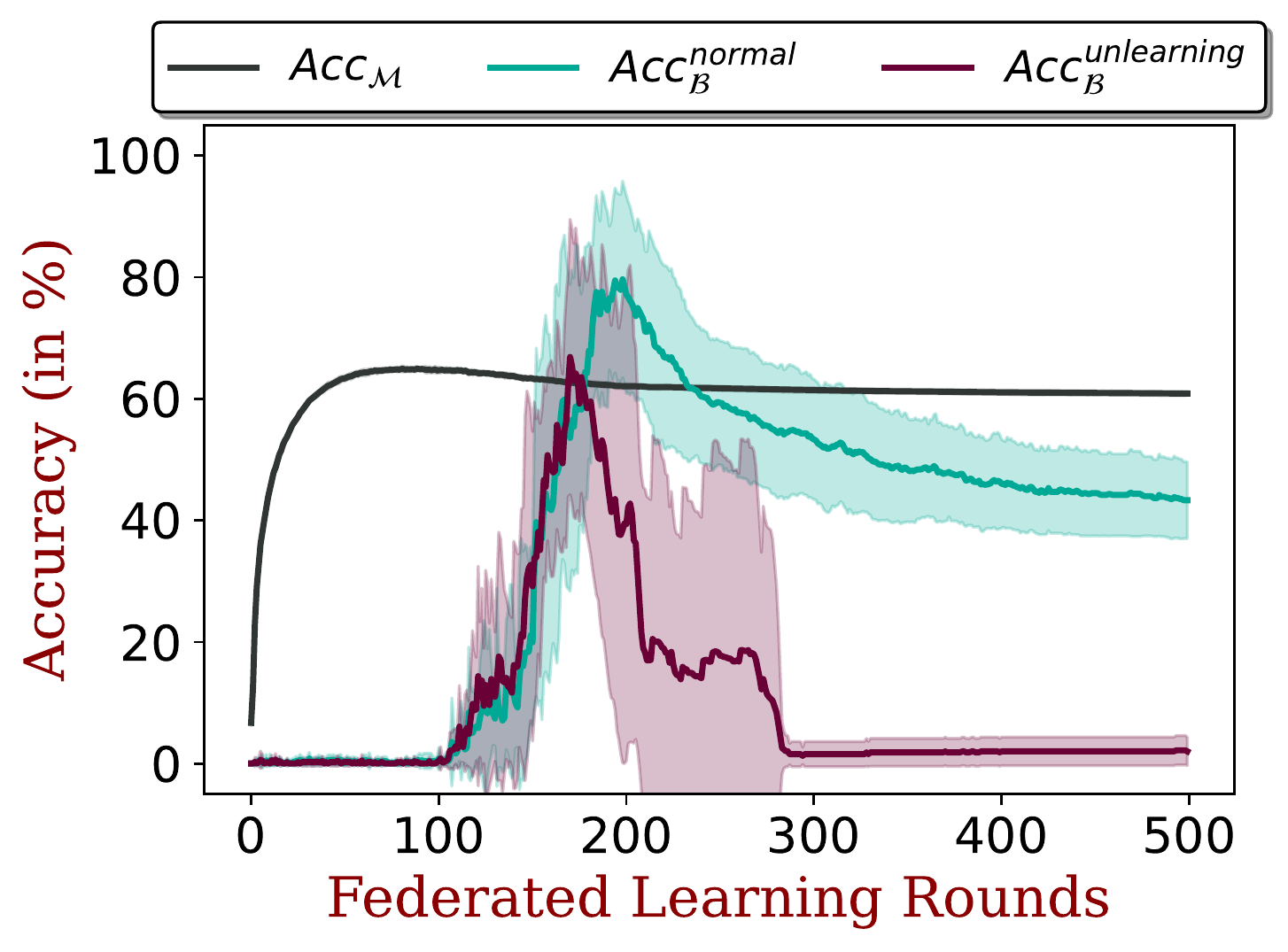}
         \caption{\textbf{ID4:} Random Selection}
     \end{subfigure}
\caption{Performance evaluation of the proposed backdoor removal approach considering default backdoor insertion settings mentioned in Table~\ref{table:attack_settings} and different poisoning strategies. It shows the mean and standard deviation of ten independent runs.}\label{fig:unlearn_performance}
\vspace{-0.085cm}
\end{figure*}

We evaluate the performance of the proposed backdoor removal approach within the context of FL, taking into account the backdoor insertion techniques outlined in Table~\ref{table:attack_settings}. We adhere to the default attack configurations specified in the respective papers to ensure a fair evaluation, including hyperparameters, the number of backdoor samples, and other relevant settings. We also consider the three poisoning strategies - continuous selection, fixed-frequency selection, and random selection - for this evaluation. We consider that the adversary starts backdoor insertion into the global model once it stabilizes from the initial learning fluctuations, as inserting backdoors at the initial rounds is unlikely to persist due to the dynamics of FL training iterations. We observe that the global model stabilizes roughly after 100 FL training rounds for both the datasets. So for each of the configurations mentioned in Table~\ref{table:attack_settings}, we initiate backdoor insertion starting from round 100. We continue the process until the backdoor accuracy in the global model reaches 90\%. Without loss of generality, we begin removing backdoors after 5 FL rounds when the backdoor accuracy attains 90\%. We proceed with backdoor removal until the backdoor patterns reach random guess accuracy, i.e., 10\% for CIFAR-10 and 1\% for CIFAR-100. We consider 6 local epochs for the unlearning step with an initial learning rate of $10^{-5}$. The learning rate is subsequently reduced by a factor of $10$ after every 2 epochs. We use the \textit{stochastic gradient descent} algorithm for optimizing the loss mentioned in Equation(~\ref{eq:final_unlearn}). We consider $\gamma=3$ for this analysis. We also provide an ablation study on the choice of $\gamma$ and its impact on the results later in Section~\ref{sec:ablation_gamma}. In this analysis, we assume that the adversary employs a continuous poisoning strategy only during the backdoor removal phase, with the aim of rapidly eliminating the traces of backdoors from the global model, irrespective of the strategy implemented in the backdoor insertion phase. We later conduct an ablation study to analyze the impact of a non-continuous poisoning strategy for backdoor removal in Section~\ref{sec:ablation_non_continuous}. The results of unlearning the backdoor patterns using the aforementioned configurations for all the backdoor insertion methods are presented in Figure~\ref{fig:unlearn_performance}. Without loss of generality the plots are shown till 300 rounds for continuous selection and 500 rounds for both fixed-frequency and random selection.

The plots displayed in Figure~\ref{fig:unlearn_performance} are generated by considering ten distinct, independent runs with varying random seeds. The solid lines represent the mean values of these runs, while the shaded regions surrounding the lines illustrate the corresponding standard deviations. In the figure, $Acc_{\mathcal{M}}$ represents the main task accuracy. $Acc^{normal}_{\mathcal{B}}$ refers to the backdoor accuracy in cases where no backdoor unlearning is applied, and the adversary stops contributing to FL training after backdoor insertion is completed. $Acc^{unlearning}_{\mathcal{B}}$, on the other hand, represents the backdoor accuracy when the proposed backdoor removal method is implemented. As the figure demonstrates, each attack setting effectively injects stealthy backdoors into the global model without impacting the main task accuracy $Acc_{\mathcal{M}}$, consistent with the state-of-the-art results. This demonstrates the effectiveness of the attack strategies in achieving their intended goals. It is to be noted that the continuous selection strategy attains the desired backdoor accuracy earlier than the other two strategies because of the continuous accumulation of malicious updates in the global model. Furthermore, we observe that the backdoor accuracy $Acc^{normal}_{\mathcal{B}}$ progressively diminishes over subsequent FL training rounds, aligning with expectations, as backdoors are not intrinsically persistent within the global model due to the inherent training dynamics of the FL framework. However, upon applying the proposed backdoor removal method, we notice that the final backdoor accuracy of the global model is reduced by 47.87\% for ID1, 41.93\% for ID2, 48.93\% for ID3, and 46.05\% for ID4 on average over all poisoning strategies
in comparison to scenarios where no unlearning is employed, where ID1, ID2, ID3, and ID4 refers to the respective configurations mentioned in Table~\ref{table:attack_settings}. It is to be noted that different combinations of datasets, neural network architectures, trigger patterns, backdoor insertion techniques, and poisoning strategies create backdoor models with varying impacts and complexities. The observation mentioned above highlights the effectiveness of the proposed approach in removing the influence of backdoors for all these combinations.

It is crucial to highlight that the standard deviation for the random selection strategy is substantially higher across all configurations compared to other poisoning strategies. Importantly, this considerable variation in accuracies is not attributed to the performance of our proposed method but rather to the inherent randomness in client participation.

\subsection{Stealthiness during Backdoor Unlearning}
\begin{figure*}[!t]
     \centering
     \begin{subfigure}[t]{0.24\linewidth}
         \centering
         \includegraphics[width=\linewidth]{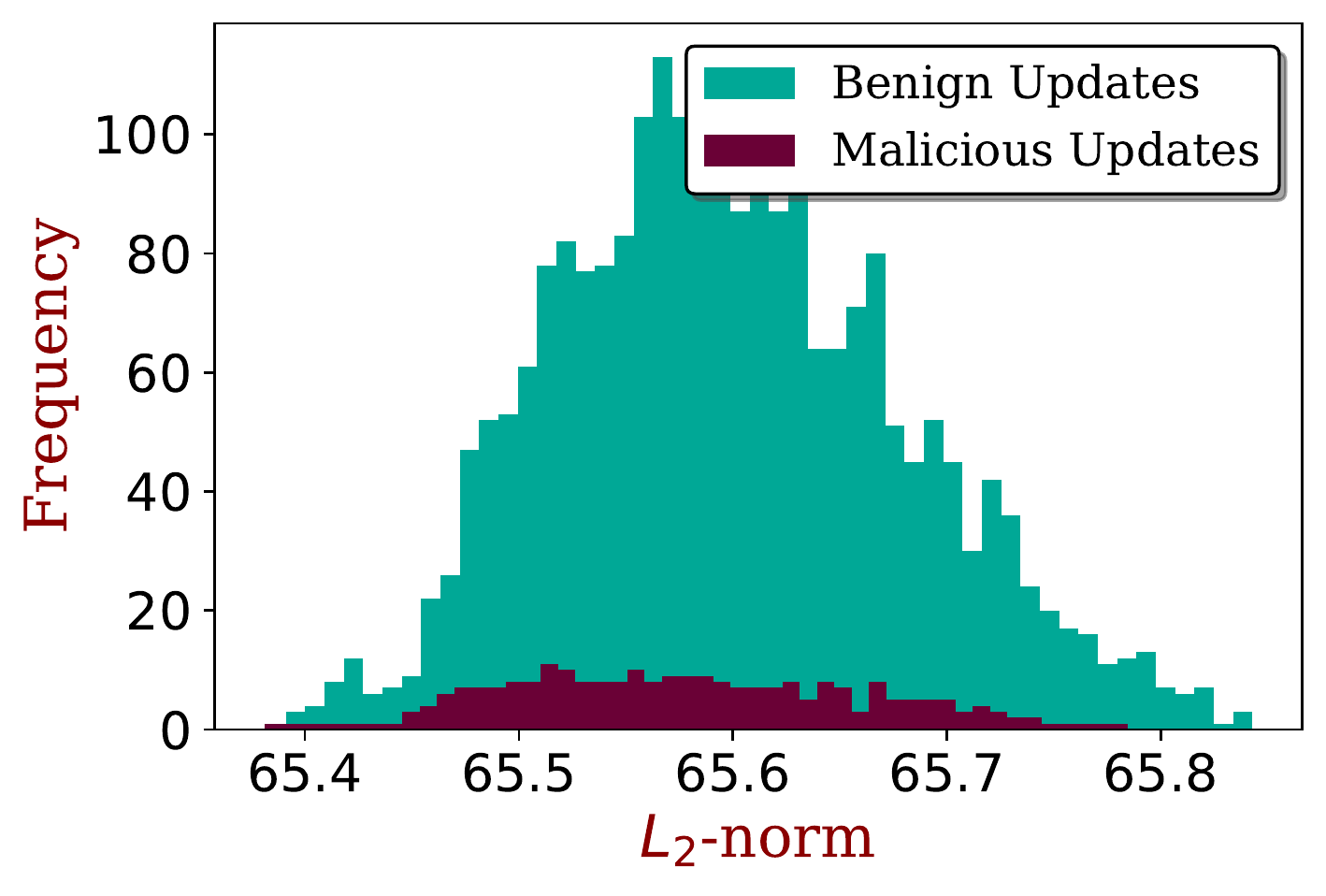}
         \caption{\textbf{ID1:} Continuous Selection}
     \end{subfigure}
     \begin{subfigure}[t]{0.24\linewidth}
         \centering
         \includegraphics[width=\linewidth]{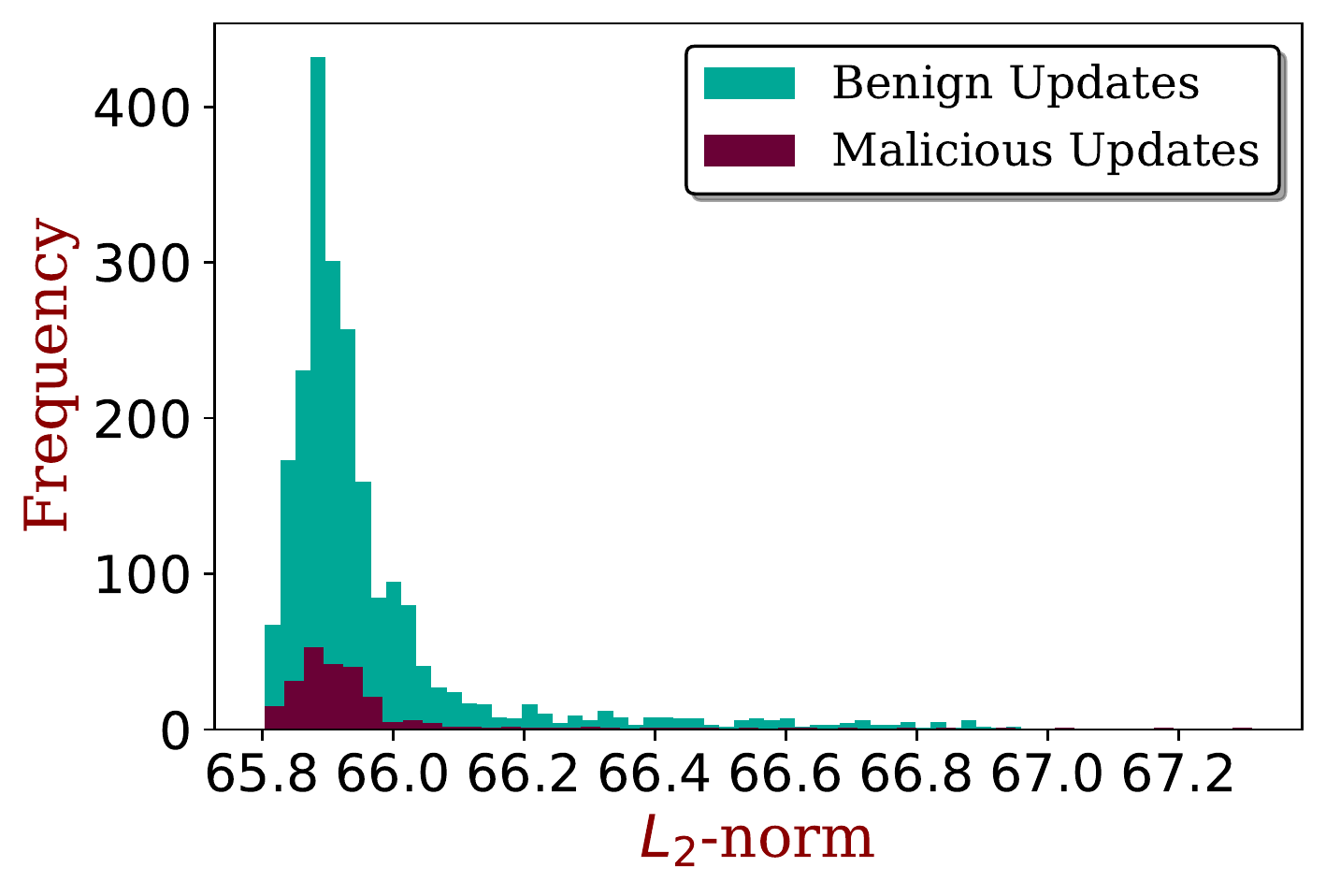}
         \caption{\textbf{ID2:} Continuous Selection}
     \end{subfigure}
     \begin{subfigure}[t]{0.24\linewidth}
         \centering
         \includegraphics[width=\linewidth]{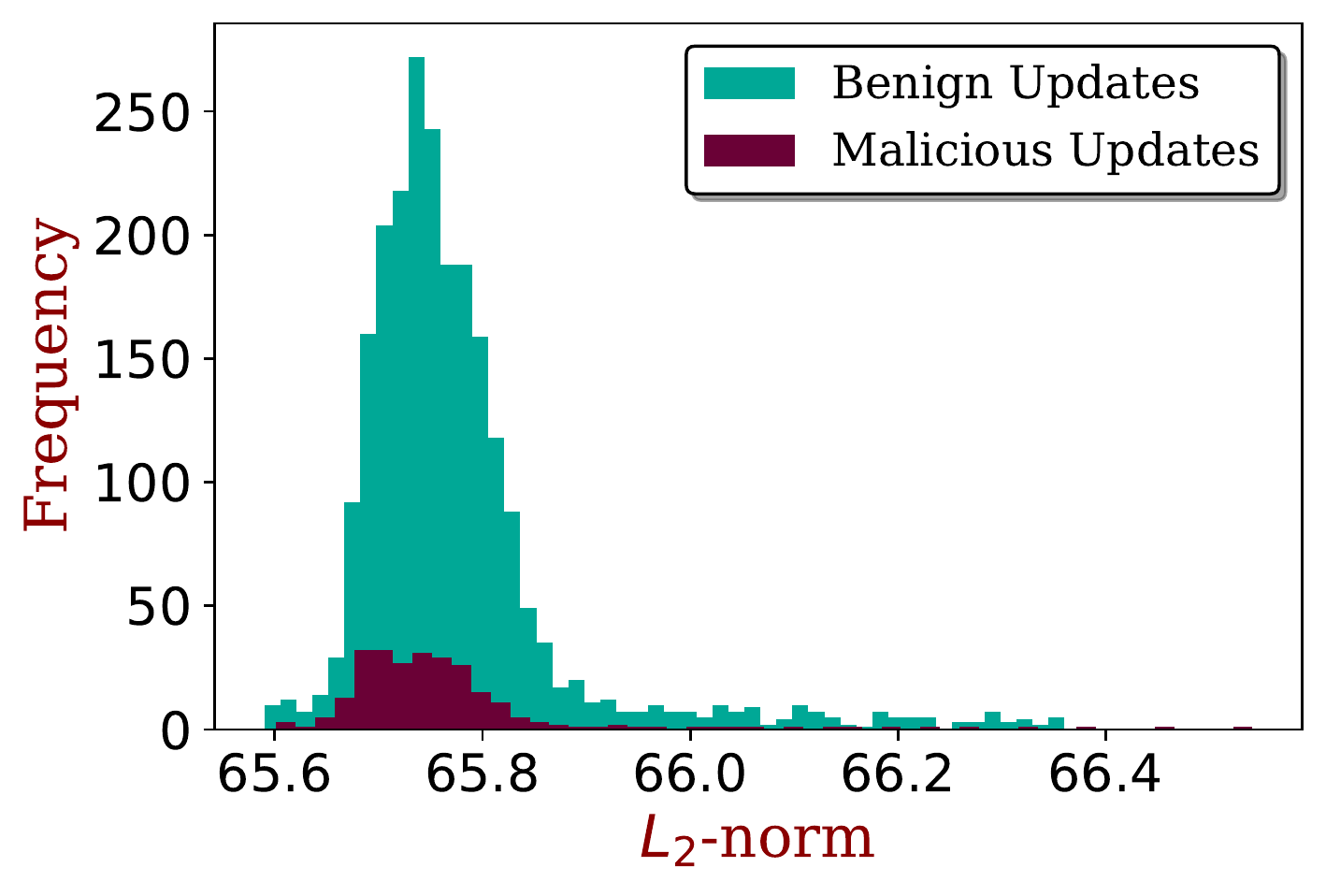}
         \caption{\textbf{ID3:} Continuous Selection}
     \end{subfigure}
     \begin{subfigure}[t]{0.24\linewidth}
         \centering
         \includegraphics[width=\linewidth]{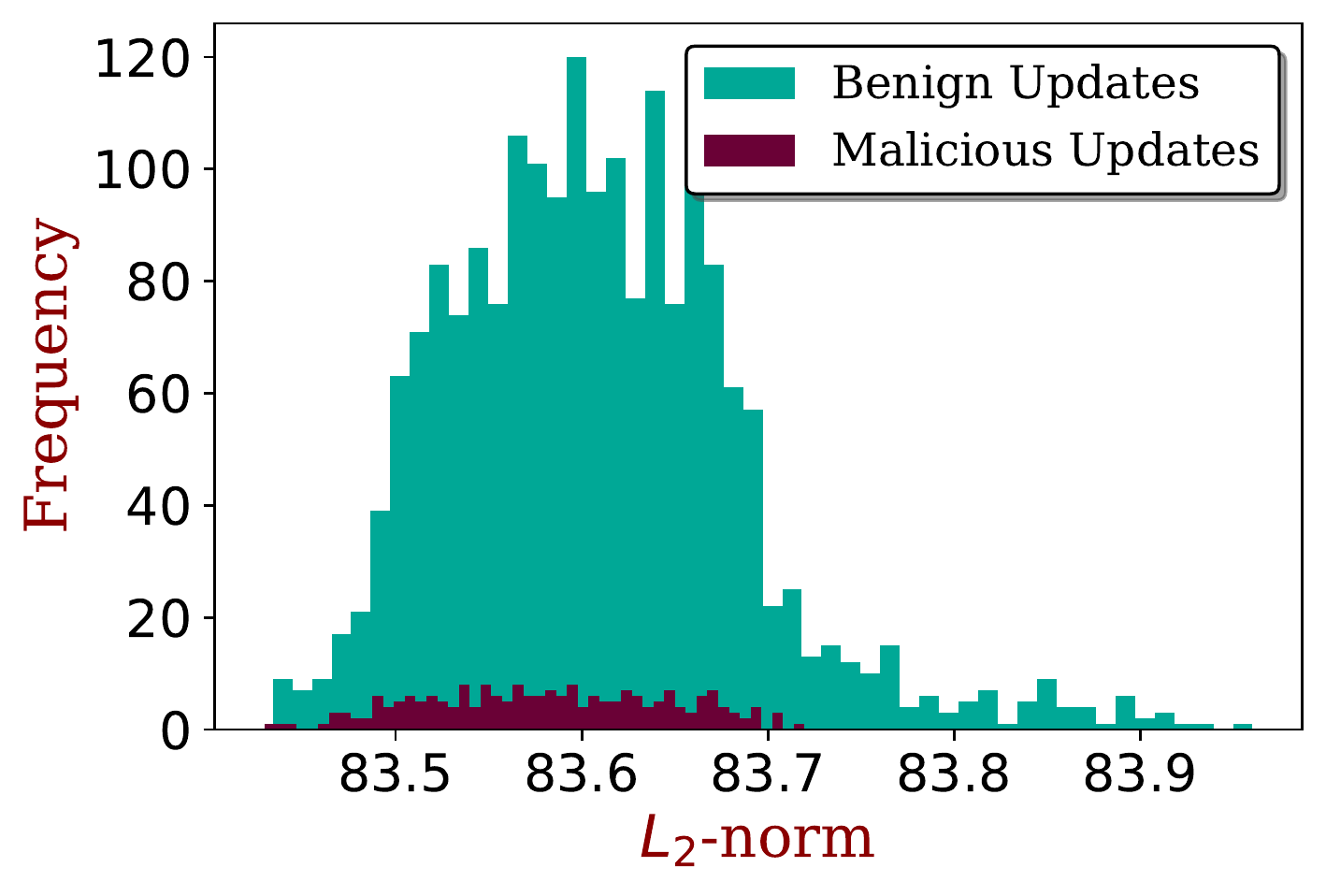}
         \caption{\textbf{ID4:} Continuous Selection}
     \end{subfigure}
     \begin{subfigure}[t]{0.24\linewidth}
         \centering
         \includegraphics[width=\linewidth]{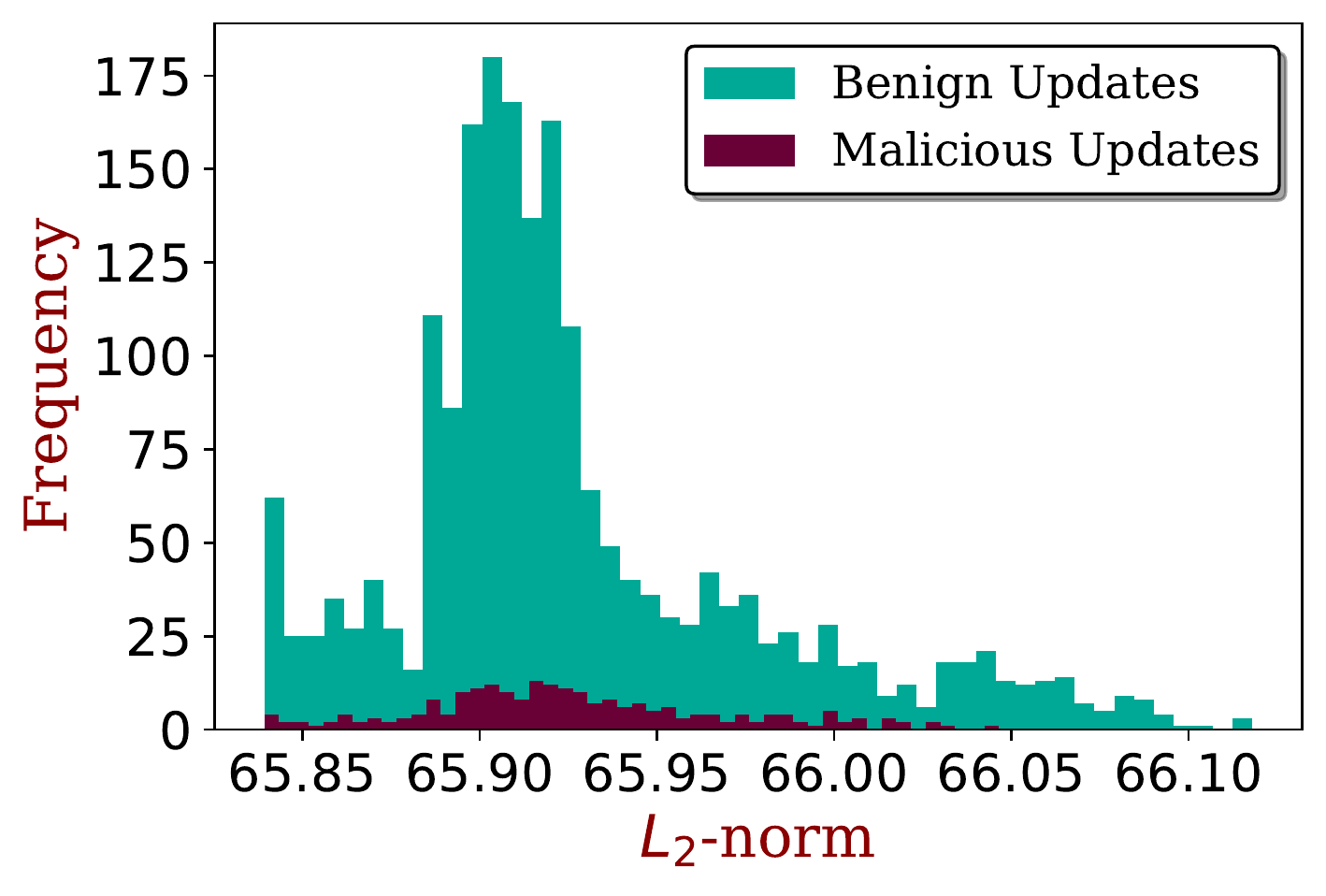}
         \caption{\textbf{ID1:} Fixed-Freq. Selection}
     \end{subfigure}
     \begin{subfigure}[t]{0.24\linewidth}
         \centering
         \includegraphics[width=\linewidth]{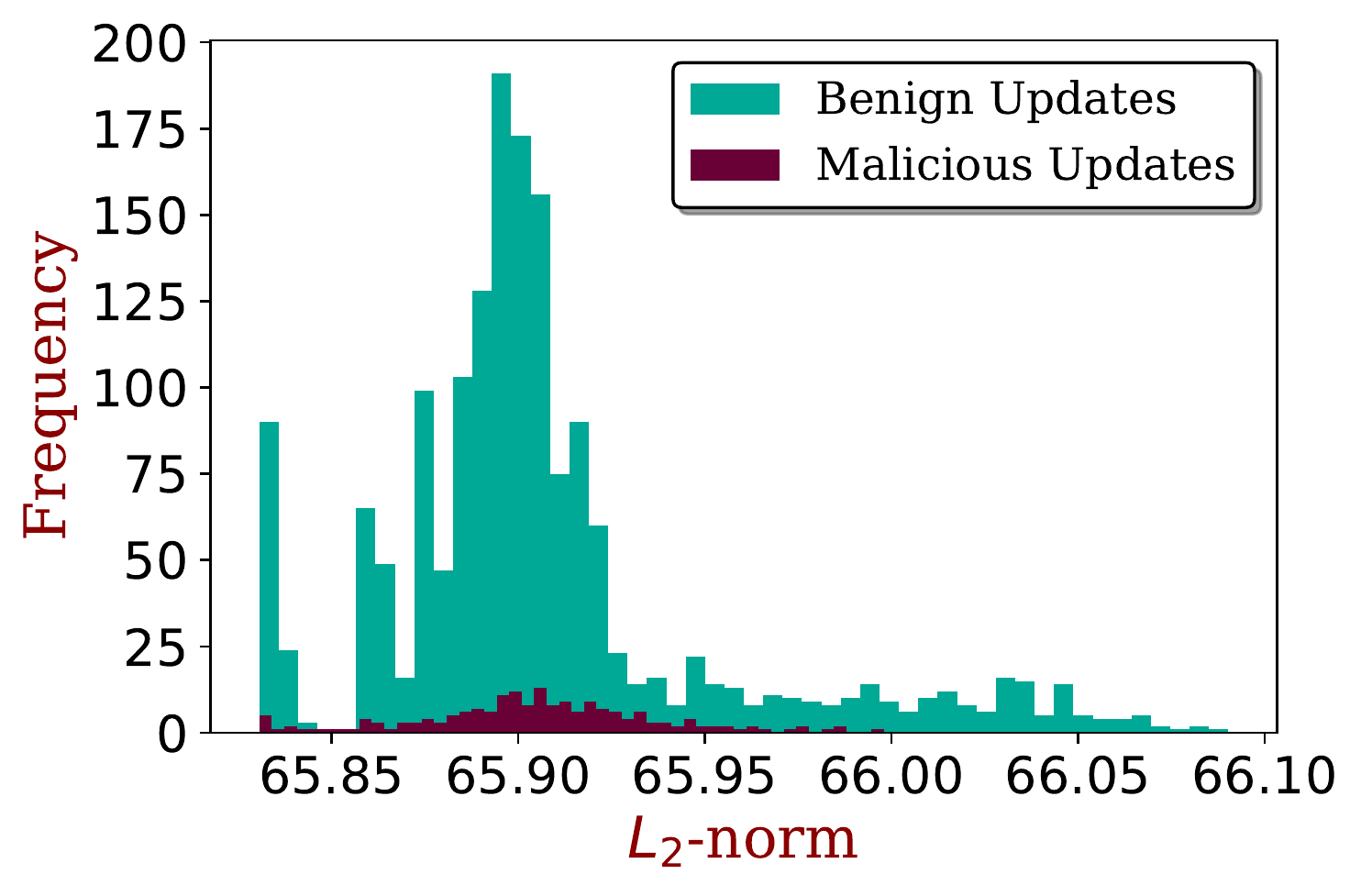}
         \caption{\textbf{ID2:} Fixed-Freq. Selection}
     \end{subfigure}
     \begin{subfigure}[t]{0.24\linewidth}
         \centering
         \includegraphics[width=\linewidth]{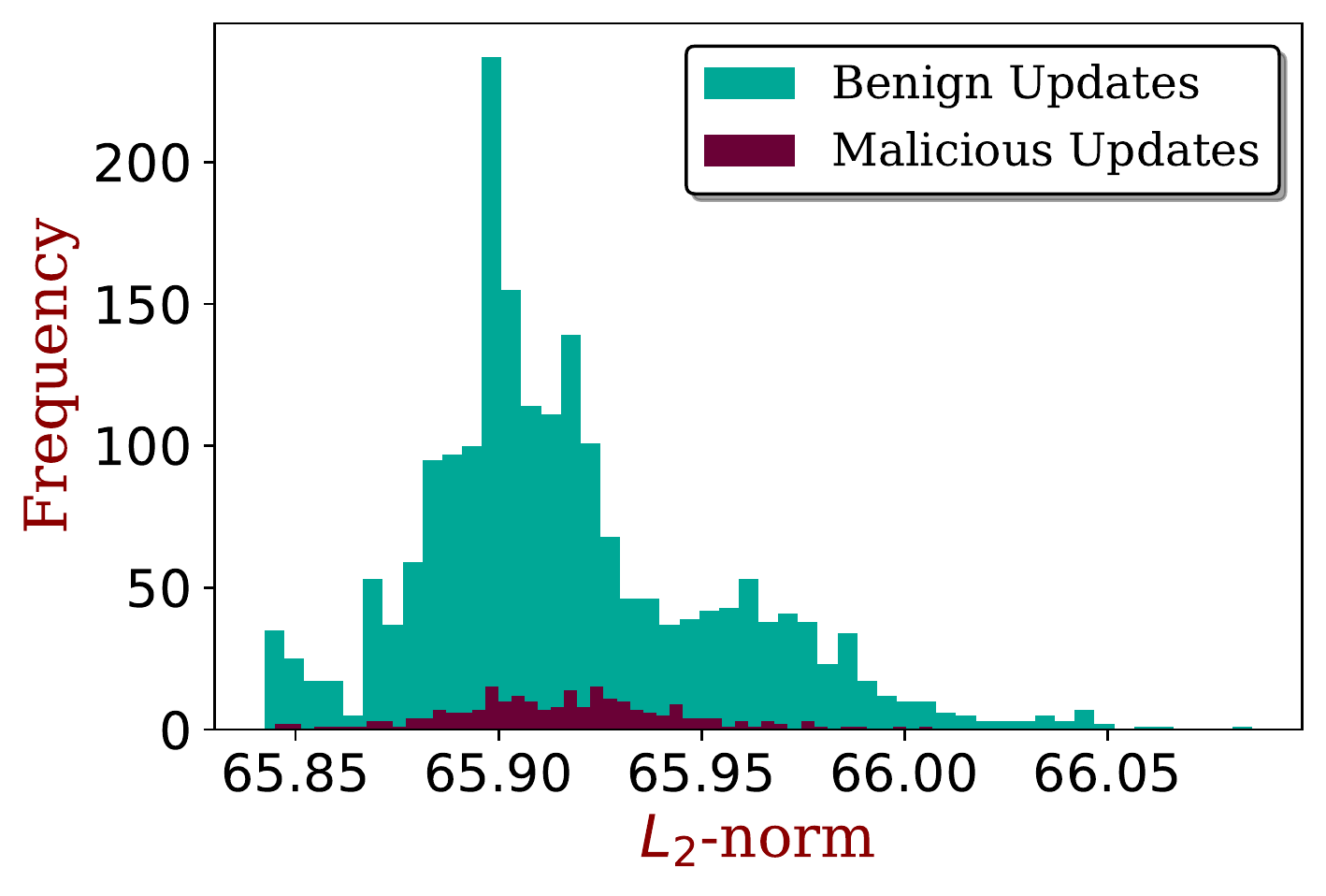}
         \caption{\textbf{ID3:} Fixed-Freq. Selection}
     \end{subfigure}
     \begin{subfigure}[t]{0.24\linewidth}
         \centering
         \includegraphics[width=\linewidth]{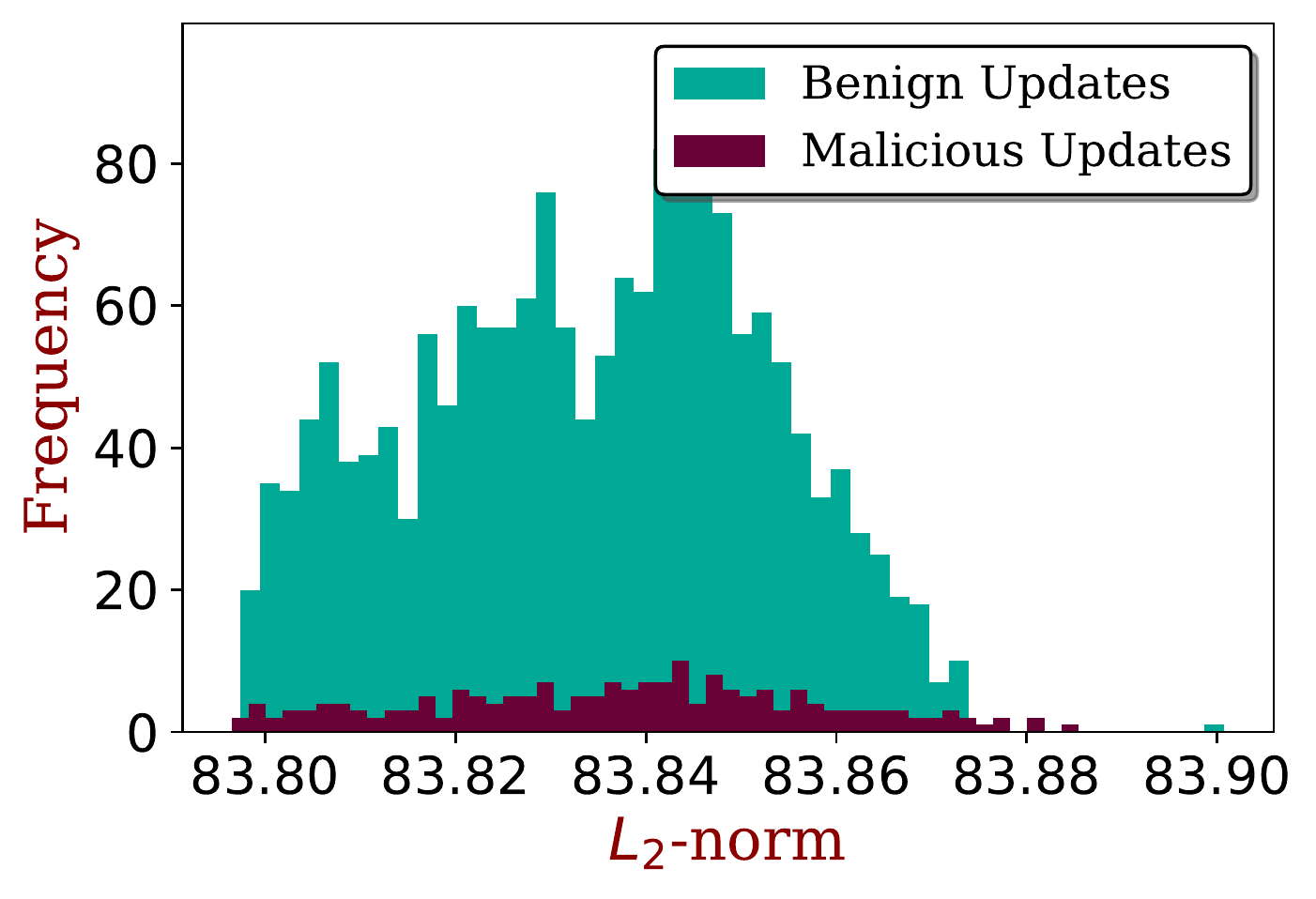}
         \caption{\textbf{ID4:} Fixed-Freq. Selection}
     \end{subfigure}
     \begin{subfigure}[t]{0.24\linewidth}
         \centering
         \includegraphics[width=\linewidth]{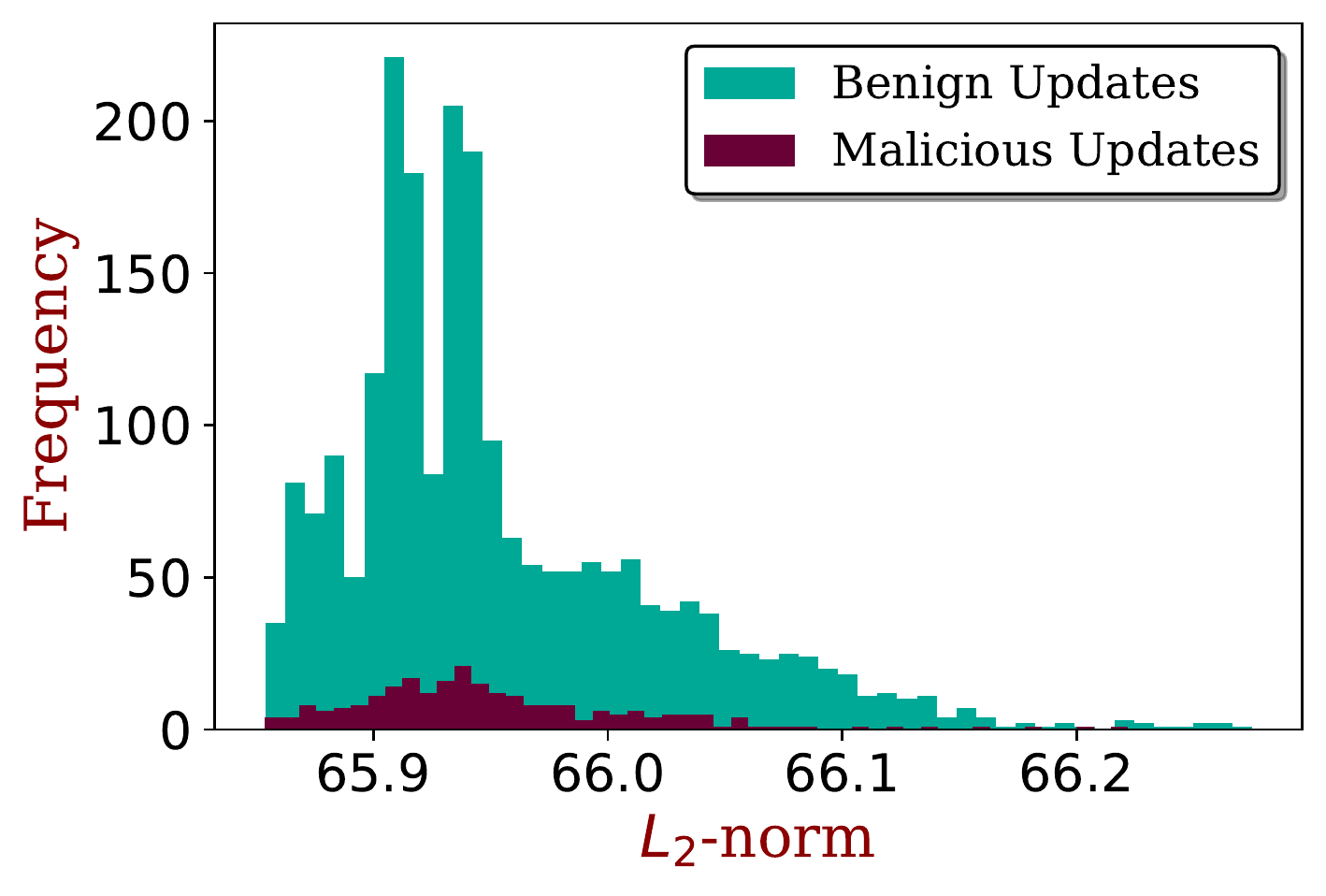}
         \caption{\textbf{ID1:} Random Selection}
     \end{subfigure}
     \begin{subfigure}[t]{0.24\linewidth}
         \centering
         \includegraphics[width=\linewidth]{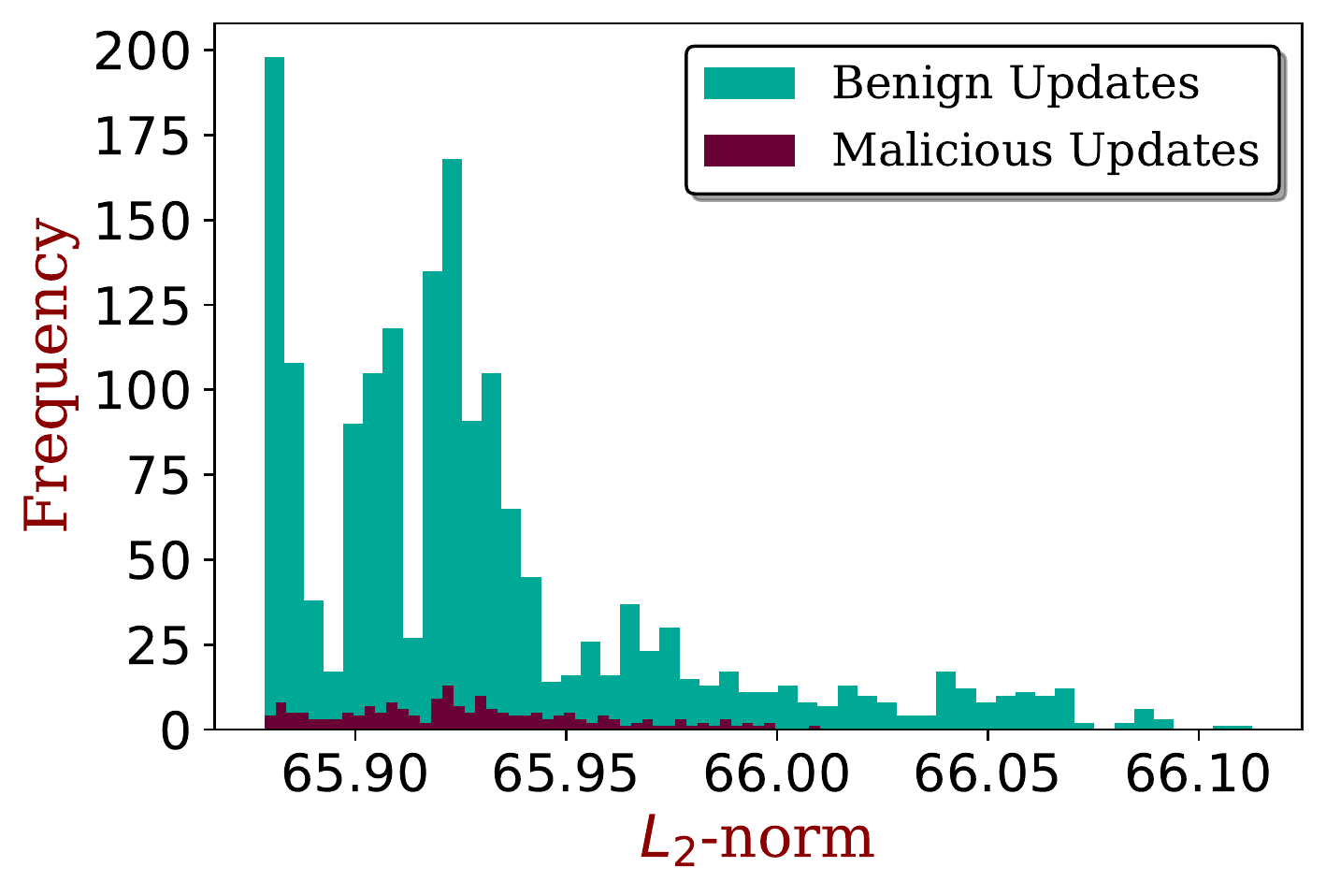}
         \caption{\textbf{ID2:} Random Selection}
     \end{subfigure}
     \begin{subfigure}[t]{0.24\linewidth}
         \centering
         \includegraphics[width=\linewidth]{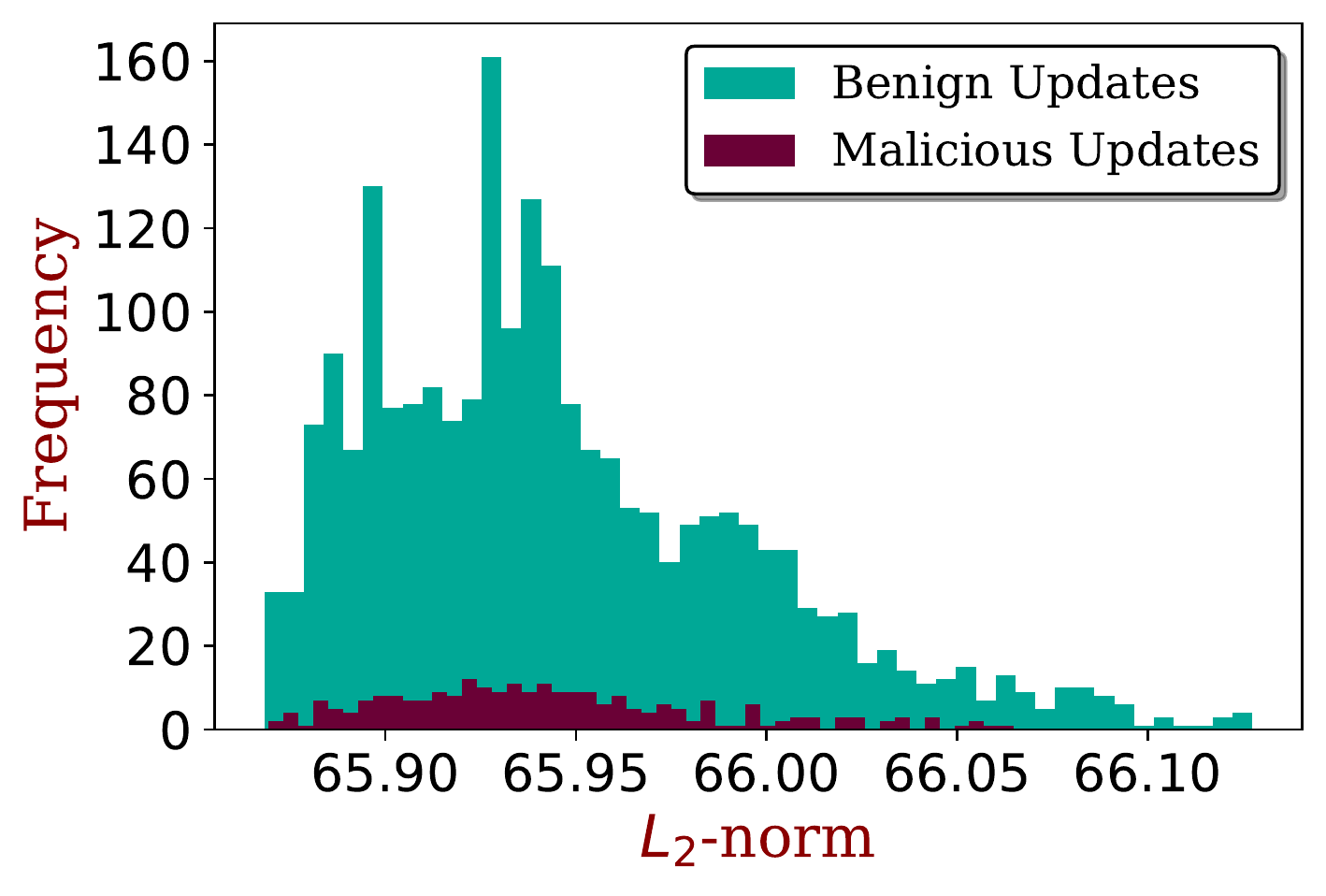}
         \caption{\textbf{ID3:} Random Selection}
     \end{subfigure}
     \begin{subfigure}[t]{0.24\linewidth}
         \centering
         \includegraphics[width=\linewidth]{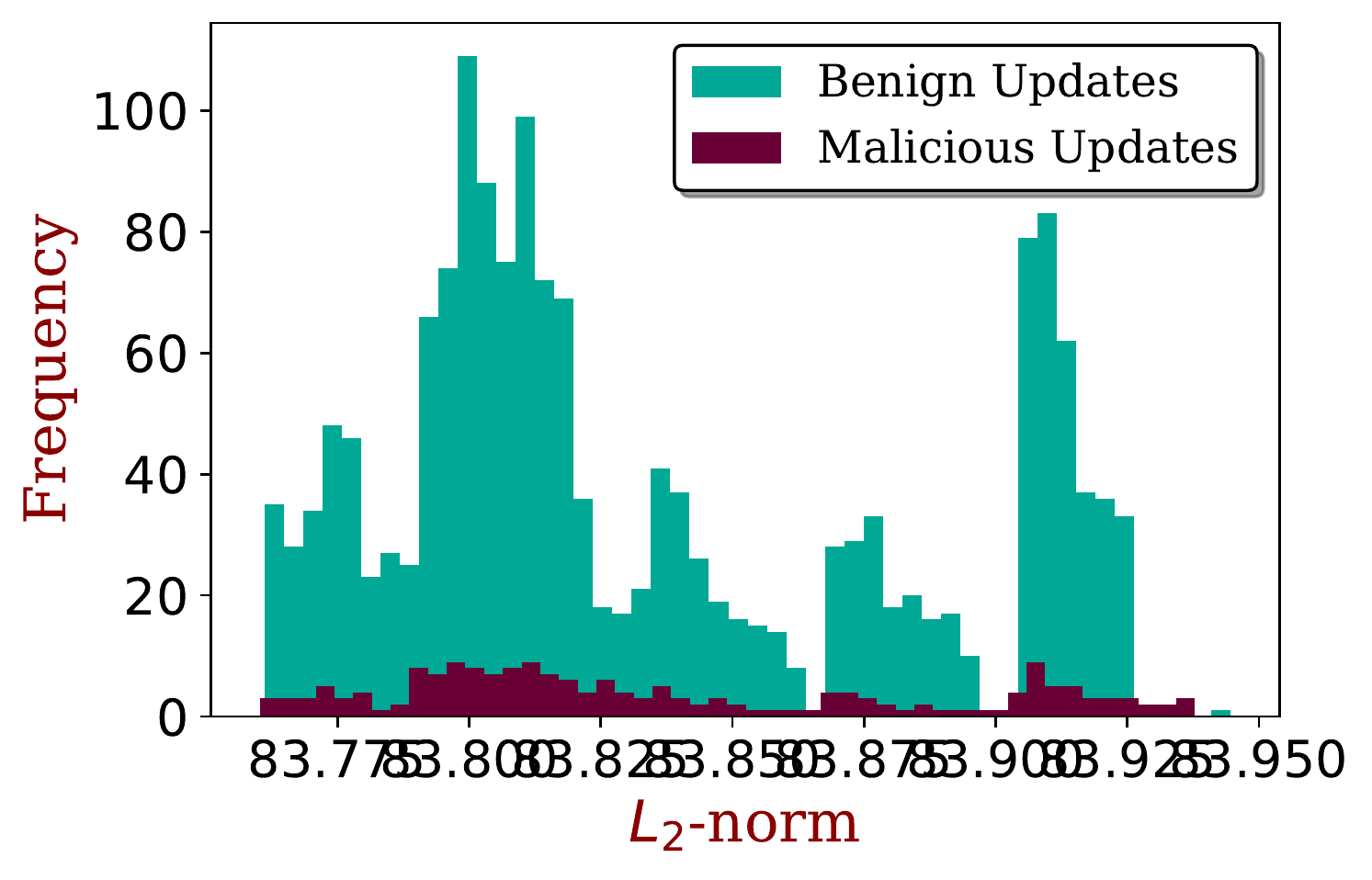}
         \caption{\textbf{ID4:} Random Selection}
     \end{subfigure}
\caption{Distribution of the $L_2$-norm of the difference between local model updates and the global model for all participants across various backdoor insertion settings and poisoning strategies.}\label{fig:unlearn_stealth}
\end{figure*}

We assess the stealthiness of the proposed backdoor removal method within the context of FL by examining the $L_2$-norm of the difference between local model updates and the global model. We take into account all participants involved in the process, considering the various backdoor insertion settings outlined in Table~\ref{table:attack_settings}, as well as three distinct poisoning strategies, as discussed previously. We analyze the distributions of these norms during both the backdoor insertion and backdoor removal phases, with the results shown in Figure~\ref{fig:unlearn_stealth}.
A key observation from the figure is that the malicious updates overlap with their benign counterparts across all configurations mentioned in Table~\ref{table:attack_settings}. As mentioned earlier, different combinations of attack configurations like datasets, neural network architectures, trigger patterns, backdoor insertion techniques, and poisoning strategies create backdoor models with varying impacts and complexities. The observation mentioned in Figure~\ref{fig:unlearn_stealth} indicates that the proposed method effectively maintains stealth throughout the process of unlearning for all these combinations. Consequently, this approach does not raise any suspicions based on the norm of updates, thereby ensuring the successful implementation of backdoor removal for diverse scenarios.

\subsection{Ablation Studies}

\subsubsection{Unlearning without Weighted Importance}
\begin{figure}[!t]
    \centering
    \includegraphics[width=0.95\linewidth]{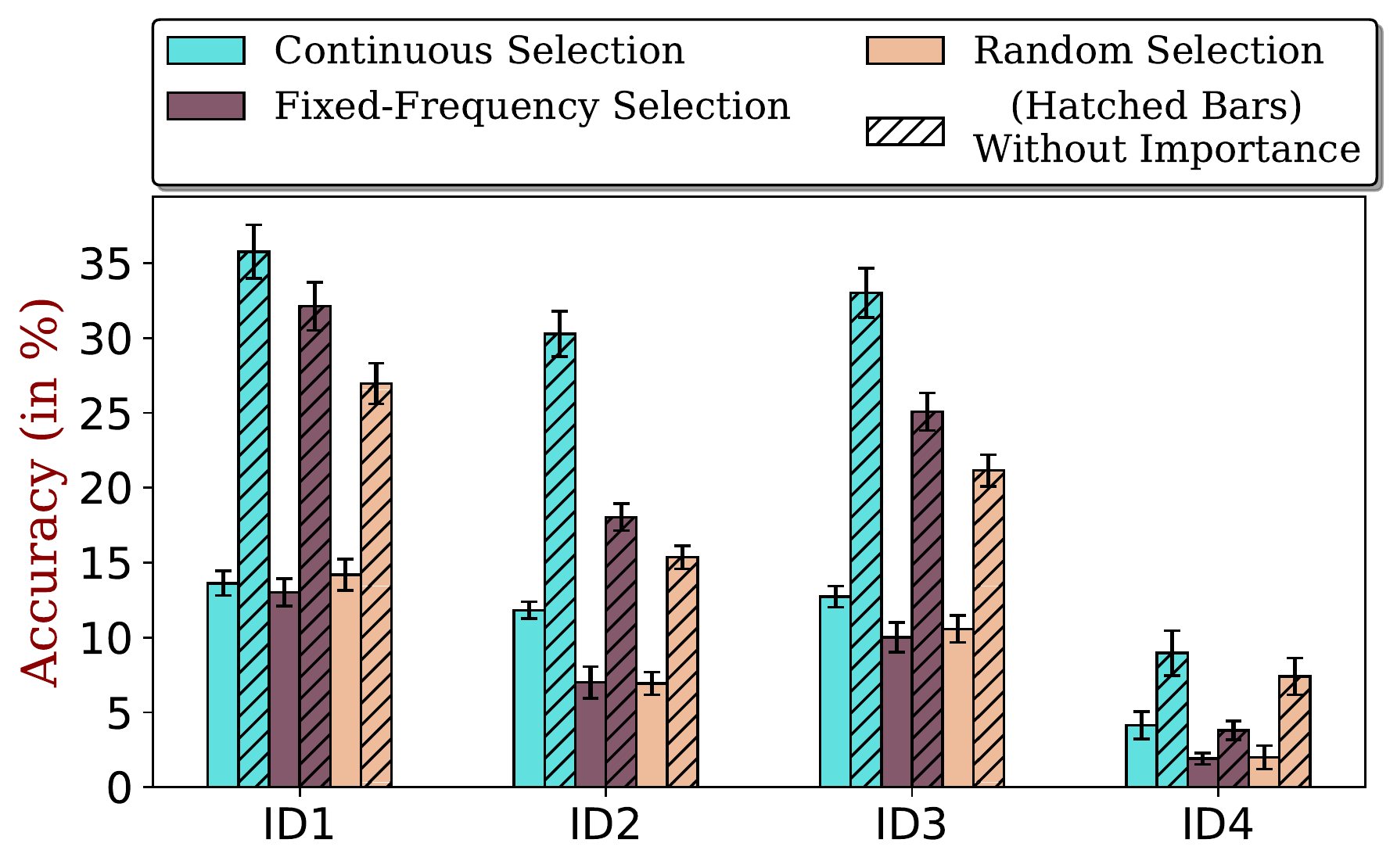}
    \caption{Comparison of final backdoor accuracy for both weighted penalty parameters (solid bars) and unweighted penalty parameters (hatched bars) across all backdoor insertion settings and poisoning strategies. Error bars represent the standard deviation across ten independent runs with different random seeds.}
    \label{fig:ablation_weighted_importance}
\end{figure}

In Section~\ref{sec:dynamic_penalization}, we introduced a dynamic penalization approach, which involves assigning weights of varying magnitudes to penalty terms. In this section, we evaluate the performance of the proposed backdoor removal method in the absence of the weighted penalty (which is analogous to the machine unlearning-based defense against backdoor attacks in a conventional machine learning framework as discussed in~\cite{DBLP:conf/infocom/LiuFCLMWM22}). We adopt Equation~(\ref{eq:unlearn_penalty}) as the unlearning loss function and set $\gamma=3$ for this analysis. Our assessment consists of all the backdoor insertion settings outlined in Table~\ref{table:attack_settings}, along with the three poisoning strategies discussed earlier. Figure~\ref{fig:ablation_weighted_importance} presents the results of this analysis, where solid bars indicate the final backdoor accuracy after 300 rounds for continuous selection and after 500 rounds for both fixed-frequency and random selection, taking into account the proposed weighted penalty parameters (i.e., for the loss function mentioned in Equation~(\ref{eq:final_unlearn})).
In contrast, hatched bars represent the final backdoor accuracy when weights are not assigned to the parameters. Each bar in the plot also includes error terms, which denote the standard deviation of the final backdoor accuracy across ten independent runs with different random seeds. As illustrated in Figure~\ref{fig:ablation_weighted_importance}, assigning weights to the penalty computation minimizes the final backdoor accuracy for all attack settings and poisoning strategies. More specifically, upon assigning the weighted penalty the final backdoor accuracy of the global model is reduced by 17.99\% for ID1, 12.64\% for ID2, 15.32\% for ID3, and 4.03\% for ID4 on average over all poisoning strategies in comparison to scenarios where no weighted penalty is used. The observation emphasizes the significance of the weighted dynamic penalization strategy for the proposed backdoor removal method.

\begin{figure*}[!t]
     \centering
     \begin{subfigure}[t]{0.24\linewidth}
         \centering
         \includegraphics[width=\linewidth]{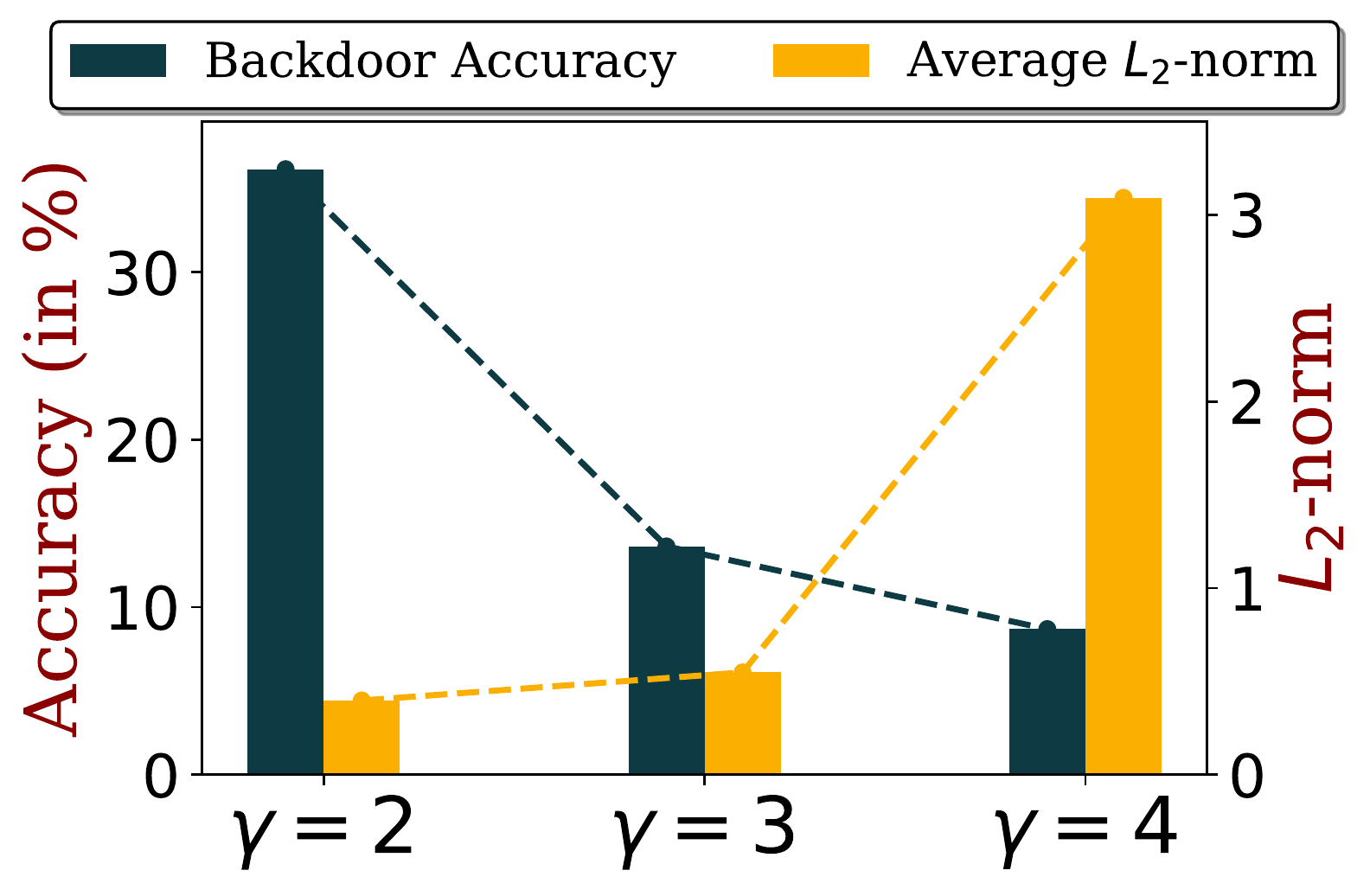}
         \caption{\textbf{ID1:} Continuous Selection}
     \end{subfigure}
     \begin{subfigure}[t]{0.24\linewidth}
         \centering
         \includegraphics[width=\linewidth]{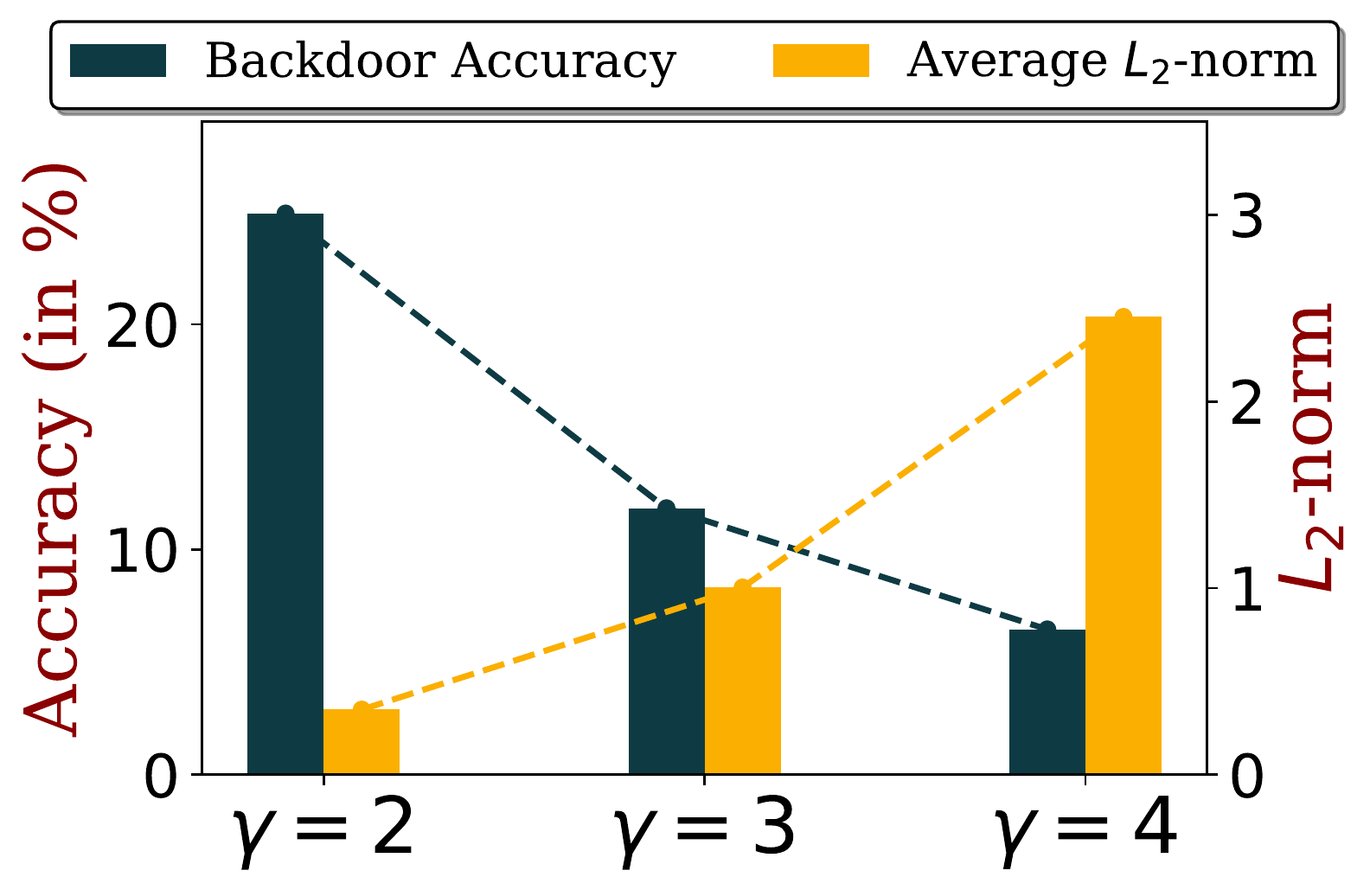}
         \caption{\textbf{ID2:} Continuous Selection}
     \end{subfigure}
     \begin{subfigure}[t]{0.24\linewidth}
         \centering
         \includegraphics[width=\linewidth]{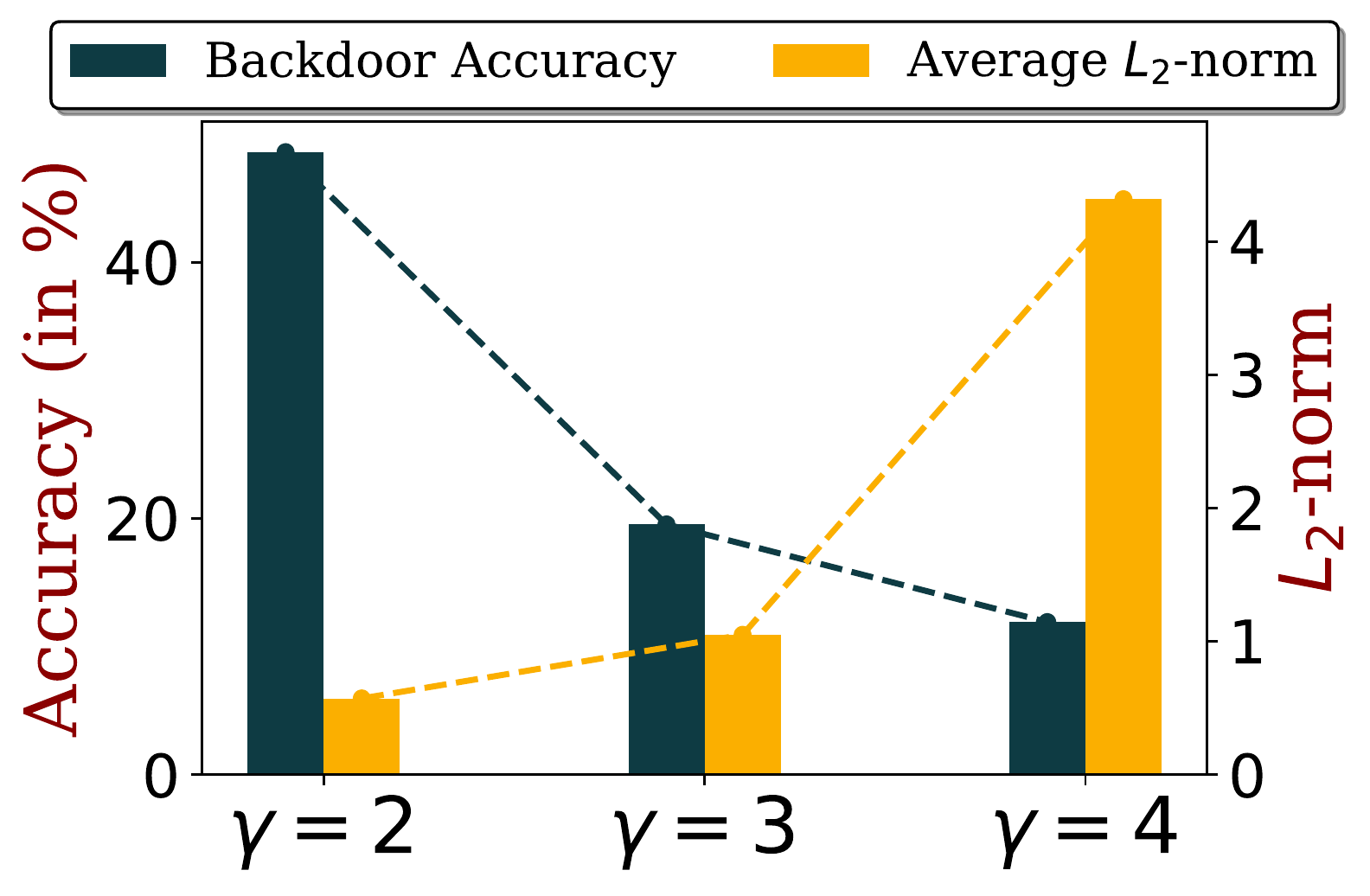}
         \caption{\textbf{ID3:} Continuous Selection}
     \end{subfigure}
     \begin{subfigure}[t]{0.24\linewidth}
         \centering
         \includegraphics[width=\linewidth]{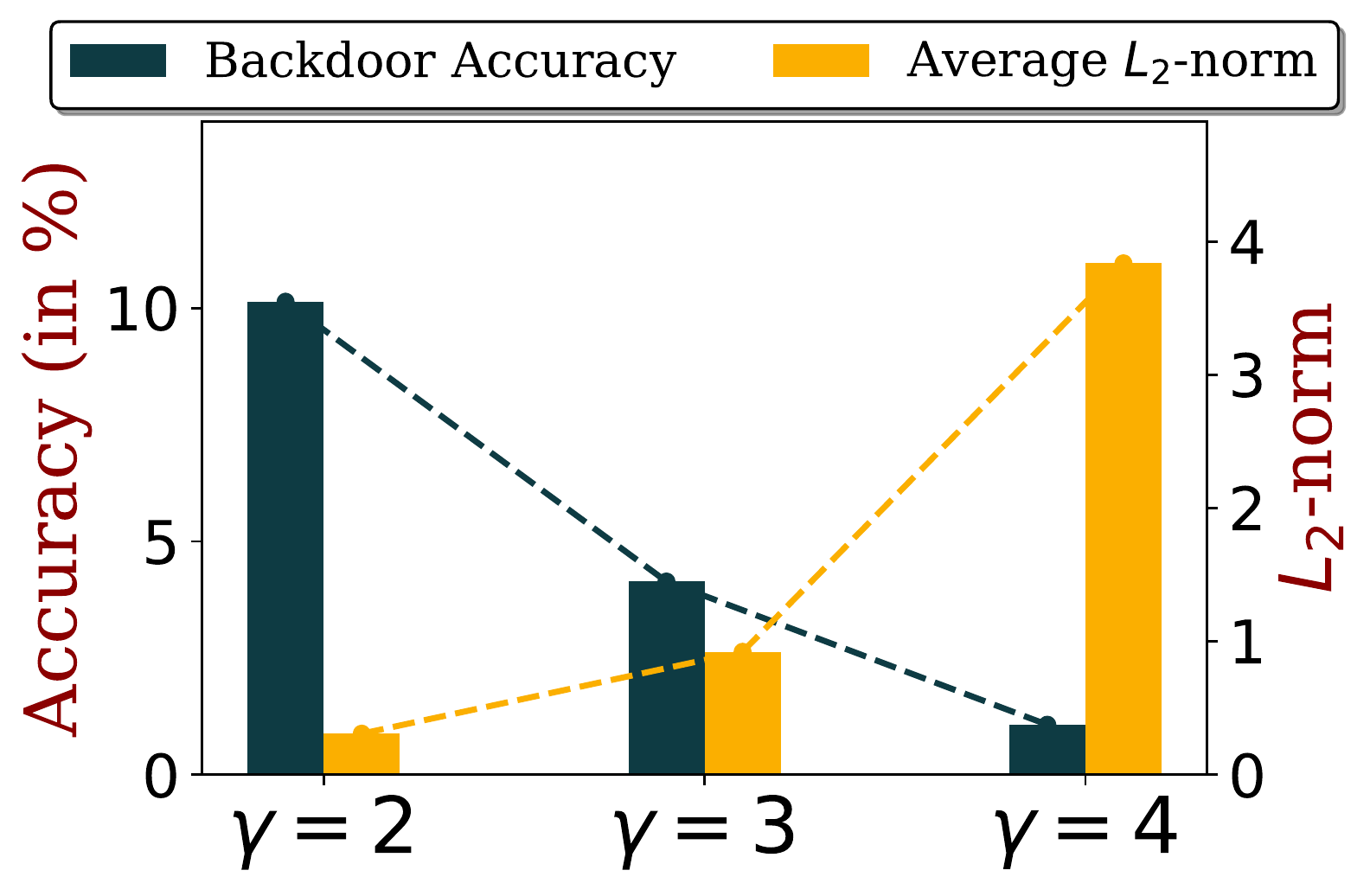}
         \caption{\textbf{ID4:} Continuous Selection}
     \end{subfigure}
     \begin{subfigure}[t]{0.24\linewidth}
         \centering
         \includegraphics[width=\linewidth]{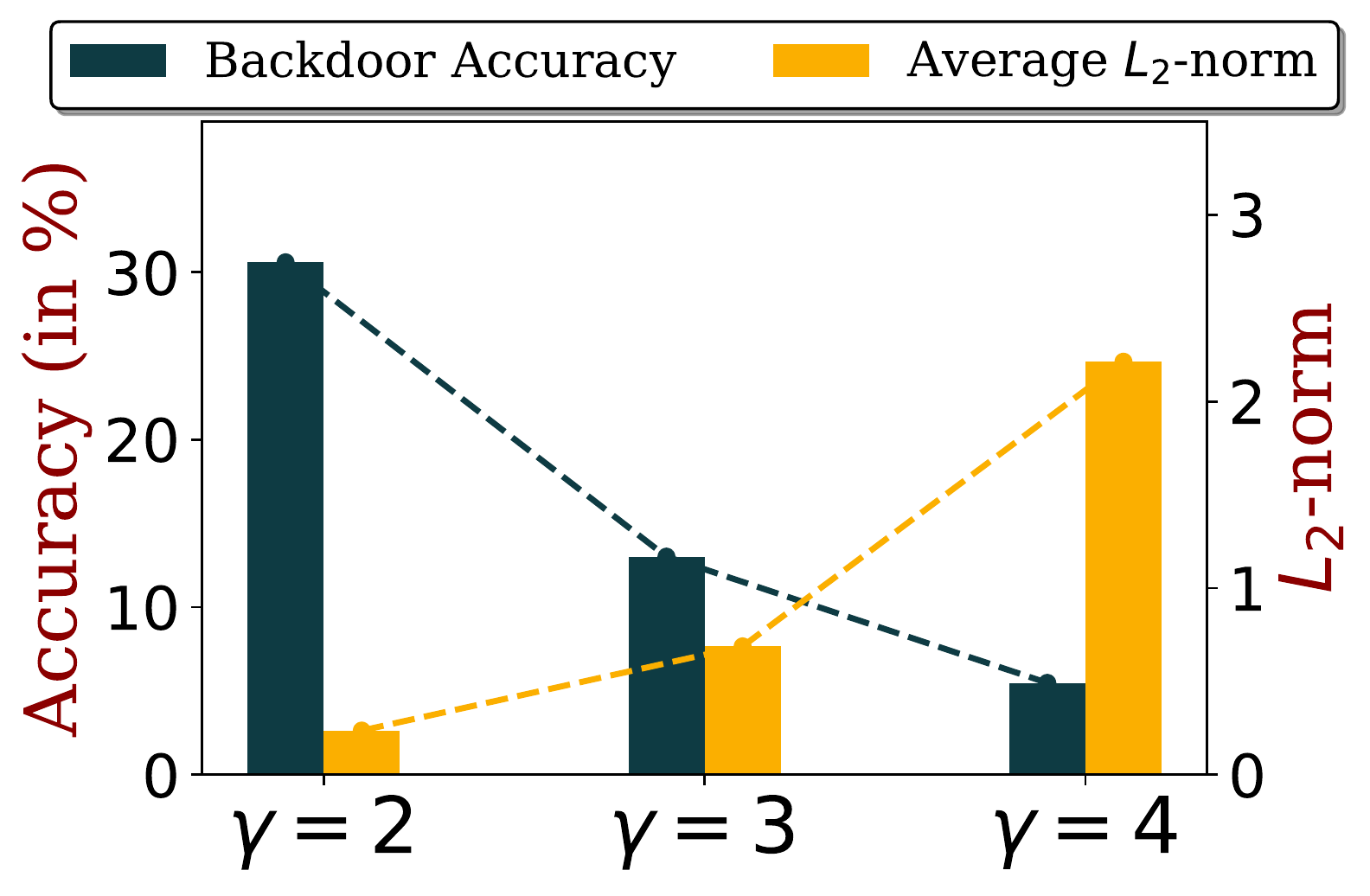}
         \caption{\textbf{ID1:} Fixed Frequency Selection}
     \end{subfigure}
     \begin{subfigure}[t]{0.24\linewidth}
         \centering
         \includegraphics[width=\linewidth]{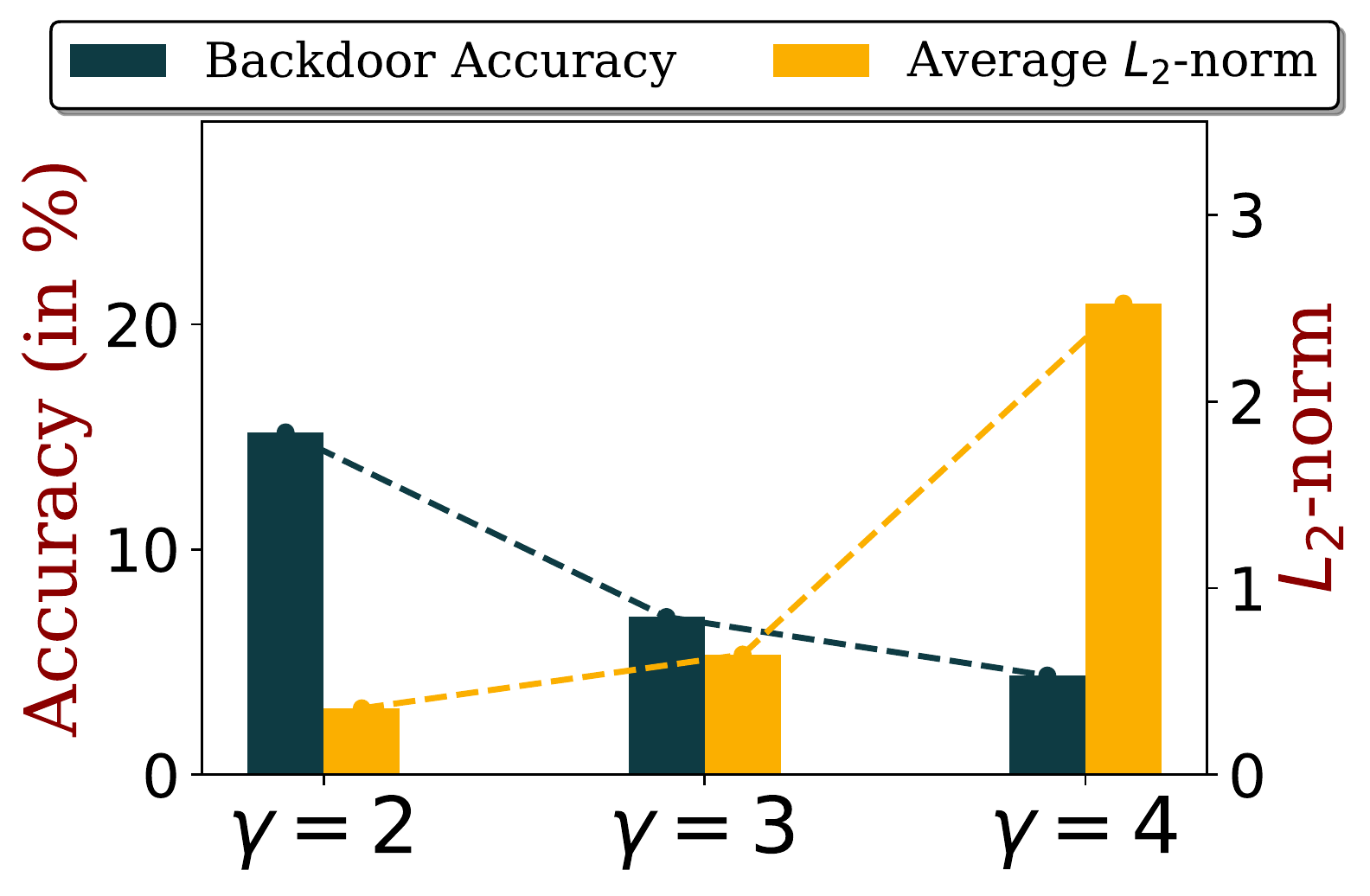}
         \caption{\textbf{ID2:} Fixed Frequency Selection}
     \end{subfigure}
     \begin{subfigure}[t]{0.24\linewidth}
         \centering
         \includegraphics[width=\linewidth]{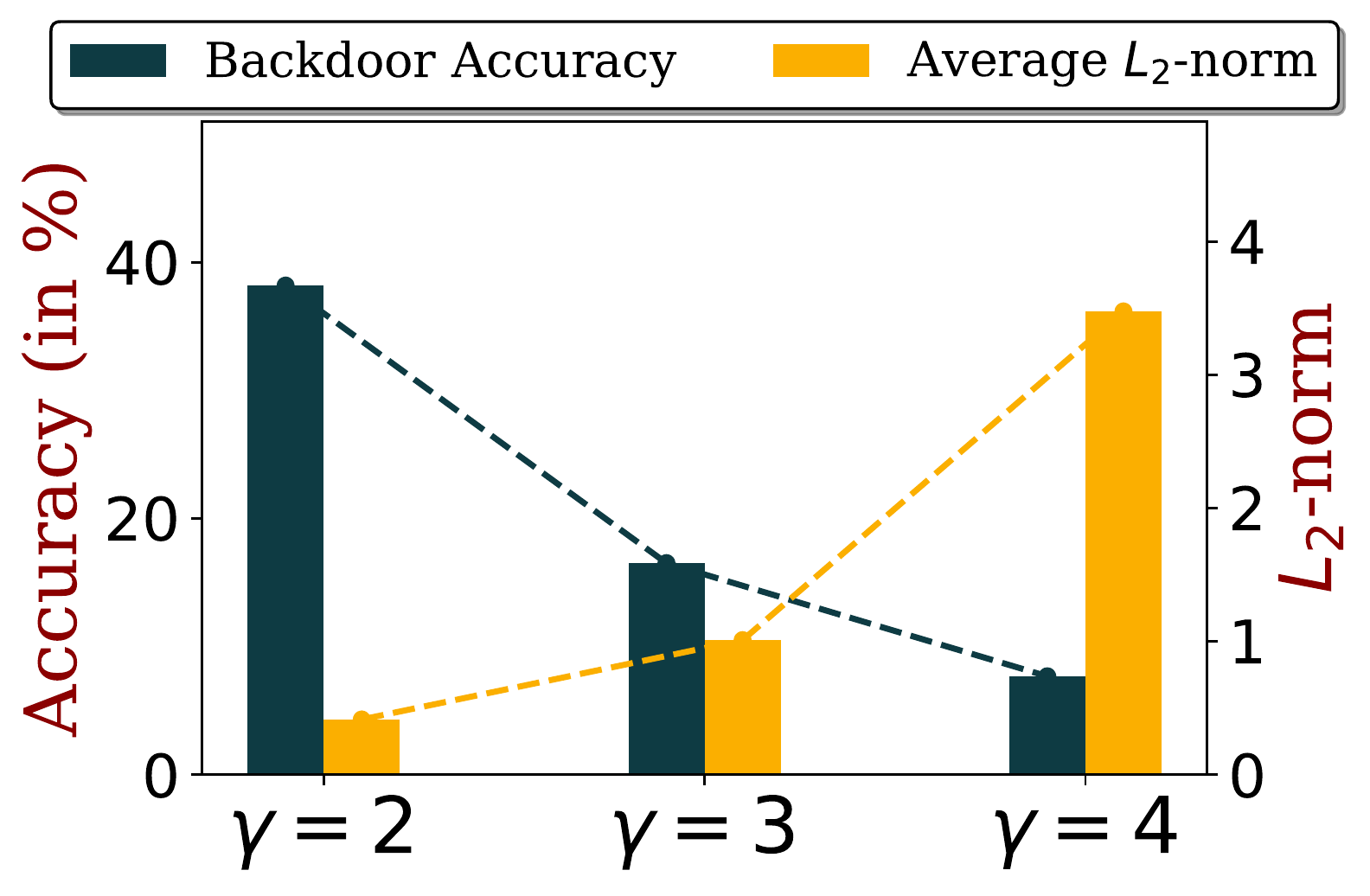}
         \caption{\textbf{ID3:} Fixed Frequency Selection}
     \end{subfigure}
     \begin{subfigure}[t]{0.24\linewidth}
         \centering
         \includegraphics[width=\linewidth]{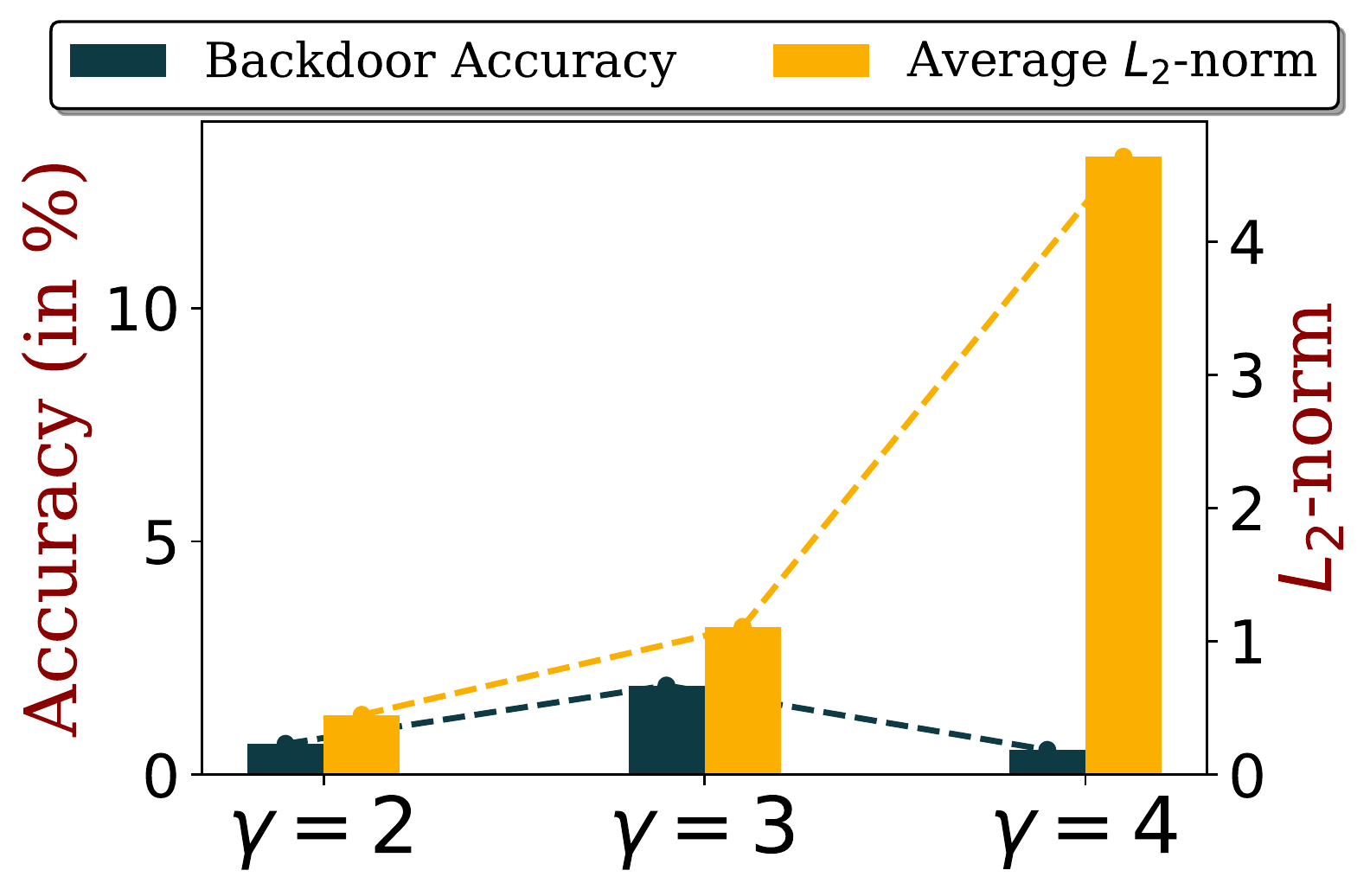}
         \caption{\textbf{ID4:} Fixed Frequency Selection}
     \end{subfigure}
     \begin{subfigure}[t]{0.24\linewidth}
         \centering
         \includegraphics[width=\linewidth]{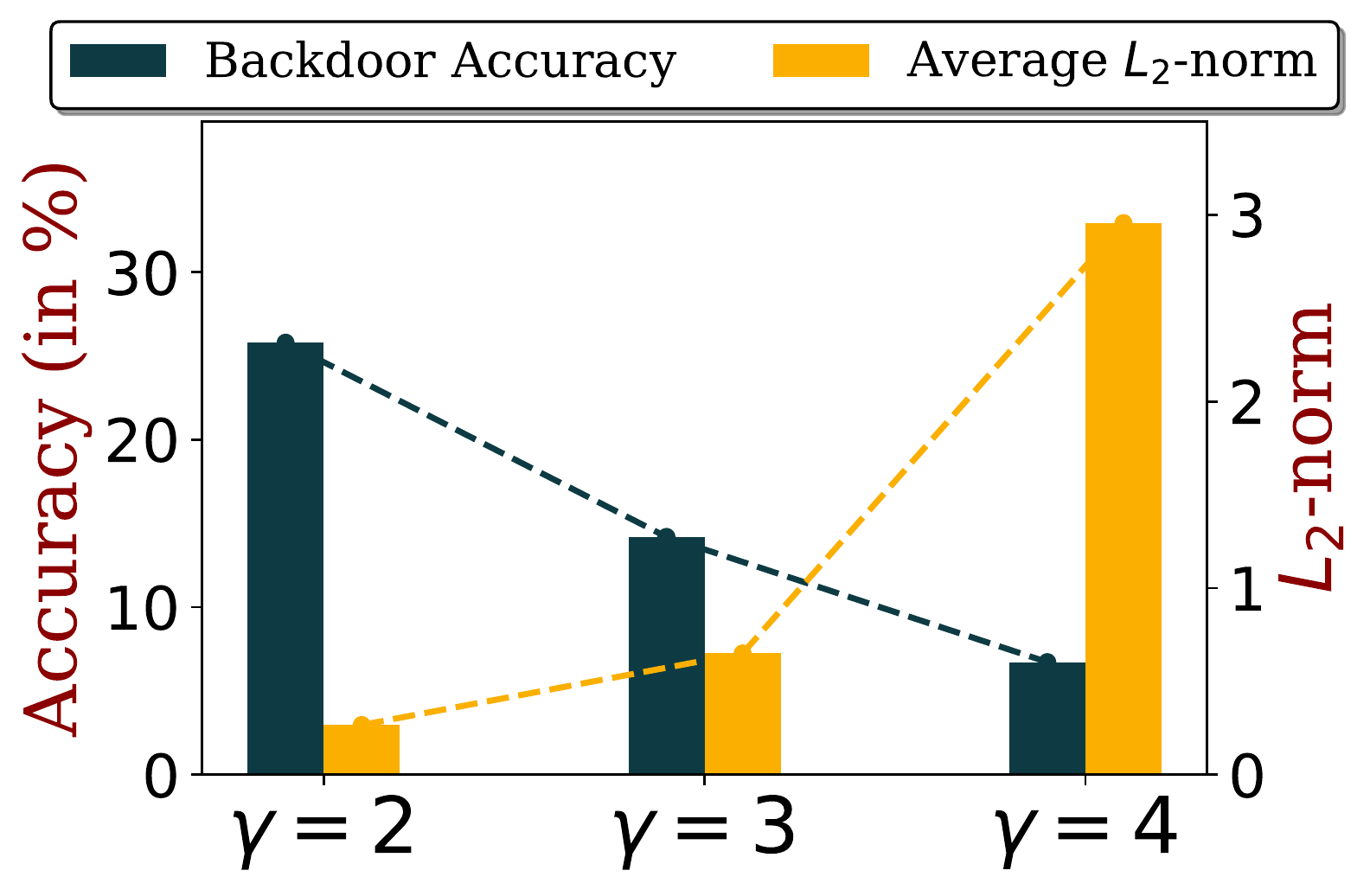}
         \caption{\textbf{ID1:} Random Selection}
     \end{subfigure}
     \begin{subfigure}[t]{0.24\linewidth}
         \centering
         \includegraphics[width=\linewidth]{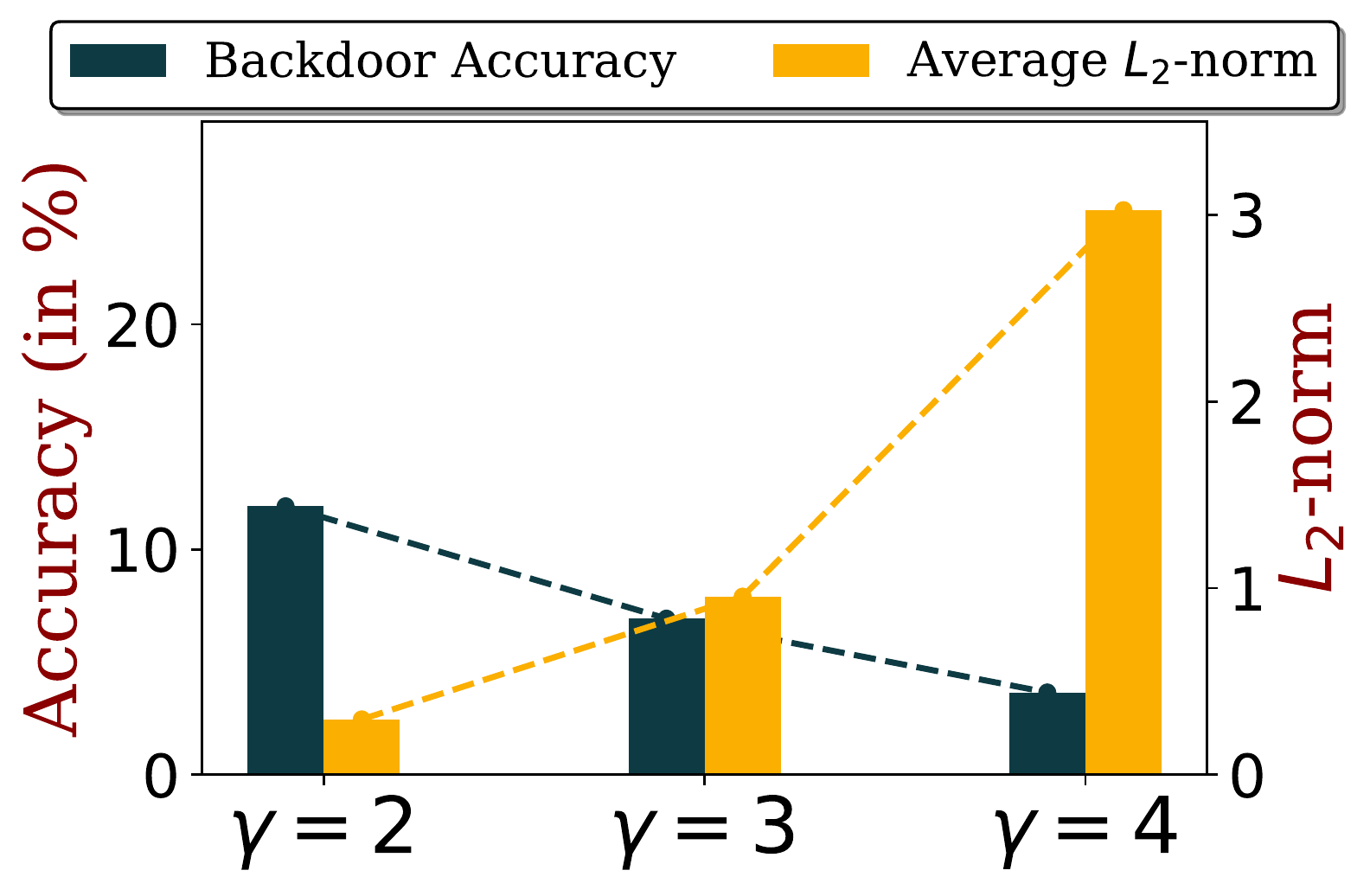}
         \caption{\textbf{ID2:} Random Selection}
     \end{subfigure}
     \begin{subfigure}[t]{0.24\linewidth}
         \centering
         \includegraphics[width=\linewidth]{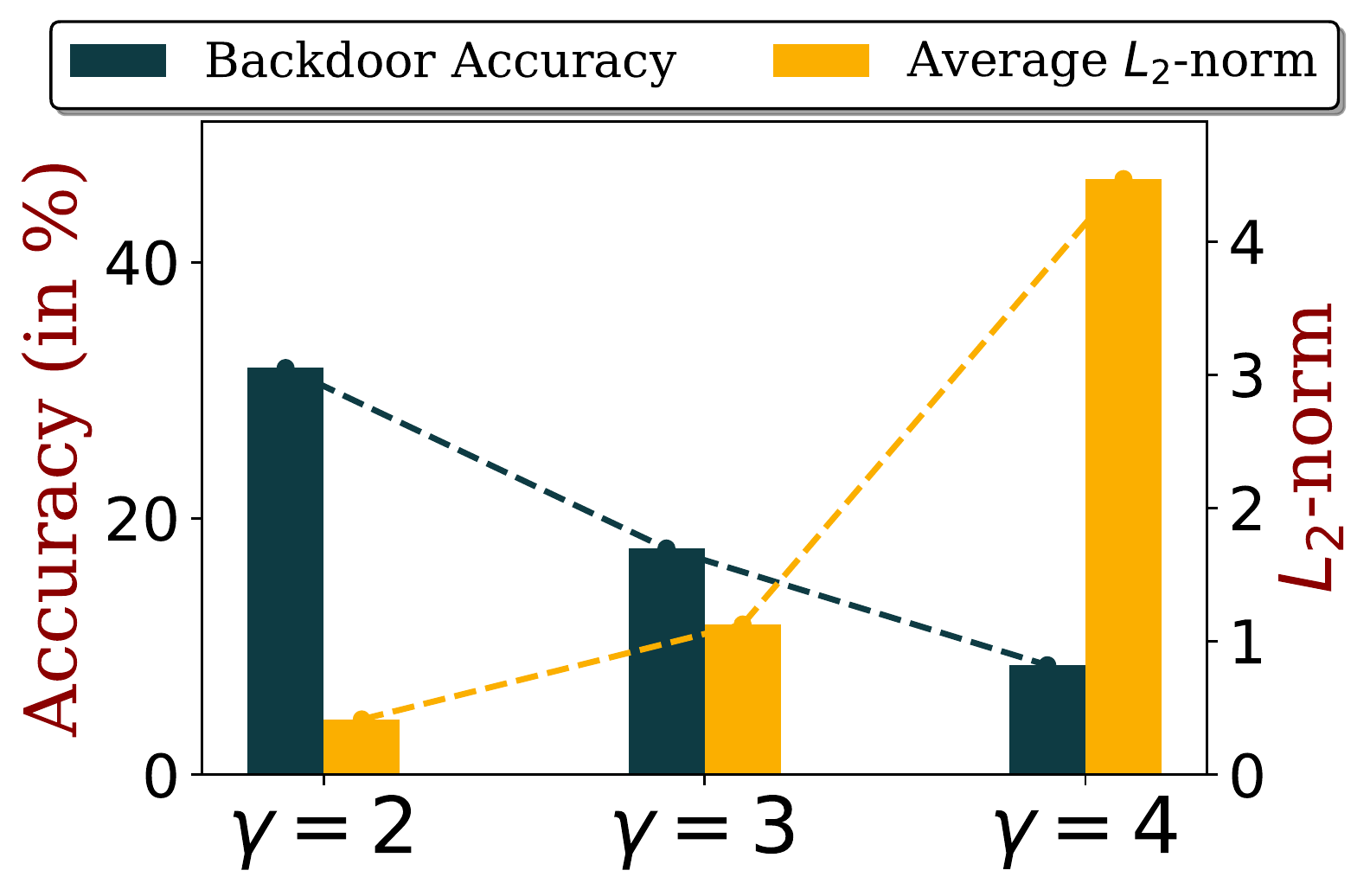}
         \caption{\textbf{ID3:} Random Selection}
     \end{subfigure}
     \begin{subfigure}[t]{0.24\linewidth}
         \centering
         \includegraphics[width=\linewidth]{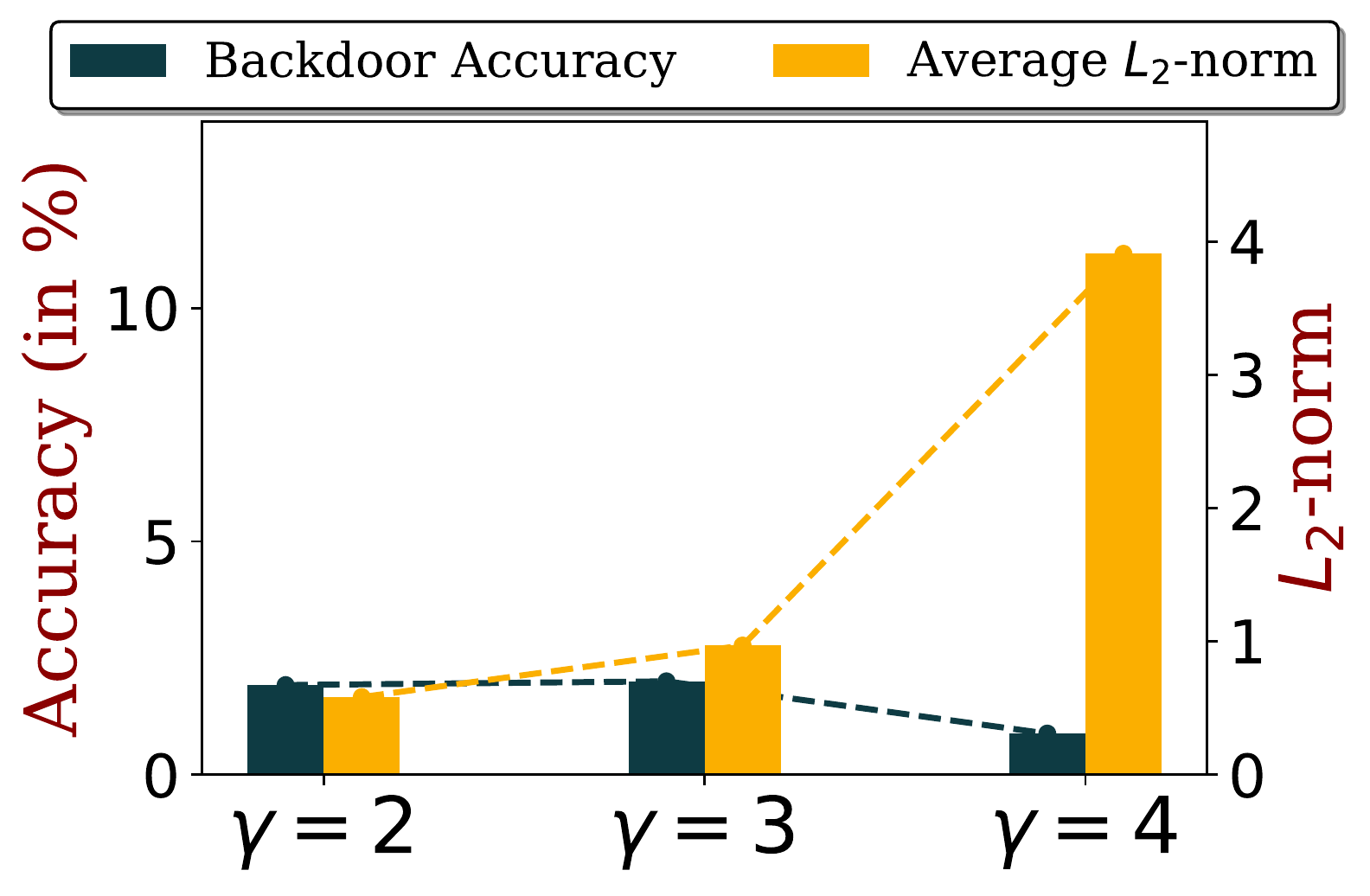}
         \caption{\textbf{ID4:} Random Selection}
     \end{subfigure}
\caption{Performance evaluation of the proposed backdoor removal method for varying $\gamma$ values across all backdoor insertion settings and poisoning strategies, illustrating the trade-off between backdoor accuracy (black bars) and the average deviation of the malicious model updates from benign model updates (yellow bars).}\label{fig:ablation:gamma}
\end{figure*}

\subsubsection{Unlearning with different \texorpdfstring{$\gamma$}{gamma} values}\label{sec:ablation_gamma}
In Section~\ref{sec:dynamic_penalization}, we presented the dynamic penalization approach, where we utilized the hyperparameter $\gamma$ to modulate the influence of the penalty term on the loss function. In this section, we assess the performance of the proposed backdoor removal method for varying gamma values. We adopt the weighted penalization strategy described in Equation~(\ref{eq:final_unlearn}) as the unlearn loss function. Without loss of generality, we consider $\gamma=2$, $\gamma=3$, and $\gamma=4$ for this analysis, taking into account all backdoor insertion settings outlined in Table~\ref{table:attack_settings} and the three poisoning strategies discussed previously. The results of our analysis are illustrated in Figure~\ref{fig:ablation:gamma}. In the figure, the black bars represent the final backdoor accuracy for different $\gamma$ values after 300 rounds for continuous selection and after 500 rounds for both fixed-frequency and random selection. The axes for these bars are located on the left side of each figure. The yellow bars correspond to the average deviation of parameters (in terms of the $L_2$-norm) for the local model of the compromised participant from the parameters of the local model of other benign participants, with their axes displayed on the right side of each figure. Each bar in the plot is computed by averaging the results of ten independent runs with distinct random seeds. Each plot also illustrates the variation trend for both backdoor accuracy and the average deviation as the $\gamma$ value increases using dashed lines with respective colors. From the figure, we can observe the trade-off between backdoor accuracy and the deviation of the malicious model updates from the benign model updates as $\gamma$ varies. As $\gamma$ increases, the final backdoor accuracy is minimized, as desired, but it also results in a model that deviates significantly from the models of other benign participants, making it detectable by the central server. This is expected, as higher $\gamma$ values penalize over-unlearning more, but in doing so, they considerably impact the loss function. The average deviation is close to or more than 3 for almost all attack settings in the figure, which is notably different when compared to the $L_2$-norms depicted in Figure~\ref{fig:unlearn_stealth}, where an average deviation of 3 in the $L_2$-norm can cause the model updates of the compromised participant to be markedly distinct from those of other participants. In our analysis, we consider $\gamma=3$ as it provides the optimal trade-off between backdoor accuracy and deviation of the compromised model from other benign~participants.

\subsubsection{Continuous vs. Non-Continuous Unlearning}\label{sec:ablation_non_continuous}
\begin{figure}[!t]
     \centering
     \begin{subfigure}[t]{0.48\linewidth}
         \centering
         \includegraphics[width=\linewidth]{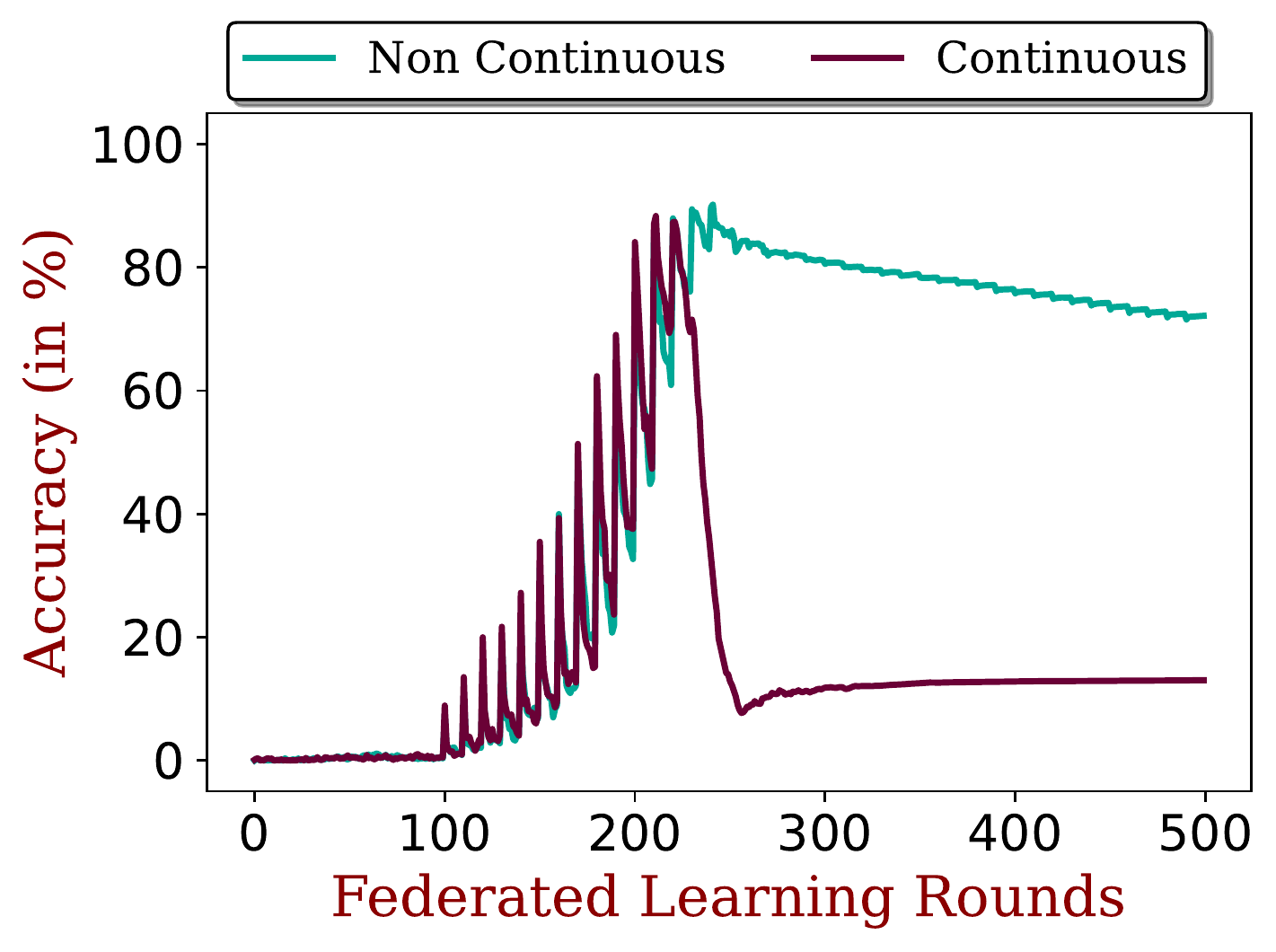}
         \caption{\textbf{ID1:} Fixed Freq. Selection}
     \end{subfigure}
     \begin{subfigure}[t]{0.48\linewidth}
         \centering
         \includegraphics[width=\linewidth]{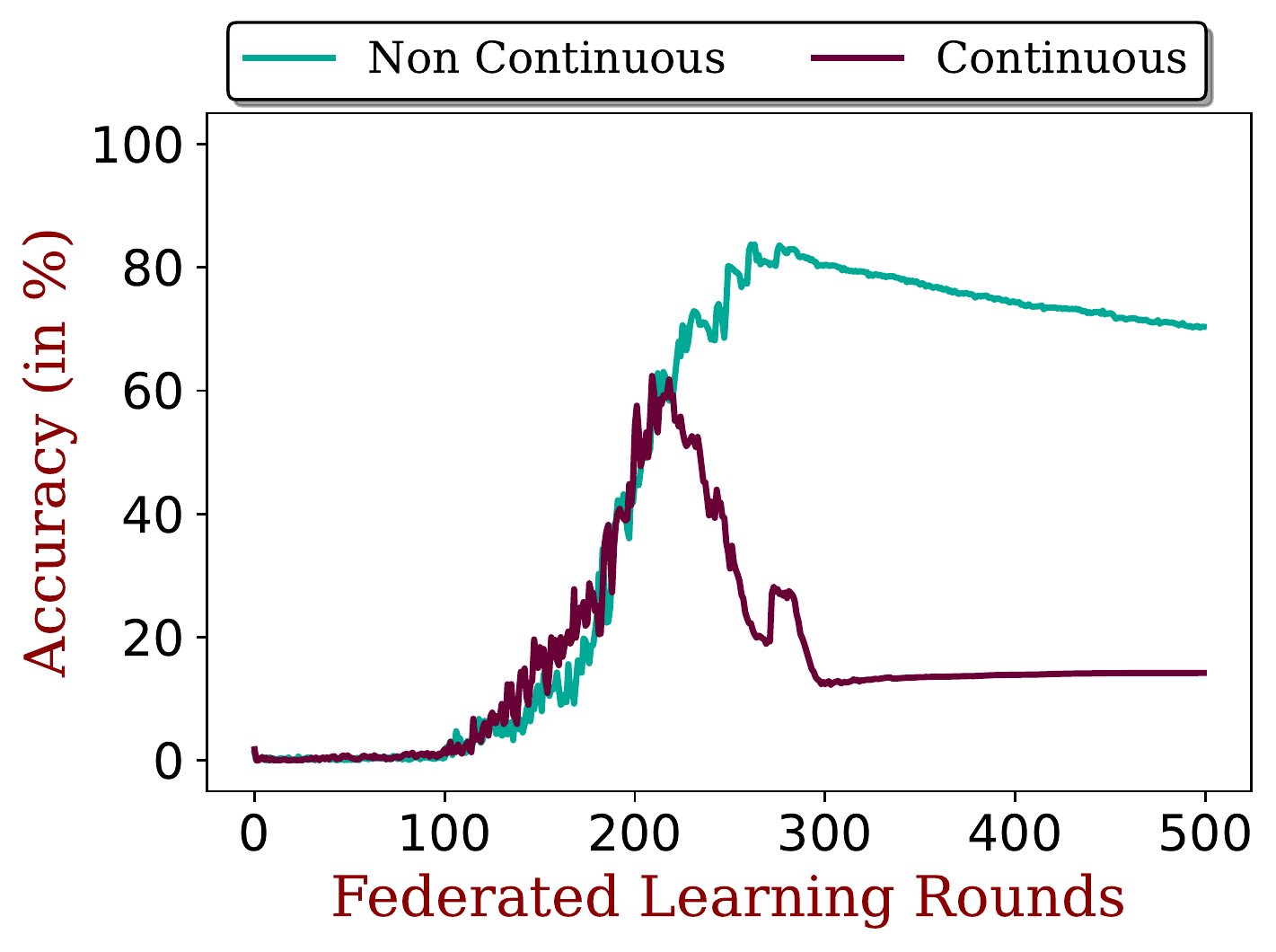}
         \caption{\textbf{ID1:} Random Selection}
     \end{subfigure}
     \begin{subfigure}[t]{0.48\linewidth}
         \centering
         \includegraphics[width=\linewidth]{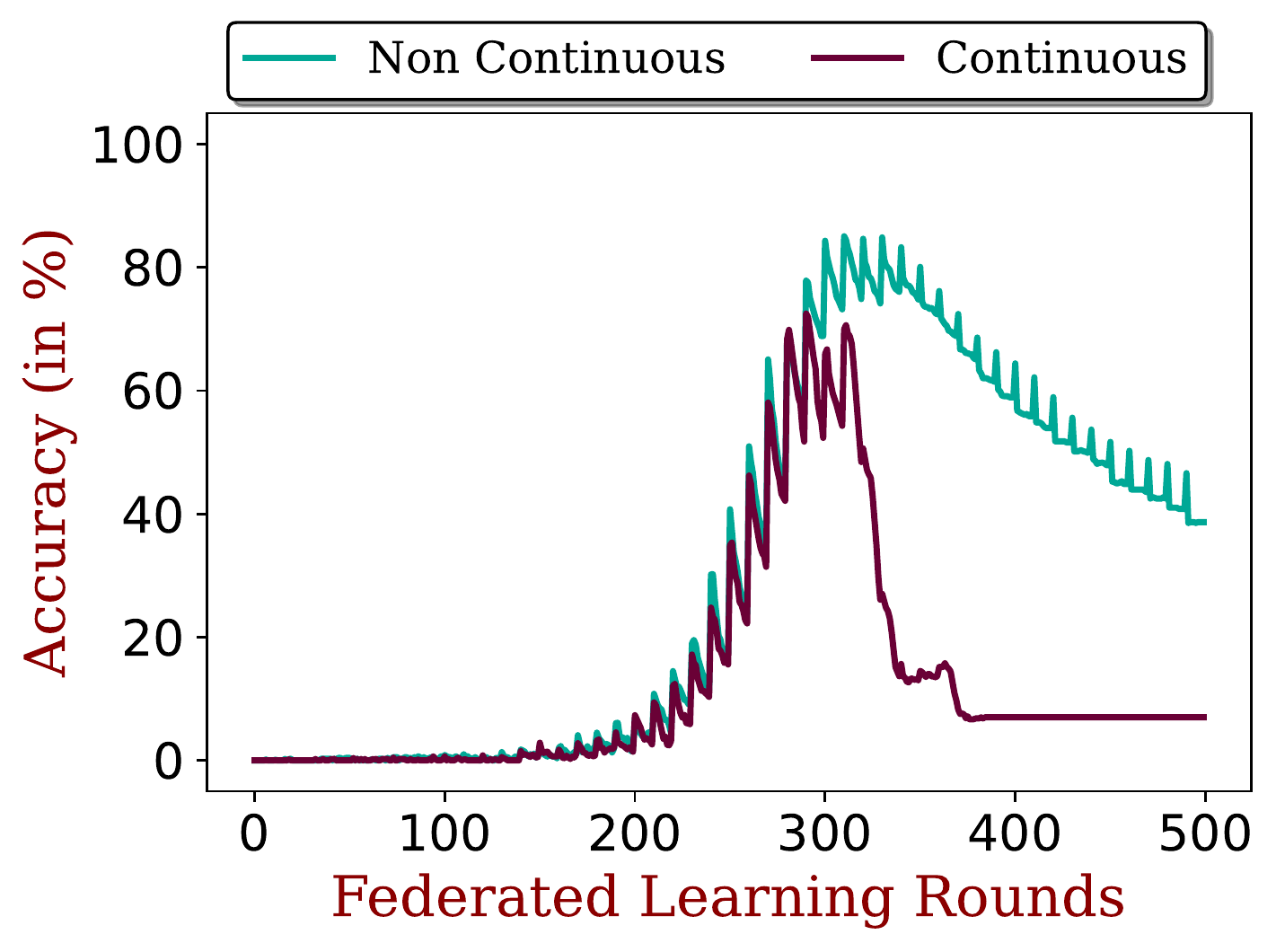}
         \caption{\textbf{ID2:} Fixed Freq. Selection}
     \end{subfigure}
     \begin{subfigure}[t]{0.48\linewidth}
         \centering
         \includegraphics[width=\linewidth]{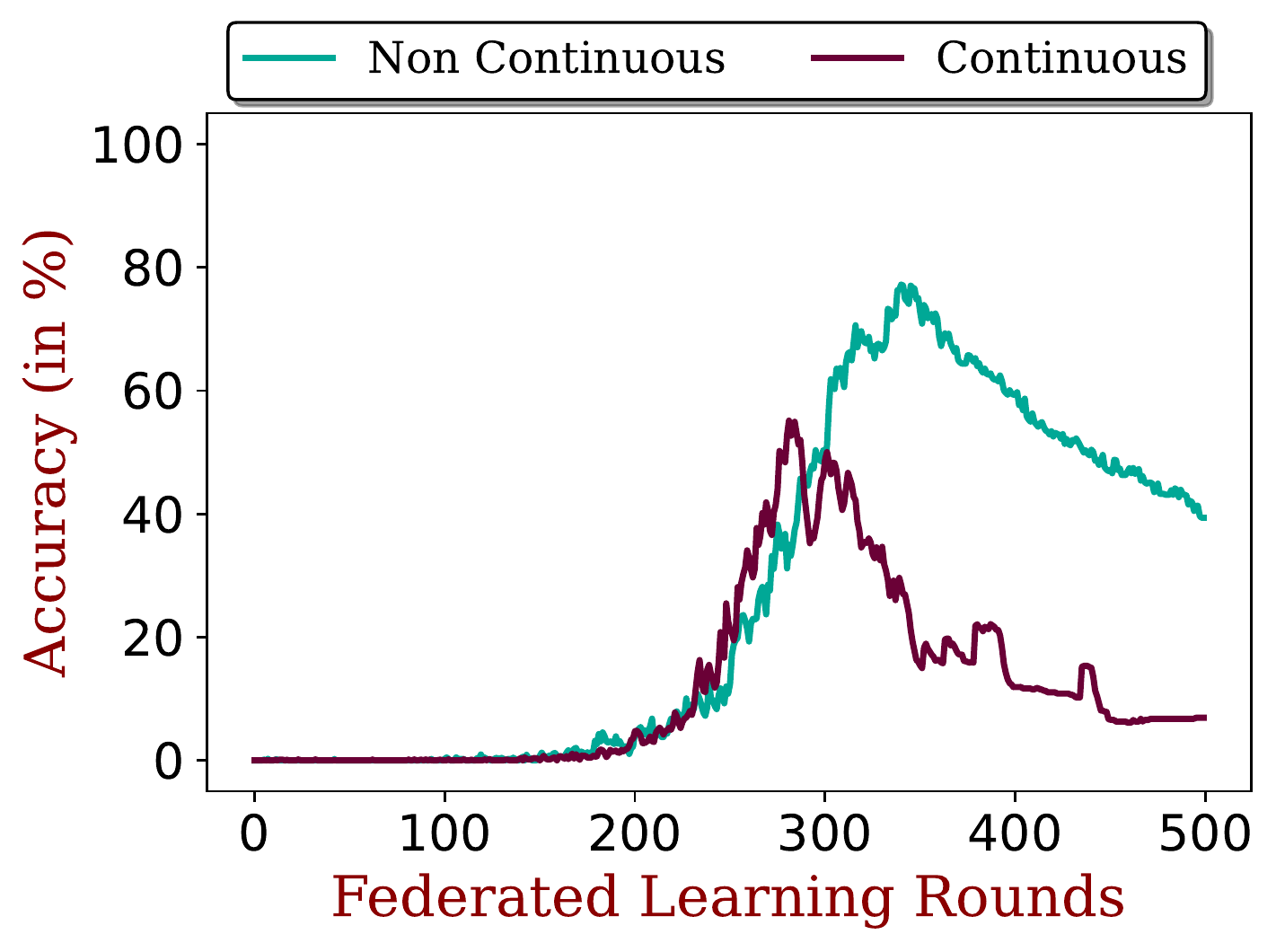}
         \caption{\textbf{ID2:} Random Selection}
     \end{subfigure}
     \begin{subfigure}[t]{0.48\linewidth}
         \centering
         \includegraphics[width=\linewidth]{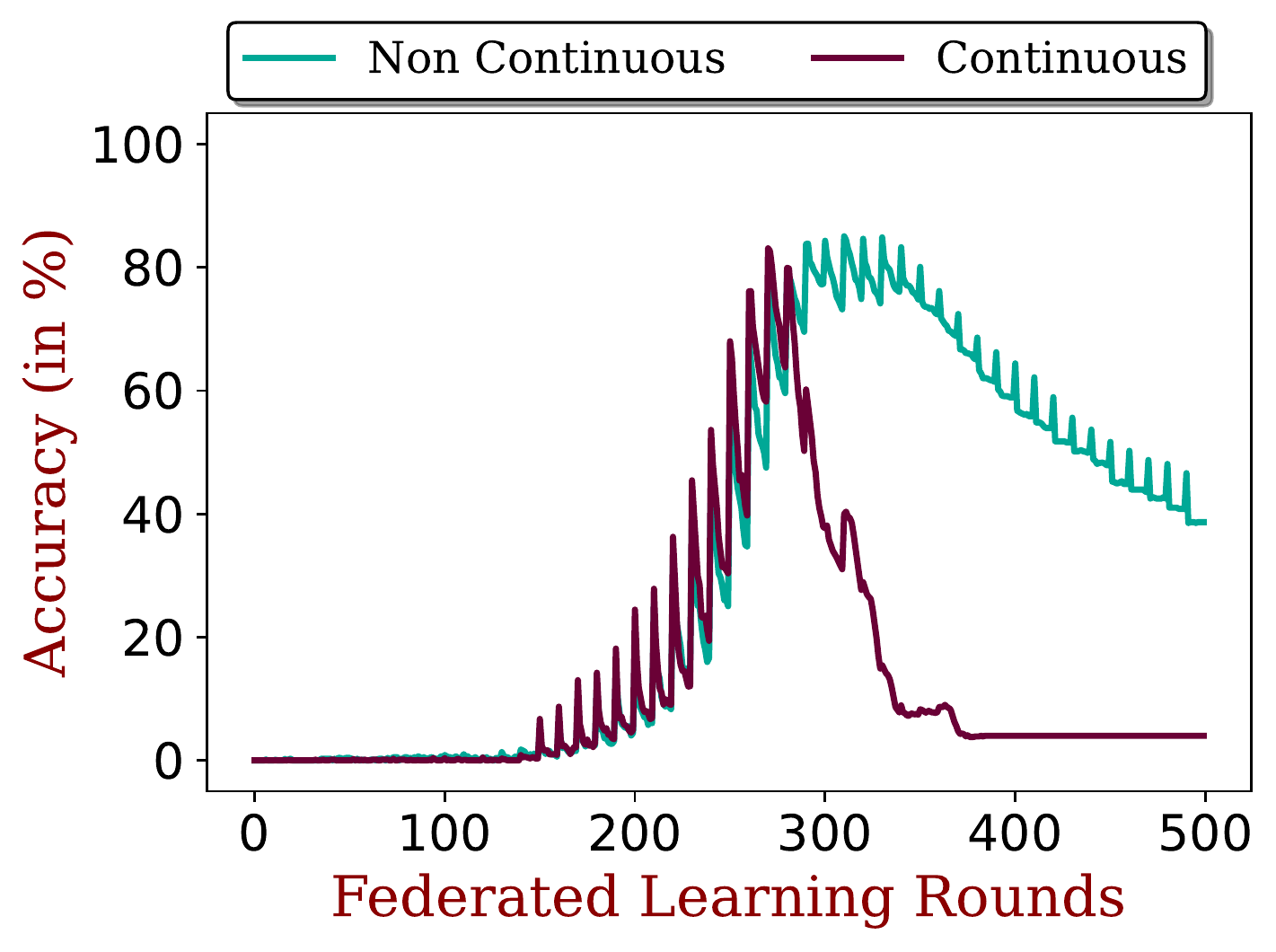}
         \caption{\textbf{ID3:} Fixed Freq. Selection}
     \end{subfigure}
     \begin{subfigure}[t]{0.48\linewidth}
         \centering
         \includegraphics[width=\linewidth]{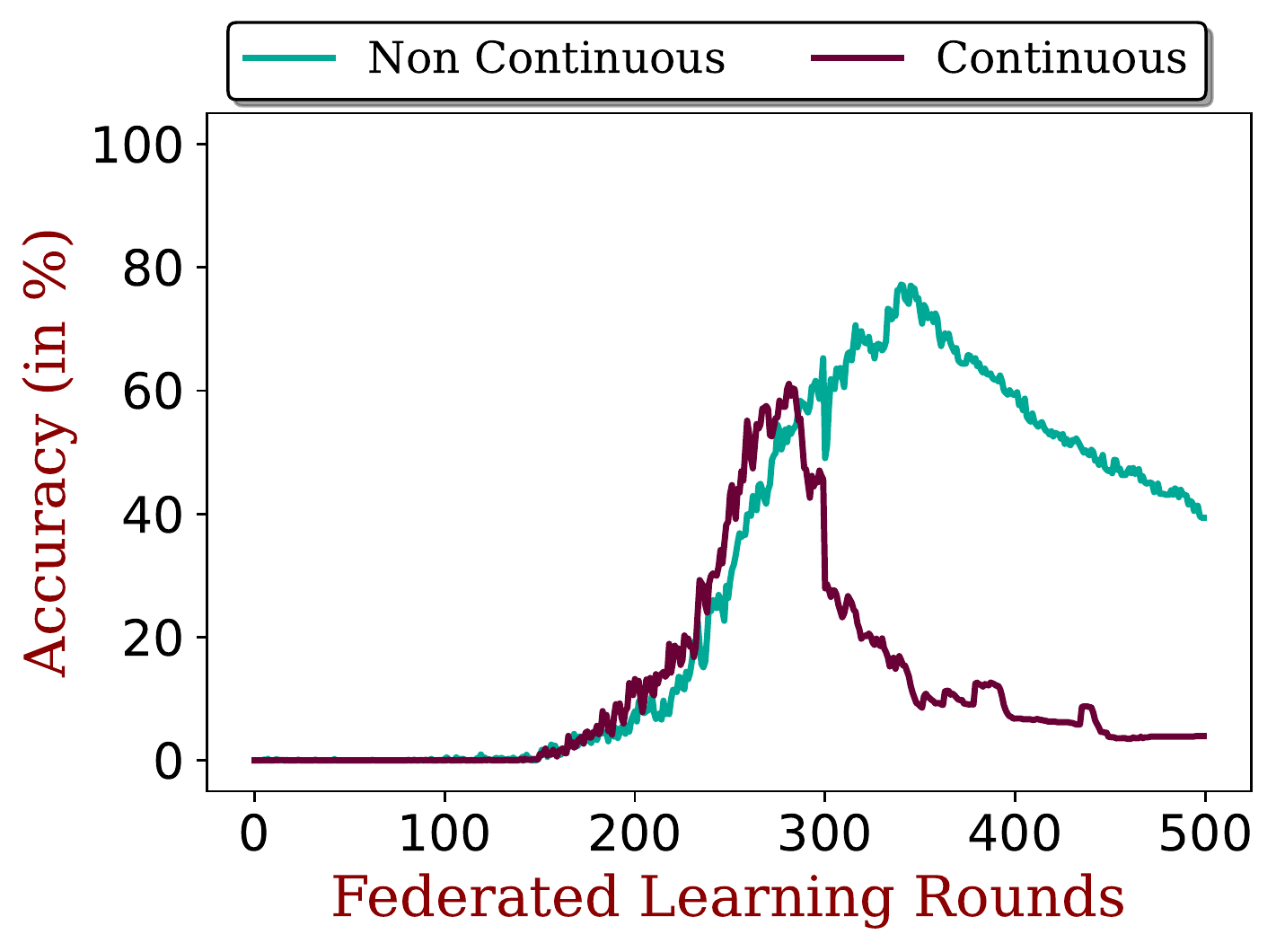}
         \caption{\textbf{ID3:} Random Selection}
     \end{subfigure}
     \begin{subfigure}[t]{0.48\linewidth}
         \centering
         \includegraphics[width=\linewidth]{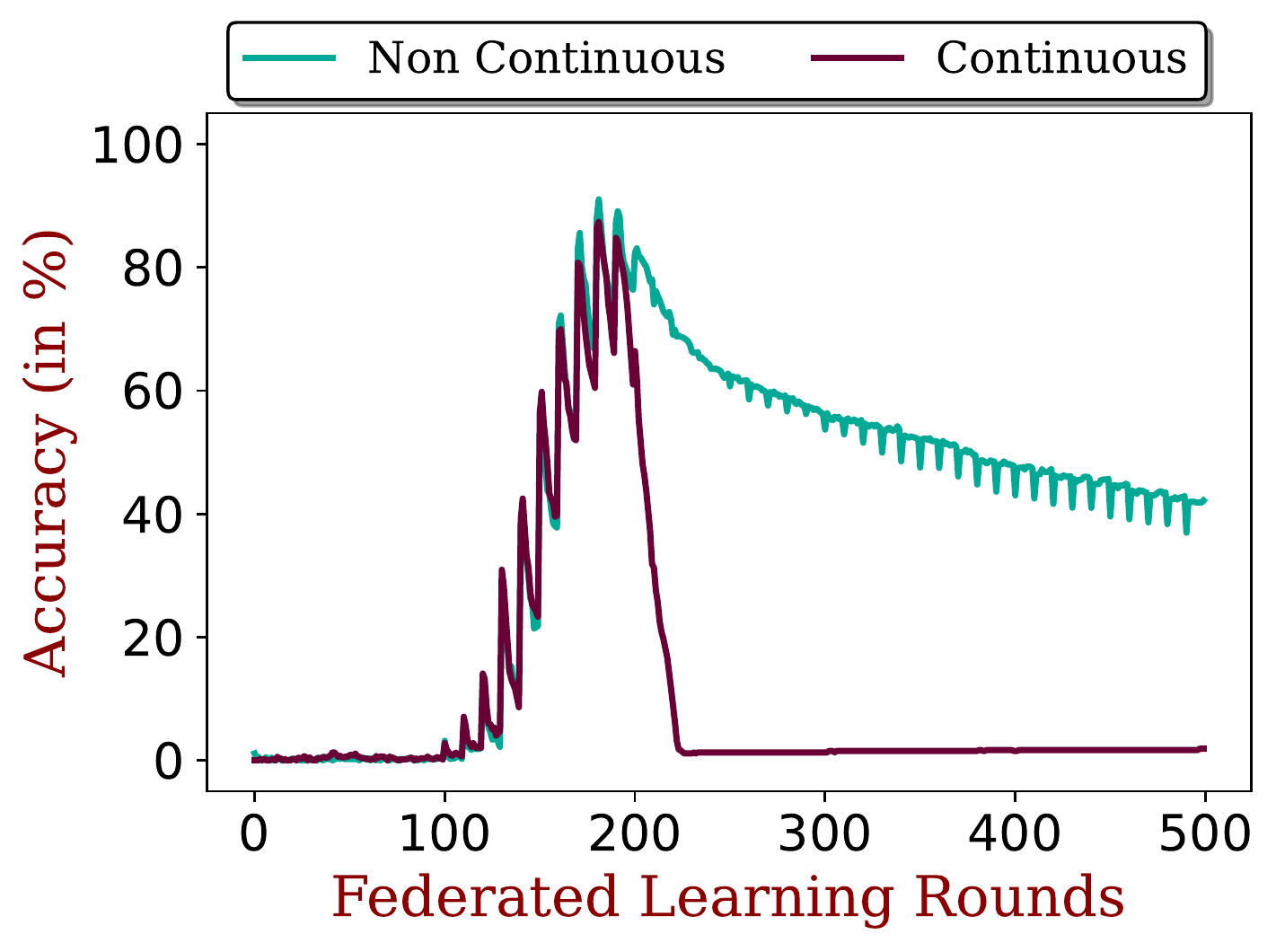}
         \caption{\textbf{ID4:} Fixed Freq. Selection}
     \end{subfigure}
     \begin{subfigure}[t]{0.48\linewidth}
         \centering
         \includegraphics[width=\linewidth]{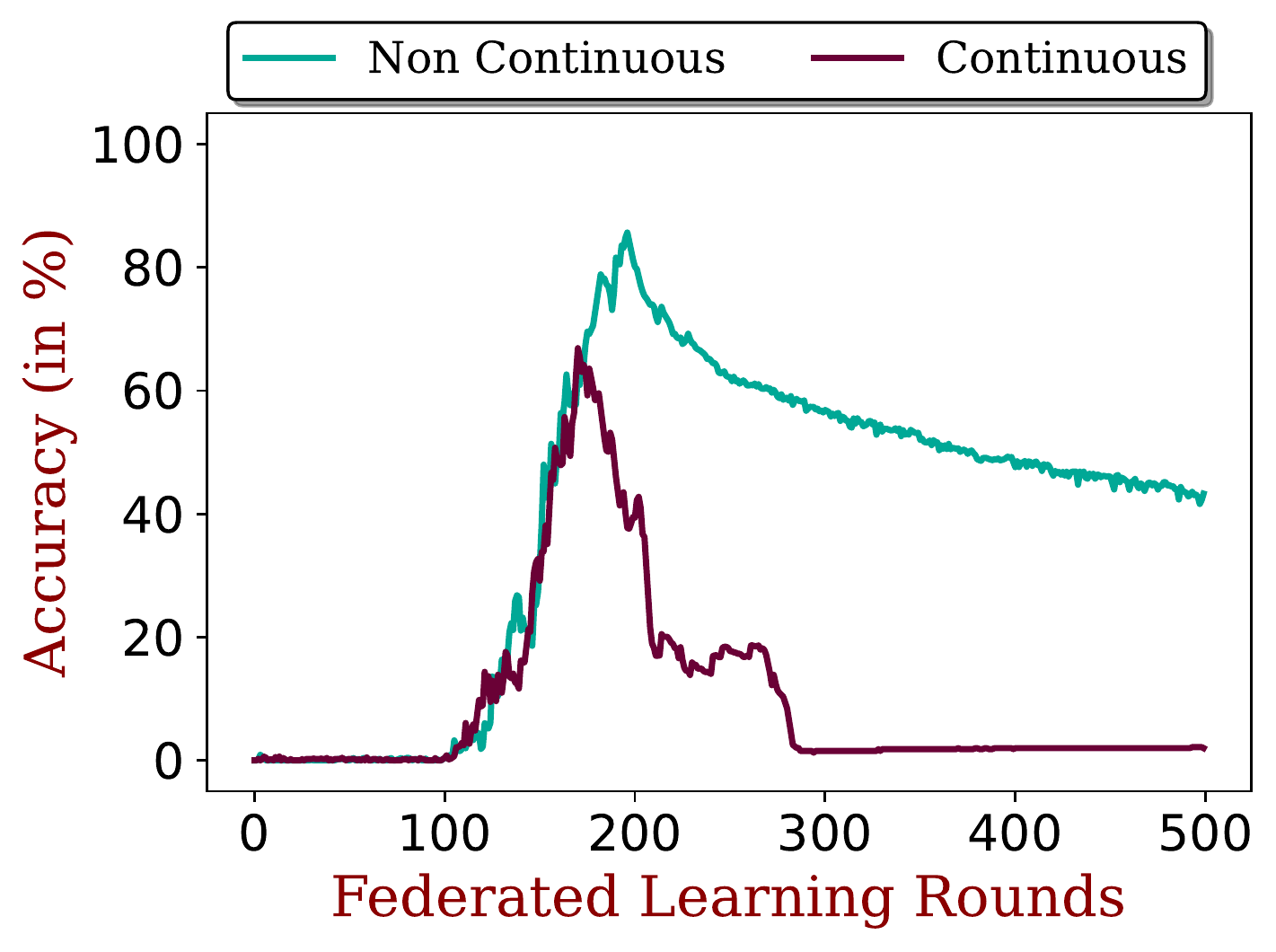}
         \caption{\textbf{ID4:} Random Selection}
     \end{subfigure}
\caption{Performance comparison between \textit{Non Continuous} and \textit{Continuous} poisoning strategies during backdoor removal. \textit{Non Continuous} employs the same strategy for both backdoor insertion and removal phases, while \textit{Continuous} uses continuous selection during removal.}\label{fig:ablation_non_continuous}
\end{figure}

In the previous discussions, we present our analyses based on the assumption that the adversary employs a continuous selection strategy only during the backdoor removal phase, with the aim of rapidly eliminating the traces of backdoors from the global model, independent of the strategy utilized in the backdoor insertion phase. In this section, we assess the performance of the proposed backdoor removal method while considering identical poisoning strategies for both the backdoor insertion and removal phases. This analysis considers all backdoor insertion settings as outlined in Table~\ref{table:attack_settings} and focuses on fixed-frequency and random selection strategies, as the results pertaining to continuous selection have already been covered in Figure~\ref{fig:unlearn_performance}. The results of this analysis are depicted in Figure~\ref{fig:ablation_non_continuous}, where the line plot labeled \textit{Non Continuous} represents backdoor accuracy when the same poisoning strategy is applied during both insertion and removal phases, while the plot labeled \textit{Continuous} represents backdoor accuracy considering the continuous selection strategy for backdoor removal.
\begin{figure*}[!t]
     \centering
     \begin{subfigure}[t]{0.355\linewidth}
         \centering
         \includegraphics[width=\linewidth]{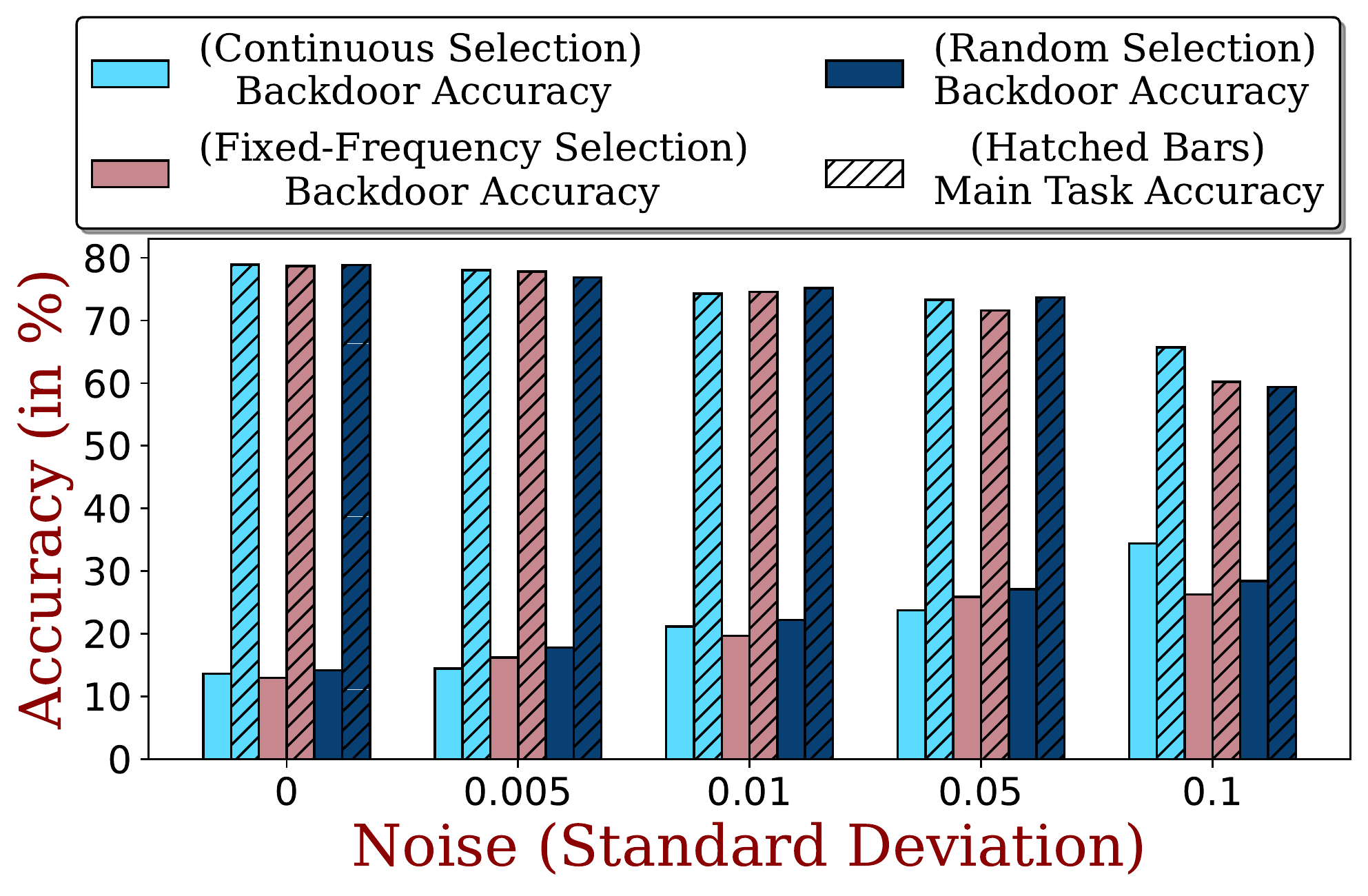}
         \caption{\textbf{ID1}}
     \end{subfigure}
     \begin{subfigure}[t]{0.355\linewidth}
         \centering
         \includegraphics[width=\linewidth]{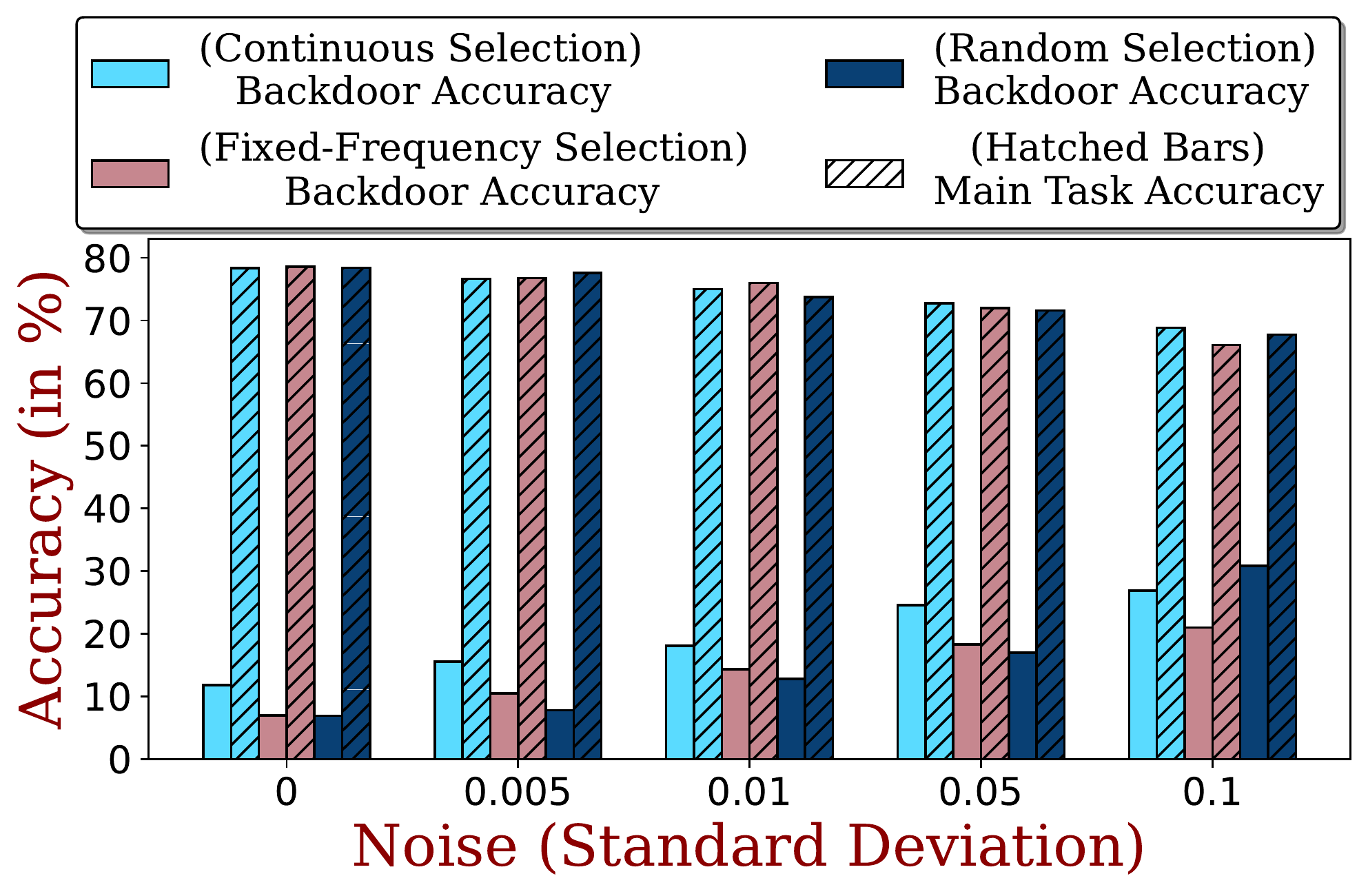}
         \caption{\textbf{ID2}}
     \end{subfigure}
     \begin{subfigure}[t]{0.355\linewidth}
         \centering
         \includegraphics[width=\linewidth]{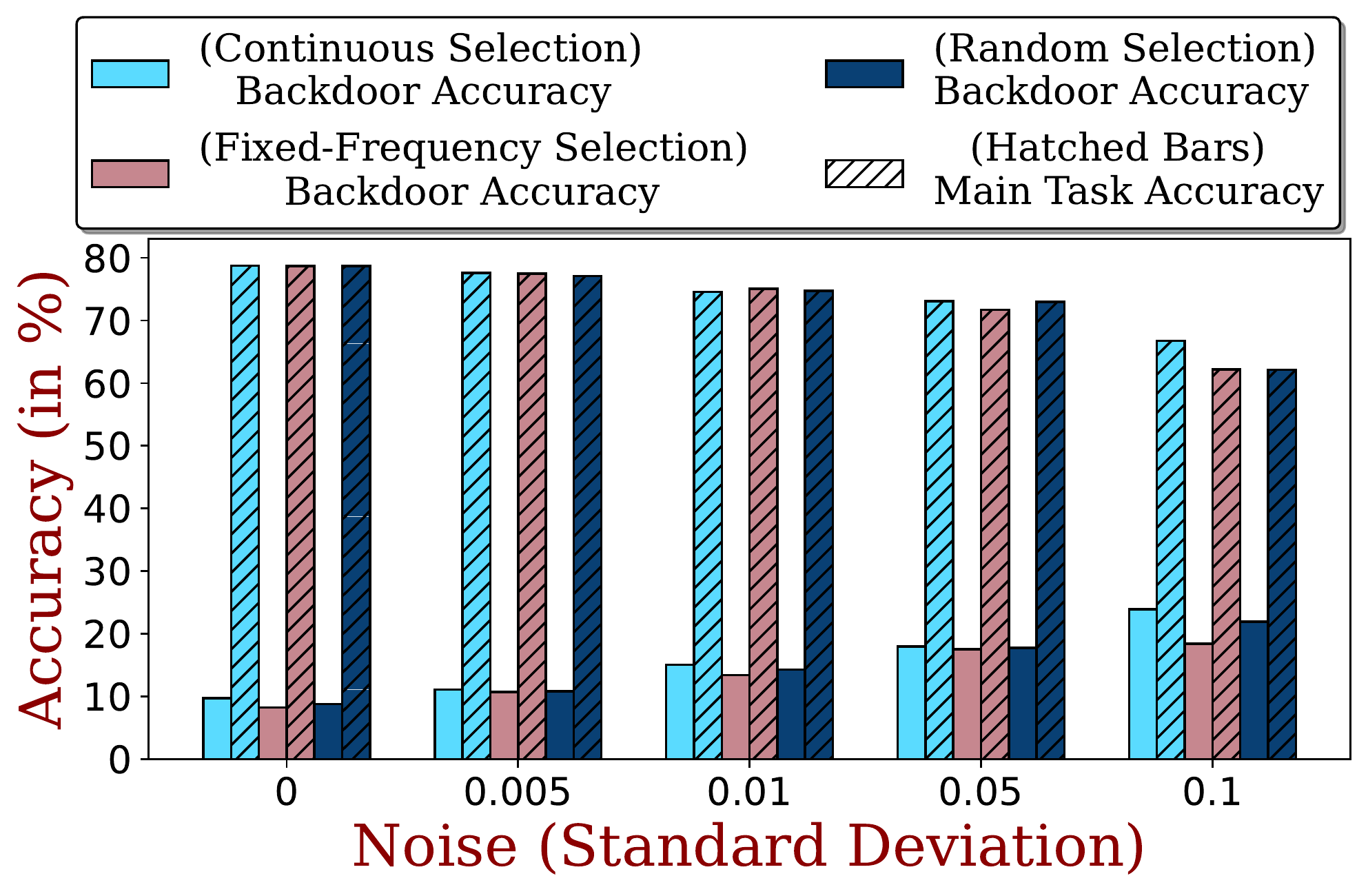}
         \caption{\textbf{ID3}}
     \end{subfigure}
     \begin{subfigure}[t]{0.355\linewidth}
         \centering
         \includegraphics[width=\linewidth]{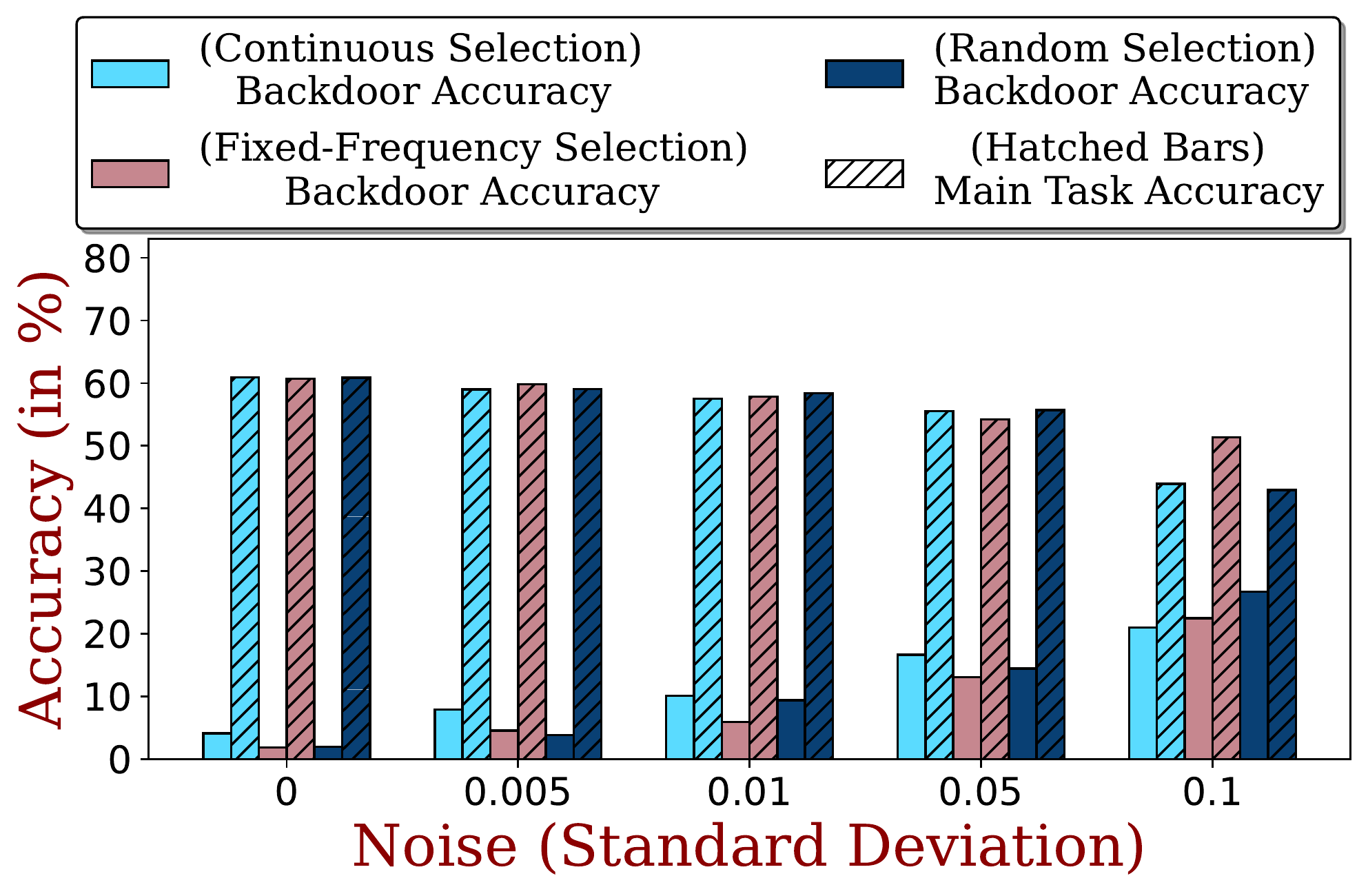}
         \caption{\textbf{ID4}}
     \end{subfigure}
\caption{The impact of Gaussian noise addition during model aggregation on the performance of the proposed backdoor removal method, with varying standard deviations. Solid bars represent final backdoor accuracy, while hatched bars indicate main task accuracy for corresponding poisoning strategies.}\label{fig:ablation_noise}
\end{figure*}
Each plot is derived by computing the mean across ten independent runs for different random seeds. As demonstrated in Figure~\ref{fig:ablation_non_continuous}, the continuous poisoning strategy successfully expedites backdoor removal from the global model for all the configurations, as anticipated. More specifically, using continuous unlearning the final backdoor accuracy of the global model is reduced by 57.62\% for ID1, 32.05\% for ID2, 35.04\% for ID3, and 40.76\% for ID4 on average over all poisoning strategies in comparison to scenarios where non-continuous unlearning is used. While the non continuous backdoor removal approach is capable of removing backdoors, it requires considerably more time compared to the continuous variant. Consequently, in our analysis, we assume that the adversary implements a continuous selection strategy exclusively during the backdoor removal phase.

\subsection{Unlearning in the presence of Noise Addition}
One of the useful techniques to prevent any malicious modification to the global model in any FL framework is to add noise to the model updates when a participant transmits the local model to the central server. Given that both backdoor attacks and removal influence various parameters within the global model with varying magnitudes, it is crucial to investigate the impact of noise addition on the proposed backdoor removal strategy. In this section, we evaluate the performance of the proposed backdoor removal method, considering that the central server introduces \textit{Gaussian} noise to the model updates during aggregation. Without loss of generality, we assume Gaussian noise with a zero mean and investigate varying standard deviations. The analysis includes all backdoor insertion settings outlined in Table~\ref{table:attack_settings} and all poisoning strategies discussed previously. Figure~\ref{fig:ablation_noise} illustrates the results for different standard deviations of the noise distribution. Solid bars represent the final backdoor accuracy after 300 rounds for continuous selection and after 500 rounds for both fixed-frequency and random selection. Hatched bars, on the other hand, represent the main task accuracy for corresponding poisoning strategies. As shown in Figure~\ref{fig:ablation_noise}, increasing the noise level diminishes the efficacy of the proposed method in removing backdoors, as backdoor accuracy rises with an increase in noise standard deviation. Simultaneously, a higher noise level also impacts the main task accuracy of the global model, which, as demonstrated in the figure, declines with an increase in noise standard deviation. Consequently, the addition of noise presents an interesting trade-off between model performance and the impact of the proposed backdoor removal method.

\section{Related Work}\label{sec:related_work}
In this section, we provide a concise overview of the relevant literature on backdoor attacks and applications of machine unlearning within FL frameworks.

\subsection{Backdoor Attacks in FL}
The backdoor attack in FL was first introduced by Bagdasaryan et al.~\cite{DBLP:conf/aistats/BagdasaryanVHES20}, demonstrating that amplifying the gradient magnitudes can effectively inject backdoors into the central model, even after the server aggregates the poisoned gradients with the gradients of other benign participants. Backdoor attacks typically operate through specific trigger patterns and input properties, with varying implications and consequences. Semantic backdoor attacks, for example, cause misclassification of inputs sharing a common semantic property, such as cars featuring green stripes~\cite{DBLP:conf/aistats/BagdasaryanVHES20,DBLP:conf/nips/WangSRVASLP20,DBLP:conf/icml/ZhangPSYMMR022}. In contrast, trigger-based backdoor attacks produce a specific output when presented with an input containing a designated ``trigger''~\cite{DBLP:conf/iclr/XieHCL20,DBLP:journals/corr/abs-2205-13523}. Semantic backdoor attacks can be further categorized into base-case attacks and edge-case attacks. Base-case attacks focus on inducing misclassification of data originating from the center of the target data distribution, such as poisoning a digit classification model to consistently predict the label ``9'' for images labeled ``5''~\cite{DBLP:journals/corr/abs-1911-07963,DBLP:conf/aistats/PandaMBCM22,DBLP:conf/icml/ZhangPSYMMR022}. Preserving benign accuracy while successfully altering the model's behavior on a significant portion of the target data distribution presents a considerable challenge~\cite{DBLP:conf/sp/ShejwalkarHKR22}. Consequently, prior works proposed edge-case attacks. Wang et al., for instance, demonstrated that backdoors derived from low-probability distribution segments could effectively bypass existing defenses~\cite{DBLP:conf/nips/WangSRVASLP20}. Recent advancements in backdoor attacks emphasize the need for the durability of injected backdoors, as they do not inherently persist through multiple rounds when adversaries cease to inject them~\cite{DBLP:conf/icml/ZhangPSYMMR022,DBLP:journals/corr/abs-2205-13523}. Zhang et al., demonstrated that targeting selective parameters of the global model during backdoor insertion can yield a backdoor with twice the durability of existing attacks in the literature~\cite{DBLP:conf/icml/ZhangPSYMMR022}.

\subsection{Machine Unlearning in FL}
Machine Unlearning in FL was first introduced by Liu et al., establishing a uniform metric known as the forgetting rate that allows for the evaluation of machine unlearning techniques in transitioning data from a ``memorized'' state to an ``unknown'' state following the unlearning process~\cite{DBLP:journals/corr/abs-2003-10933}. The technique has since been employed within FL frameworks to uphold the legal ``right to be forgotten'' for participants~\cite{DBLP:conf/spawc/GongSK21,DBLP:conf/iwqos/LiuMYWL21,DBLP:journals/corr/abs-2111-12056,DBLP:journals/network/WuGWHZD22,DBLP:conf/www/Wang0XQ22}. In light of the increasing significance of machine unlearning, extensive research has been dedicated to optimizing the efficiency and time requirements of various unlearning mechanisms~\cite{DBLP:conf/wsdm/YuanYWZHW23,DBLP:conf/infocom/LiuXYWL22}. Gao et al. further proposed a verifiable machine unlearning methodology, which enables participants to verify the effectiveness of the unlearning process before disengaging from the FL system~\cite{DBLP:journals/corr/abs-2205-12709}. The methodology guarantees that the unlearning effect is appropriately achieved, maintaining data privacy and compliance with regulations. Wu et al. introduced a strategy for removing a participant's contribution from the central model without the need for any participant data. Instead, the method relies on historical updates from the participant, providing an additional layer of privacy protection~\cite{DBLP:journals/corr/abs-2201-09441}. The approach proves particularly valuable in scenarios where access to participant data is restricted or undesirable. Halimi et al. proposed an efficient method for unlearning a participant's contribution from the FL system through a constrained maximization problem. The methodology does not necessitate global access to training data or historical updates, reducing the need for extensive data storage and management~\cite{DBLP:journals/corr/abs-2207-05521}. These unlearning methods primarily focus on removing benign data from the global model. In contrast, our proposed method targets a more difficult task of eliminating backdoor data, which closely resemble benign ones and may even appear as natural images. Hence, removing backdoor samples requires additional care to prevent accidental reduction of global model accuracy, as a reduced performance could alert the central server.

\section{Conclusion}\label{sec:conclusion}
In this paper, we introduce a machine unlearning-based methodology for backdoor removal in an FL framework, which assists an adversary in effectively removing persistent backdoors from the centralized model upon achieving their objectives or upon suspicion of possible detection. The proposed approach presents strategies that preserve the performance of the centralized model and simultaneously prevent over-unlearning of information unrelated to backdoor patterns, making the adversaries stealthy while removing backdoors. Exhaustive evaluation considering image classification scenarios demonstrates the effectiveness of the proposed method in efficient backdoor removal from the centralized model, injected by state-of-the-art attacks across multiple configurations.

In this work, we assume that the participants' data are independently and identically distributed (IID). This assumption is particularly relevant when developing an FL framework tailored to training image classification models utilizing medical datasets. Under the IID assumption, medical images collected by different clients are considered to be drawn from the same underlying distribution, which can be a reasonable approximation when clients are medical institutions with similar patient demographics, imaging equipment, and data collection protocols~\cite{DBLP:journals/npjdm/RiekeH0MRABGLMO20,DBLP:journals/jamia/ChangBLYBBRRK18,DBLP:journals/npjdm/SadilekLNKSRIMK21}. However, real-world data can often be non-IID, and extending the proposed backdoor unlearning framework to accommodate non-IID data distributions presents a promising direction for future~research.

\section*{Resources}
All implementations of this paper will be made open-source upon acceptance.

\bibliographystyle{ieeetr}
\bibliography{references}
\end{document}